\documentclass[10pt,twocolumn,letterpaper]{article}

\usepackage{cvpr}
\usepackage{times}
\usepackage{epsfig}
\usepackage{graphicx}
\usepackage{amsmath}
\usepackage{amssymb}
\usepackage{lipsum}
\usepackage{tabulary}
\usepackage{tabularx}
\usepackage{multirow}
\usepackage{subcaption}
\usepackage{footmisc}
\usepackage{verbatim}
\usepackage{color}
\usepackage{makecell}
\usepackage{hhline}

\usepackage{array}
\newcolumntype{C}[1]{>{\centering\let\newline\\\arraybackslash\hspace{0pt}}m{#1}}

\def\etal{\emph{et al.}}

\usepackage[pagebackref=true,breaklinks=true,letterpaper=true,colorlinks,bookmarks=false]{hyperref}

 \cvprfinalcopy 

\def\cvprPaperID{827} 
\def\cvprPaperID{827} 

\ifcvprfinal\fi
\begin{document}

\title{DSSD : Deconvolutional Single Shot Detector }

\author{Cheng-Yang Fu$^{1 }$\thanks{Equal Contribution} , Wei Liu$^{1}$\footnotemark[1], Ananth  Ranga$^2$,  Ambrish Tyagi$^2$, Alexander C. Berg$^1$\\
$^1$UNC Chapel Hill, $^2$Amazon Inc.\\
{\tt\small \{cyfu, wliu\}@cs.unc.edu, \{ananthr, ambrisht\}@amazon.com, aberg@cs.unc.edu}
}

\maketitle

\begin{abstract}
The main contribution of this paper is an approach for introducing
additional context into state-of-the-art general object detection. To achieve this we first combine a state-of-the-art
classifier (Residual-101~\cite{Residual}) with a fast detection framework
(SSD~\cite{SSD}). We then augment SSD+Residual-101 with deconvolution
layers to introduce additional large-scale context in object detection
and improve accuracy, especially for small objects, calling our resulting system DSSD for deconvolutional single shot detector.  While
these two contributions are easily described at a high-level, a naive
implementation does not succeed. Instead we show that carefully adding additional stages of learned transformations, specifically a module for feed-forward connections in deconvolution and a new output module, enables this new approach and forms a potential way forward for further detection research. Results are shown on both PASCAL VOC and COCO detection. Our DSSD with $513 \times  513$ input achieves 81.5\% mAP on VOC2007 \texttt{test}, 80.0\% mAP on VOC2012 \texttt{test}, and 33.2\% mAP on COCO, outperforming a state-of-the-art method R-FCN~\cite{R-FCN} on each dataset.

\end{abstract}

\section{Introduction}
The main contribution of this paper is an approach for introducing
additional context into state-of-the-art general object detection. The
end result achieves the current highest accuracy for detection with a single
network on PASCAL VOC~\cite{everingham2010PASCAL} while also
maintaining comparable speed with a previous state-of-the-art detection~\cite{R-FCN}.  To achieve this we first combine a state-of-the-art classifier (Residual-101~\cite{Residual}) with a fast detection framework
(SSD~\cite{SSD}). We then augment SSD+Residual-101 with deconvolution
layers to introduce additional large-scale context in object detection
and improve accuracy, especially for small objects, calling our resulting system DSSD for deconvolutional single shot detector.  While
these two contributions are easily described at a high-level, a naive
implementation does not succeed.  Instead we show that carefully adding additional stages of learned transformations, specifically a module for feed forward connections in deconvolution and a new output module, enables this new approach and forms a potential way forward for further detection research.



Putting this work in context, there has been a recent move in object detection back toward sliding-window techniques in the last two years. The idea is that instead of first proposing potential bounding boxes for objects in an image and then classifying them, as exemplified in selective search\cite{uijlings2013selective} and R-CNN\cite{R-CNN} derived methods, a classifier is applied to a fixed set of possible bounding boxes in an image.  While sliding window approaches never completely disappeared, they had gone out of favor after the heydays of HOG~\cite{hog} and DPM~\cite{felzenszwalb2008discriminatively} due to the increasingly large number of box locations that had to be considered to keep up with state-of-the-art.  They are coming back as more powerful machine learning frameworks integrating deep learning are developed.  These allow fewer potential bounding boxes to be considered, but in addition to a classification score for each box, require predicting an offset to the actual location of the object---snapping to its spatial extent.  Recently these approaches have been shown to be effective for bounding box proposals~\cite{multibox,faster-RCNN} in place of bottom-up grouping of segmentation~\cite{uijlings2013selective,R-CNN}. Even more recently, these approaches were used to not only score bounding boxes as potential object locations, but to simultaneously predict scores for object categories, effectively combining the steps of region proposal and classification. This is the approach taken by You Only Look Once (YOLO)~\cite{redmon2015you} which computes a global feature map and uses a fully-connected layer to predict detections in a fixed set of regions.  Taking this single-shot approach further by adding layers of feature maps for each scale and using a convolutional filter for prediction, the Single Shot MultiBox Detector (SSD)~\cite{SSD} is significantly more accurate and is currently the best detector with respect to the speed-vs-accuracy trade-off.

\begin{figure*}[ht]
  \includegraphics[width=\textwidth]{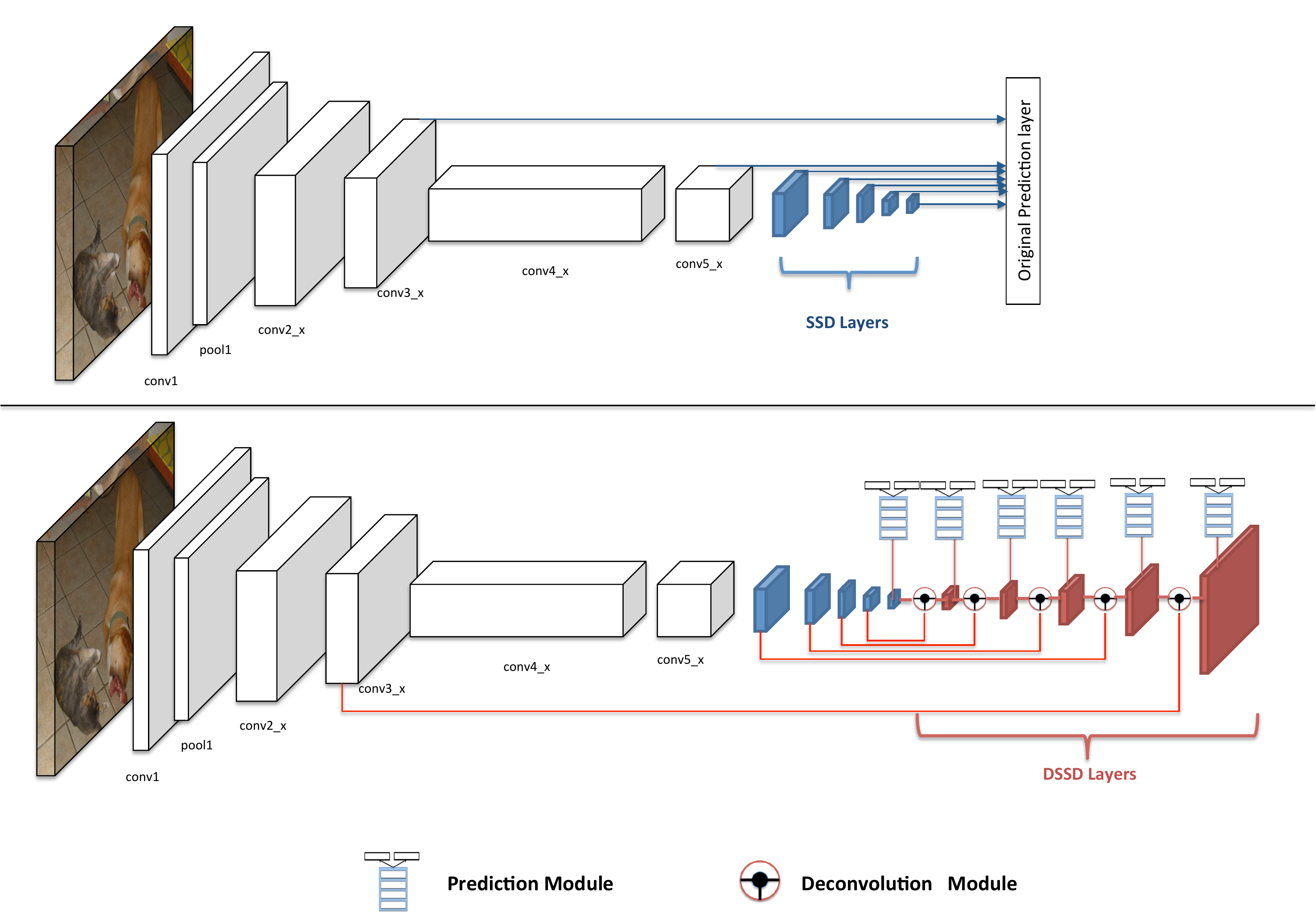}
  \caption{\textbf{Networks of SSD and DSSD on residual network.} The blue modules are the layers added in SSD framework, and we call them SSD Layers. In the bottom figure, the red layers are DSSD layers.  }
     \label{fig:arci}
\end{figure*}

When looking for ways to further improve the accuracy of detection, obvious targets are better feature networks and adding more context, especially for small objects, in addition to improving the spatial resolution of the bounding box prediction process.  Previous versions of SSD were based on the VGG~\cite{vgg} network, but many researchers have achieved better accuracy for tasks using Residual-101~\cite{Residual}.  Looking to concurrent research outside of detection, there has been a work on integrating context using so called ``encoder-decoder'' networks where a bottleneck layer in the middle of a network is used to encode information about an input image and then progressively larger layers decode this into a map over the whole image.  The resulting wide, narrow, wide structure of the network is often referred to as an hour-glass. These approaches have been especially useful in recent works on semantic segmentation~\cite{D-segmentation}, and human pose estimation~\cite{hourglasses}.

Unfortunately neither of these modifications, using the much deeper Residual-101, or adding deconvolution layers to the end of SSD feature layers, work ``out of the box''.  Instead it is necessary to carefully construct combination modules for integrating deconvolution, and output modules to insulate the Residual-101 layers during training and allow effective learning.

The code will be open sourced with models upon publication.

\begin{figure*}[htbp]
  \includegraphics[width=\textwidth]{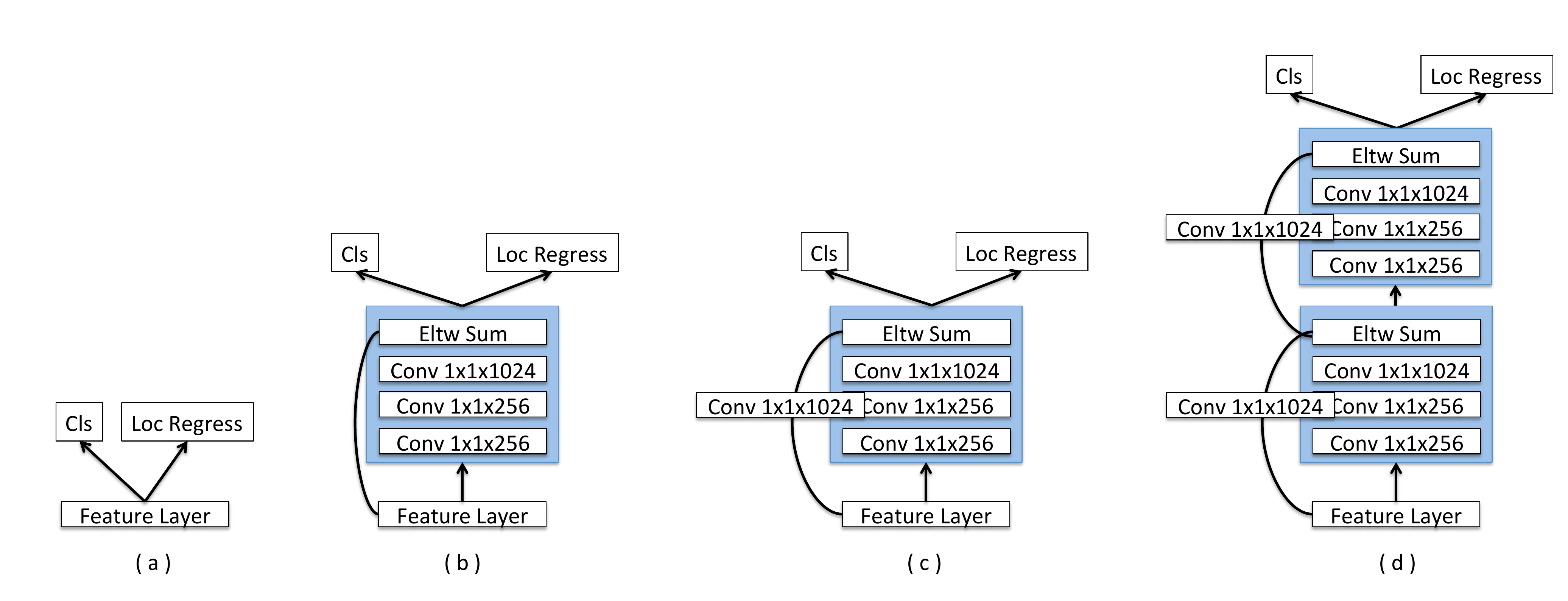}
  \caption{Variants of the prediction module }
     \label{fig:prediction_module}
\end{figure*}

\section{Related Work}

The majority of object detection methods, including SPPnet~\cite{he2014spatial}, Fast R-CNN~\cite{fast-RCNN}, Faster R-CNN~\cite{faster-RCNN}, R-FCN~\cite{R-FCN} and YOLO~\cite{redmon2015you}, use the top-most layer of a ConvNet to learn to detect objects at different scales. Although powerful, it imposes a great burden for a single layer to model all possible object scales and shapes.

There are variety of ways to improve detection accuracy by exploiting multiple layers within a ConvNet. The first set of approaches combine feature maps from different layers of a ConvNet and use the combined feature map to do prediction. ION~\cite{insideout-net} uses L2 normalization~\cite{liu2015parsenet} to combine multiple layers from VGGNet and pool features for object proposals from the combined layer. HyperNet~\cite{kong2016hypernet} also follows a similar method and uses the combined layer to learn object proposals and to pool features. Because the combined feature map has features from different levels of abstraction of the input image, the pooled feature is more descriptive and is better suitable for localization and classification. However, the combined feature map not only increases the memory footprint of a model significantly but also decreases the speed of the model.

Another set of methods uses different layers within a ConvNet to predict objects of different scales. Because the nodes in different layers have different receptive fields, it is natural to predict large objects from layers with large receptive fields (called higher  or later layers within a ConvNet) and use layers with small receptive fields to predict small objects. SSD~\cite{SSD} spreads out default boxes of different scales to multiple layers within a ConvNet and enforces each layer to focus on predicting objects of certain scale. MS-CNN~\cite{cai2016unified} applies deconvolution on multiple layers of a ConvNet to increase feature map resolution before using the layers to learn region proposals and pool features. 
However, in order to detect small objects well, these methods need to use some information from shallow layers with small receptive fields and dense feature maps, which may cause low performance on small objects because shallow layers have less semantic information about objects. By using deconvolution layers and skip connections, we can inject more semantic information in dense (deconvolution) feature maps, which in turn helps predict small objects.

There is another line of work which tries to include context information for prediction. Multi-Region CNN~\cite{MR-CNN} pools features not only from the region proposal but also pre-defined regions such as half parts, center, border and the context area. Following many existing works on semantic segmentation~\cite{D-segmentation} and pose estimation~\cite{hourglasses}, we propose to use an encoder-decoder hourglass structure to pass context information before doing prediction. The deconvolution layers not only addresses the problem of shrinking resolution of feature maps in convolution neural networks, but also brings in context information for prediction.

\section{Deconvolutional Single Shot Detection (DSSD) model}

We begin by reviewing the structure of SSD and then describe the new prediction module that produces significantly improved training effectiveness when using Residual-101 as the base network for SSD.  Next we discuss how to add deconvolution layers to make a hourglass network, and how to integrate the the new deconvolutional module to pass semantic context information for the final DSSD model.

\subsection*{SSD}
The Single Shot MultiBox Detector (SSD~\cite{SSD}) is built on top of a "base" network that ends (or is truncated to end) with some convolutional layers.  SSD adds a series of progressively smaller convolutional layers as shown in blue on top of Figure \ref{fig:arci} (the base network is shown in white). Each of the added layers, and some of the earlier base network layers are used to predict scores and offsets for some predefined default bounding boxes.  These predictions are performed by 3x3x\#channels dimensional filters, one filter for each category score and one for each dimension of the bounding box that is regressed.  It uses non-maximum suppression (NMS) to post-process the predictions to get final detection results.  More details can be found in \cite{SSD}, where the detector uses VGG~\cite{vgg} as the base network.

\subsection{Using Residual-101 in place of VGG}
Our first modification is using Residual-101 in place of VGG used in the original SSD paper, in particular we use the Residual-101 network from~\cite{Residual}.  The goal is to improve accuracy.  Figure \ref{fig:arci} top shows SSD with Residual-101 as the base network.  Here we are adding layers after the conv5\_x block, and predicting scores and box offsets from conv3\_x, conv5\_x, and the additional layers.  By itself this does not improve results.  Considering the ablation study results in Table~\ref{tab:ablation},  the top row shows a mAP of 76.4 of SSD with Residual-101 on $321 \times 321$ inputs for PASCAL VOC 2007 test.  This is lower than the 77.5 for SSD with VGG on $300 \times 300$ inputs (see Table\ref{tab:voc07}).  However adding an additional prediction module, described next, increases performance significantly.

\subsection*{Prediction module}
In the original SSD~\cite{SSD}, the objective functions are applied on the selected feature maps directly and a L2 normalization layer is used for the conv4\_3 layer, because of the large magnitude of the gradient.  MS-CNN\cite{cai2016unified} points out that improving the sub-network of each task can improve accuracy.  Following this principle, we add one residual block for each prediction layer as shown in Figure ~\ref{fig:prediction_module} variant (c).  We also tried the original SSD approach (a) and a version of the residual block with a skip connection (b) as well as two sequential residual blocks (d).  Ablation studies with the different prediction modules are shown in Table~\ref{tab:ablation} and discussed in Section~\ref{sec:experiments}.  We note that Residual-101 and the prediction module seem to perform significantly better than VGG without the prediction module for higher resolution input images.



\begin{figure}[htbp]
  \begin{center}
  \includegraphics[width=\linewidth]{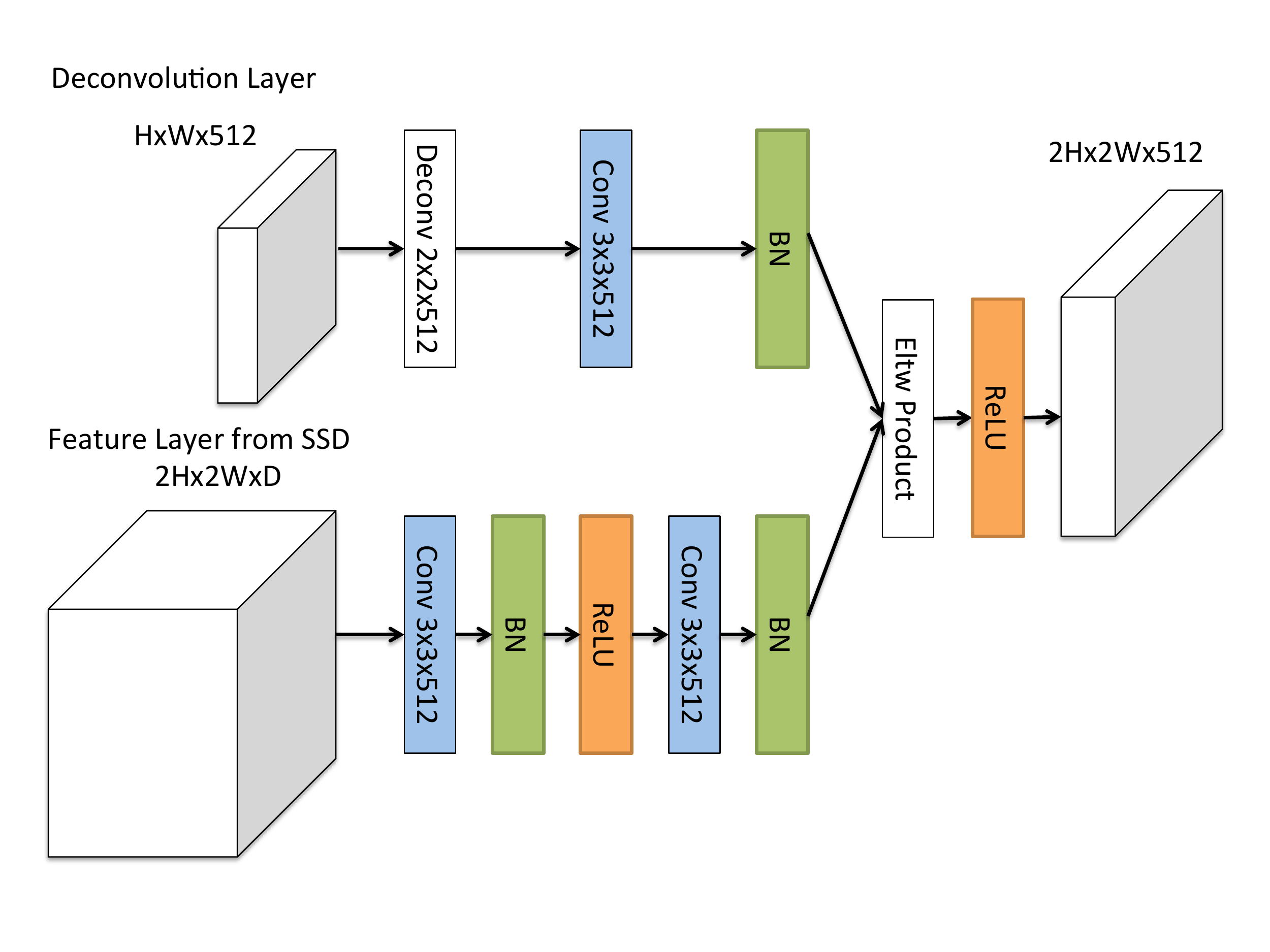}
  \end{center}
  \caption{Deconvolution module }
     \label{fig:Deconvolution_module}
\end{figure}

\subsection*{Deconvolutional SSD}
In order to include more high-level context in detection, we move prediction to a series of deconvolution layers placed after the original SSD setup, effectively making an asymmetric hourglass network structure, as shown in Figure~\ref{fig:arci} bottom.  The DSSD model in our experiments is built on SSD with Residual-101. Extra deconvolution layers are added to successively increase the resolution of feature map layers. In order to strengthen features, we adopt the "skip connection" idea from the Hourglass model \cite{hourglasses}. Although the hourglass model contains symmetric layers in both the Encoder and Decoder stage, we make the decoder stage extremely shallow for two reasons. First, detection is a fundamental task in vision and may need to provide information for the downstream tasks. Therefore, speed is an important factor. Building the symmetric network means the time for inference will double. This is not what we want in this fast detection framework. Second, there are no pre-trained models which include a decoder stage trained on the classification task of ILSVRC CLS-LOC dataset~\cite{Imagenet} because classification gives a single whole image label instead of a local label as in detection. State-of-the-art detectors rely on the power of transfer learning. The model pre-trained on the classification task of ILSVRC CLS-LOC dataset~\cite{Imagenet} makes the accuracy of our detector higher and converge faster compared to a randomly initialized model. Since there is no pre-trained model for our decoder, we cannot take the advantage of transfer learning for the decoder layers which must be trained starting from random initialization.  An important aspect of the deconvolution layers is computational cost, especially when adding information from the previous layers in addition to the deconvolutional process.

\subsection*{Deconvolution Module}
In order to help integrating information from earlier feature maps and the deconvolution layers, we introduce a deconvolution module as shown in Figure~\ref{fig:Deconvolution_module}.  This module fits into the overall DSSD architecture as indicated by the solid circles in the bottom of Figure~\ref{fig:arci}.  The deconvolution module is inspired by Pinheiro \etal~\cite{RefineMask} who suggested that a factored version of the deconvolution module for a refinement network has the same accuracy as a more complicated one and the network will be more efficient. We make the following modifications and show them in Figure~\ref{fig:Deconvolution_module}. First, a batch normalization layer is added after each convolution layer. 
Second, we use the learned deconvolution layer instead of bilinear upsampling. 
Last, we test different combination methods: element-wise sum and element-wise product. The experimental results show that the element-wise product provides the best accuracy (See Table~\ref{tab:ablation} bottom sections).

\subsection*{Training}
We follow almost the same training policy as SSD. First, we have to match a set of default boxes to target ground truth boxes. For each ground truth box, we match it with the best overlapped default box and any default boxes whose Jaccard overlap is larger than a threshold (e.g. 0.5). Among the non-matched default boxes, we select certain boxes as negative samples based on the confidence loss so that the ratio with the matched ones is 3:1. Then we minimize the joint localization loss (e.g. Smooth L1) and confidence loss (e.g Softmax). Because there is no feature or pixel resampling stage as was done in Fast or Faster R-CNN, it relies on extensive data augmentation which is done by randomly cropping the original image plus random photometric distortion and random flipping of the cropped patch. Notably, the latest SSD also includes a random expansion augmentation trick which has proved to be extremely helpful for detecting small objects, and we also adopt it in our DSSD framework.

We have also made a minor change in the prior box aspect ratio setting. In the original SSD model, boxes with aspect ratios of 2 and 3 were proven useful from the experiments. In order to get an idea of the aspect ratios of bounding boxes in the training data (PASCAL VOC 2007 and 2012 \texttt{trainval}), we run K-means clustering on the training boxes with square root of box area as the feature. We started with two clusters, and increased the number of clusters if the error can be improved by more than 20\%. We converged at seven clusters and show the result in Table ~\ref{tab:aspect_ratio}.
Because the SSD framework resize inputs to be square and most training images are wider, it is not surprising that most bounding boxes are taller. 
 Based on this table we can see that most box ratios fall within a range of 1-3. Therefore we decide to add one more aspect ratio, 1.6, and use (1.6, 2.0, 3.0) at every prediction layer.

 \begin{table}
 \centering
   \setlength{\tabcolsep}{5.4pt}
    \begin{tabular}{|l|c|c|c|c|c|c|c|}
 	
    \hline
    \% & 22.6 & 21.3 &  19.0 & 13.6&  12.8 &  6.7&4.1\\
    \hline
    W/H & 1.0 & 0.7 & 0.5 & 0.3 & 1.6 & 0.2 &  2.9 \\ 
   \hline 
   \tiny Max(W/H, H/W) & 1.0 & 1.4 & 2 & 3.3 & 1.6 &5& 2.9 \\
   \hline
	\end{tabular}
    \caption{Clustering results of aspect ratio of bounding boxes of training data}
    \label{tab:aspect_ratio}
\end{table}

\section{Experiments}
\label{sec:experiments}
\subsubsection*{Base network} Our experiments are all based on Residual-101~\cite{Residual}, which is pre-trained on the ILSVRC CLS-LOC dataset~\cite{Imagenet}. Following R-FCN~\cite{R-FCN}, we change the conv5 stage's effective stride from 32 pixels to 16 pixels to increase feature map resolution. The first convolution layer with stride 2 in the conv5 stage is modified to 1. Then following the \emph{\`{a} trous} algorithm~\cite{holschneider1990real}, for all convolution layers in conv5 stage with kernel size larger than 1, we increase their dilation from 1 to 2 to fix the "holes" caused by the reduced stride. Following SSD, and fitting the Residual architecture, we use Residual blocks to add a few extra layers with decreasing feature map size.

Table~\ref{tab:Selected} shows the selected feature layers in the original VGG architecture and in Residual-101. The depth is the position of the selected layer in the network. Only the convolution and the pooling layers are considered. It is important to note the depth of the first prediction layer in these two networks. Although Residual-101 contains 101 layers, we need to use dense feature layers to predict smaller objects and so we have no choice but to select the last feature layer in conv3\_x block. If we only consider the layers whose kernel size is larger than 1, this number will drop to 9. This means the receptive field of neurons in this layer may be smaller than the neurons in conv4\_3 in VGG. Compared to other layers in Residual-101, this layer gives worse prediction performance due to weak feature strength.

\begin{table}[t]
    \begin{center}
    \setlength{\tabcolsep}{1.5pt}
    \begin{tabular}{|l|c|c|c|c|c|c|c|}
    \hline

        \footnotesize  VGG & \footnotesize conv4\_3 & \footnotesize conv7 & \footnotesize conv8\_2 &  \footnotesize conv9\_2 & \footnotesize  conv10\_2 & \footnotesize  conv11\_2  \\
        \hline 
         \footnotesize \footnotesize  Resolution & \footnotesize$38\times 38$ &\footnotesize $19\times19$ & \footnotesize$10\times10$ & \footnotesize$5\times5$ & \footnotesize$3\times3$ & \footnotesize 1\\
        \footnotesize \footnotesize   Depth & 13 & 20 & 22 & 24 & 26 & 27\\
        
        \hline \hline
       \footnotesize  Residual-101 &\footnotesize  conv3\_x & \footnotesize conv5\_x & \footnotesize conv6\_x & \footnotesize  conv7\_x & \footnotesize conv8\_x & \footnotesize  conv9\_x  \\
       
       \hline
    \footnotesize    Resolution &  \footnotesize$40\times40$ &  \footnotesize$20\times20$  &  \footnotesize$10\times 10$ &  \footnotesize$5 \times 5$&  \footnotesize$3\times3$ &  \footnotesize1 \\
     \footnotesize   Depth & 23 & 101 & 104 & 107 & 110 & 113 \\
     \hline
 
	\end{tabular}
    \end{center}
    \caption{Selected feature layers in VGG and Residual-101 }
    \label{tab:Selected}
\end{table}

\begin{table*}[htbp]
	\centering
	\setlength{\tabcolsep}{1.53pt}
\begin{tabular*}{\textwidth}{l|c|c|cccccccccccccccccccc}
\tiny Method  & \tiny network & \tiny mAP & \tiny aero & \tiny bike & \tiny bird & \tiny boat & \tiny bottle & \tiny bus & \tiny car & \tiny cat & \tiny chair & \tiny cow & \tiny table & \tiny dog & \tiny horse & \tiny mbike & \tiny person & \tiny plant & \tiny sheep & \tiny sofa & \tiny train & \tiny tv \\
        \hline
       
     \tiny Faster \cite{faster-RCNN}  &
      \tiny VGG & 73.2 & 76.5 & 79.0 & 70.9 & 65.5 & 52.1 & 83.1 & 84.7 & 86.4 & 52.0 & 81.9 & 65.7 & 84.8 & 84.6 & 77.5 & 76.7 & 38.8 & 73.6 & 73.9 & 83.0 & 72.6\\
      
     \tiny ION \cite{insideout-net} &
     \tiny VGG &  75.6 & 79.2 & 83.1 & 77.6 & 65.6 & 54.9 & 85.4 & 85.1 & 87.0 & 54.4 & 80.6 &  73.8 & 85.3 & 82.2 & 82.2 & 74.4 & 47.1 & 75.8 & 72.7 & 84.2 & 80.4\\
 
     \tiny Faster~\cite{Residual} &
     \tiny Residual-101 & 76.4 & 79.8 & 80.7 & 76.2 & 68.3 & 55.9 & 85.1 & 85.3 & \textbf{89.8} & 56.7 & 87.8 & 69.4 & 88.3 & 88.9 & 80.9 & 78.4 &  41.7 & 78.6 & 79.8 & 85.3 & 72.0\\
     
     \tiny MR-CNN \cite{MR-CNN} & 
     \tiny VGG &  78.2 & 80.3 & 84.1 & 78.5 & 70.8 & 68.5 & 88.0 & 85.9 & 87.8 & 60.3 & 85.2 & 73.7 & 87.2 & 86.5 & 85.0 & 76.4 & 48.5 & 76.3 & 75.5 & 85.0 & 81.0 \\
     
     \tiny R-FCN~\cite{R-FCN} & 
     \tiny Residual-101 & 80.5 & 79.9 & \textbf{87.2} & 81.5 & 72.0 & \textbf{69.8} & 86.8 & 88.5 & 89.8 & \textbf{67.0} & \textbf{88.1} &  74.5 & \textbf{89.8} & \textbf{90.6} & 79.9 & 81.2 & \textbf{53.7} & 81.8 & 81.5 & 85.9 & 79.9\\
	  \hline
      
     \tiny SSD300*\cite{SSD} & 
     \tiny VGG & 77.5 &  79.5 & 83.9& 76.0& 69.6 &50.5 &87.0 &85.7& 88.1& 60.3& 81.5& 77.0& 86.1& 87.5& 83.97& 79.4& 52.3& 77.9& 79.5& 87.6& 76.8\\
  
     \tiny SSD 321 &
     \tiny Residual-101 & 77.1 & 76.3 & 84.6 & 79.3 & 64.6 &
     47.2 & 85.4 & 84.0 & 88.8 & 60.1 & 82.6 & 76.9 & 86.7 & 87.2 & 85.4 & 79.1 & 50.8 & 77.2 & \textbf{82.6} & \textbf{87.3} & 76.6\\

     \tiny DSSD 321 & 
     \tiny Residual-101 & 78.6 & 81.9 & 84.9 & 80.5 & 68.4 & 53.9 & 85.6 & 86.2 & 88.9 & 61.1 & 83.5 & 78.7 & 86.7 & 88.7 & 86.7 & 79.7 & 51.7 & 78.0 & 80.9 & 87.2 & 79.4  \\
	\hline
    
       \tiny SSD512*\cite{SSD} & 
     \tiny VGG & 79.5 & 84.8 & 85.1& 81.5 &73.0& 57.8& 87.8 &88.3 &87.4 &63.5 &85.4 &73.2& 86.2& 86.7 &83.9& 82.5 &55.6 &81.7 &79.0& 86.6 &80.0\\
    
    \tiny SSD 513 &
     \tiny Residual-101 & 80.6 & 84.3 & 87.6 & \textbf{82.6} & 71.6 & 59.0 & 88.2 & 88.1 & 89.3 & 64.4 & 85.6 & 76.2 & 88.5 & 88.9 & 87.5 & 83.0 & 53.6 & 83.9 & 82.2 & 87.2 & 81.3 \\
     
     \tiny DSSD 513 & 
     \tiny Residual-101 & \textbf{81.5} & \textbf{86.6} & 86.2 & \textbf{82.6} & \textbf{74.9} & 62.5 &\textbf{89.0} & \textbf{88.7} & 88.8 & 65.2 & 87.0 & \textbf{78.7} & 88.2 & 89.0 & \textbf{87.5} & \textbf{83.7} & 51.1 & \textbf{86.3} & 81.6 & 85.7 & \textbf{83.7} \\
	
		\end{tabular*}
		\caption{\textbf{PASCAL VOC2007 \texttt{test} detection results.} R-CNN series and R-FCN use input images whose minimum dimension is 600. The two SSD models have exactly the same settings except that they have different input sizes ($321\times 321$ vs. $513\times 513$).
        It order to fairly compare models, although Faster R-CNN with Residual network~\cite{Residual} and R-FCN~\cite{R-FCN} provide the number using multiple cropping or ensemble method in testing. We only list the number without these techniques. }
\label{tab:voc07}
\end{table*}

\subsection*{PASCAL VOC 2007}
We trained our model on the union of 2007  \texttt{trainval} and 2012 \texttt{trainval}. 

For the original SSD model, we used a batch size of 32 for the model with $321 \times 321$ inputs and 20 for the model with $513 \times 513$ inputs, and started the learning rate at $10^{-3}$  for the first 40k iterations. We then decreased it to $10^{-4}$ at 60K and $10^{-5}$ at 70k iterations. We take this well-trained SSD model as the pre-trained model for the DSSD. For the first stage, we only train the extra deconvolution side by freezing all the weights of original SSD model. We set the learning rate at $10^{-3}$ for the first 20k iterations, then continue training for 10k iterations with a $10^{-4}$ learning rate. For the second stage, we fine-tune the entire network with learning rate of $10^{-3}$ for the first 20k iteration and decrease it to $10^{-4}$ for next 20k iterations.

Table~\ref{tab:voc07} shows our results on the PASCAL VOC2007 \texttt{test} detection. SSD300* and SSD512* are the latest SSD results with the new expansion data augmentation trick, which are already better than many other state-of-the-art detectors. By replacing VGGNet to Residual-101, the performance is similar if the input image is small. For example, SSD321-Residual-101 is similar to SSD300*-VGGNet, although Residual-101 seems converge much faster (e.g. we only used half of the iterations of VGGNet to train our version of SSD). Interestingly, when we increase the input image size, Residual-101 is about 1\% better than VGGNet. We hypothesize that it is critical to have big input image size for Residual-101 because it is significant deeper than VGGNet so that objects can still have strong spatial information in some of the very deep layers (e.g. conv5\_x). More importantly, we see that by adding the deconvolution layers and skip connections, our DSSD321 and DSSD513 are consistently about 1-1.5\% better than the ones without these extra layers. This proves the effectiveness of our proposed method. Notably DSSD513 is much better than other methods which try to include context information such as MR-CNN~\cite{MR-CNN} and ION~\cite{insideout-net}, even though DSSD does not require any hand crafted context region information. Also, our single model accuracy is better than the current state-of-the-art detector R-FCN~\cite{R-FCN} by 1\%.

In conclusion, DSSD shows a large improvement for classes with specific backgrounds and small objects in both test tasks. For example the airplane, boat, cow, and sheep classes have very specific backgrounds. The sky for airplanes, grass for cow, etc. instances of bottle are usually small. This shows the weakness of small object detection in SSD is fixed by the proposed DSSD model, and better performance is achieved for classes with unique context.

\begin{table}[b]
    \setlength{\tabcolsep}{6pt}
    \begin{tabular}{|l|c|}
    \hline
    Method & mAP \\
    \hline
    SSD 321 & 76.4 \\
    \hline
    SSD 321 + PM(b) & 76.9\\ 
    SSD 321 + PM(c) &77.1 \\
    SSD 321 + PM(d) &77.0  \\
    \hline
    SSD 321 + PM(c) + DM(Eltw-sum) & 78.4\\
    SSD 321 + PM(c) + DM(Eltw-prod) & \textbf{78.6} \\
    \hline
    SSD 321 + PM(c) + DM (Eltw-prod) + Stage 2 & 77.9 \\
    \hline 
	\end{tabular}
    \caption{Ablation study : Effects of various prediction module and deconvolution module on PASCAL VOC 2007 \texttt{test }. \textbf{PM}: Prediction module in Figure\ref{fig:prediction_module}, \textbf{DM}:Feature Combination.}
    \label{tab:ablation}
\end{table}

\begin{table*}[ht]\small
\centering
	\setlength{\tabcolsep}{1.95pt}
\begin{tabular*}{\textwidth}{l|c|c|c|cccccccccccccccccccc}
\tiny Method & \tiny data  & \tiny network & \tiny mAP & \tiny aero & \tiny bike & \tiny bird & \tiny boat & \tiny bottle & \tiny bus & \tiny car & \tiny cat & \tiny chair & \tiny cow & \tiny table & \tiny dog & \tiny horse & \tiny mbike & \tiny person & \tiny plant & \tiny sheep & \tiny sofa & \tiny train & \tiny tv \\
        \hline
       
       \tiny ION~\cite{insideout-net} & \tiny 07+12+S & \tiny VGG & 76.4 & 87.5 & 84.7 & 76.8 & 63.8 & 58.3 & 82.6 & 79.0 & 90.9 & 57.8 & 82.0 & 64.7 & 88.9 & 86.5 & 84.7 & 82.3 & 51.4 & 78.2 & 69.2 & 85.2 & 73.5 \\
       
       \tiny Faster~\cite{Residual} & \tiny 07++12 & \tiny Residual-101 & 73.8 & 86.5 & 81.6 & 77.2 & 58.0 & 51.0 & 78.6 & 76.6 & 93.2 & 48.6 & 80.4 & 59.0 & 92.1 & 85.3 & 84.8 & 80.7 & 48.1 & 77.3 & 66.5 & 84.7 & 65.6 \\

       \tiny R-FCN~\cite{R-FCN} & \tiny 07++12& \tiny Residual-101 & 77.6 & 86.9 & 83.4 & \textbf{81.5}& 63.8& \textbf{62.4} & 81.6 & 81.1 & 93.1 & 58.0 & 83.8 & 60.8& 92.7 & 86.0 & 84.6 & 84.4 & \textbf{59.0} & 80.8 & 68.6& 86.1 & 72.9 \\
       \hline
      
      
      \tiny SSD300*\cite{SSD} & \tiny 07++12&
      \tiny VGG & 75.8 & 88.1 & 82.9 & 74.4 & 61.9 & 47.6 & 82.7 & 78.8 & 91.5 & 58.1 & 80.0 & 64.1 & 89.4 & 85.7 & 85.5 & 82.6 & 50.2 & 79.8 & 73.6 & 86.6 & 72.1\\
 	

	\tiny SSD321 & \tiny 07++12 & \tiny Residual-101 & 75.4 & 87.9& 82.9&  73.7 & 61.5 & 45.3 & 81.4 & 75.6 & 92.6 & 57.4 & 78.3 & 65.0 & 90.8 & 86.8 & 85.8 & 81.5 & 50.3 &78.1 &75.3 &85.2 & 72.5 \\

     
     \tiny DSSD 321 & \tiny 07++12 & 
     \tiny Residual-101 & 76.3 & 87.3 & 83.3 & 
75.4 & 64.6 & 46.8 & 82.7 & 76.5 & 92.9 & 59.5 & 78.3 & 64.3 &  91.5 & 86.6 & 86.6 & 82.1 & 53.3 & 79.6 & 75.7 & 85.2 & 73.9 \\
        	 \hline
    
     
  	 \tiny SSD512*\cite{SSD} & \tiny 07++12&
     \tiny VGG & 78.5 & 90.0 & 85.3 & 77.7 & 64.3 & 58.5 & \textbf{85.1} & 84.3 & 92.6 & 61.3 & 83.4 & 65.1 & 89.9 & 88.5 & \textbf{88.2} & 85.5 & 54.4 & 82.4 & 70.7 & 87.1 & 75.6\\
     
     
     \tiny SSD 513 & \tiny 07++12 & 
     \tiny Residual-101 & 79.4 & 90.7 & \textbf{87.3} & 78.3 & 66.3 & 56.5 & 84.1 & 83.7 &  94.2 & 62.9 & 84.5 & \textbf{66.3} & 92.9 & \textbf{88.6} & 87.9 & 85.7 & 55.1 & 83.6  & \textbf{74.3} & \textbf{88.2} & \textbf{76.8} \\
     
     
     \tiny DSSD 513 & \tiny 07++12& 
     \tiny Residual-101 & \textbf{80.0} & \textbf{92.1} & 86.6 & 80.3 & 68.7 & 58.2 & 84.3 & \textbf{85.0} & \textbf{94.6} & \textbf{63.3} & \textbf{85.9} & 65.6 & \textbf{93.0} & 88.5 & 87.8 & \textbf{86.4} & 57.4 & \textbf{85.2} & 73.4 & 87.8 & \textbf{76.8}  \\

	\end{tabular*}
	\caption{\textbf{PASCAL 2012 \texttt{test} detection results.}
\textbf{07+12}: 07 \texttt{trainval} + 12 \texttt{trainval}, \textbf{07+12+S}: 07+12 plus segmentation labels,
\textbf{07++12}: 07 \texttt{trainval} + 07 \texttt{test} + 12 \texttt{trainval}
\label{tab:voc12}
}
\end{table*}

\subsubsection*{Ablation Study on VOC2007}
In order to understand the effectiveness of our additions to SSD, we run models with different settings on VOC2007 and record their evaluations in Table~\ref{tab:ablation}. PM = Prediction module, and DM = Deconvolution module. 
The pure SSD using Residual-101 with 321 inputs is 76.4\% mAP. This number is actually worse than the VGG model. By adding the prediction module we can see the result is improving, and the best is when we use one residual block as the intermediate layer before prediction. The idea is to avoid allowing the gradients of the objective function to directly flow into the backbone of the Residual network. We do not see much difference if we stack two PM before prediction.

When adding the deconvolution module (DM), Elementwise-product shows the best (78.6\%) among all the methods. The results are similar in~\cite{Fukui2016bp_vqa} which evaluate different methods combining vision and text features. We also try to use the approximate bilinear pooling method\cite{compact_bilinear_pooling}, the low-dimensional approximation of the original method proposed by Lin \etal ~\cite{lin2015bilinear}, but training speed is slowing down and the training error decrease very slowly as well. Therefore, we did not use or evaluate it here. The better feature combination can be considered as further work to improve the accuracy of the DSSD model.

We also tried to fine-tune the whole network after adding and fine-tuning the DM component, however we did not see any improvements but instead decreased performance.

\subsection*{PASCAL VOC 2012}

For VOC2012 task, we follow the setting of VOC2007 and with a few differences described here. We use 07++12 consisting of VOC2007 \texttt{trainval}, VOC2007 \texttt{test}, and VOC2012 \texttt{trainval} for training and VOC2012 \texttt{test} for testing. Due to more training data, increasing the number of training iterations is needed. For the SSD model, we train the first 60k iterations with $10^{-3}$ learning rate, then 30k iterations with $10^{-4}$ learning rate, and use $10^{-5}$ learning rate for the last 10k iterations. For the DSSD, we use the well-trained SSD as the pre-trained model. According to the Table \ref{tab:ablation} ablation study, we only need to train the Stage 1 model. Freezing all the weights of the original SSD model, we train the deconvolution side with learning rate at $10^{-3}$ for the first 30k iterations, then $10^{-4}$ for the next 20k iterations. 
The results, shown in Table ~\ref{tab:voc12}, once again validates that DSSD outperforms all others. It should be noted that our model is the only model that achieves 80.0\% mAP without using extra training data (i.e. COCO), multiple cropping, or an ensemble method in testing.

\begin{table*}[tpb]
	\centering
	\setlength{\tabcolsep}{2.6pt}
	\begin{tabular*}{\textwidth}{l|c|c|C{3.4em}C{2.2em}C{2.2em}|C{2em}C{2em}C{2em}|C{2.2em}C{2.2em}C{2.2em}|C{2em}C{2em}C{2em}}
     
    	\multirow{2}{*}{Method} & \multirow{2}{*}{data} &     
\multirow{2}{*}{network} &        
\multicolumn{3}{c|}{\scriptsize{Avg. Precision, IoU:}} & \multicolumn{3}{c|}{\scriptsize{Avg. Precision, Area:}} & \multicolumn{3}{c|}{\scriptsize{Avg. Recall, \#Dets:}} & \multicolumn{3}{c}{\scriptsize{Avg. Recall, Area:}}\\
        & & &0.5:0.95 & 0.5 & 0.75 & S & M & L & 1 & 10 & 100 & S & M & L\\
        \hline
     
        Faster~\cite{faster-RCNN} & trainval & VGG & 21.9 & 42.7 & - & - & - & - & - & - & - & - & - & -\\
        ION~\cite{insideout-net} & train & VGG & 23.6 & 43.2 & 23.6 & 6.4 & 24.1 & 38.3 & 23.2 & 32.7 & 33.5 & 10.1 & 37.7 & 53.6\\
        
        Faster+++~\cite{Residual} & trainval & Residual-101 & 34.9 & 55.7 & - & - & - & - & - & - & - & - & - & \\

          R-FCN~\cite{R-FCN} & trainval & Residual-101 & 29.9&51.9  & - & 10.8 & 32.8 & 45.0 & - & - &-&-&-&-  \\
        \hline
        SSD300*~\cite{SSD} & trainval35k & VGG & 25.1 & 43.1 & 25.8 & 6.6 & 25.9 & 41.4 & 23.7 & 35.1 & 37.2 & 11.2 & 40.4 & 58.4\\
        SSD321 & trainval35k & Residual-101 & 28.0 & 45.4 & 29.3 & 6.2 & 28.3 & 49.3  & 25.9 & 37.8 & 39.9 & 11.5 & 43.3 & 64.9  \\
        DSSD321 & trainval35k & Residual-101 & 
        28.0 & 46.1 & 29.2 & 7.4 & 28.1 & 47.6 &  25.5 & 37.1 & 39.4& 12.7 & 42.0 & 62.6\\
        
        \hline 
        SSD512*~\cite{SSD} & trainval35k & VGG & 28.8 & 48.5 & 30.3 & 10.9 & 31.8 & 43.5 & 26.1 & 39.5 & 42.0 & 16.5 & 46.6 & 60.8\\
        
        SSD513 & trainval35k & Residual-101 & 31.2 & 50.4 & 33.3 & 10.2 & 34.5 & 49.8 & 28.3 & 42.1 & 44.4 & 17.6 & 49.2 & 65.8 \\
        
        DSSD513 & trainval35k & Residual-101 & 33.2 & 53.3 & 35.2& 13.0 & 35.4 & 51.1 & 28.9 & 43.5 & 46.2 & 21.8 & 49.1 & 66.4 \\ 
    \end{tabular*}
    \caption{\textbf{COCO \texttt{test-dev2015} detection results.}}
    \label{tab:coco}
\end{table*}

\subsection*{COCO}
Because Residule-101 uses batch normalization, to get more stable results, we set the batch size to 48 (which is the largest batch size we can use for training on a machine with 4 P40 GPUs) for training SSD321 and 20 for training SSD513 on COCO. We use a learning rate of $10^{-3}$ for the first 160k iteration, then $10^{-4}$ for 60k iteration and $10^{-5}$ for the last 20k. According to our observation, a batch size smaller than 16 and trained on 4 GPUs can cause unstable results in batch normalization and hurt accuracy.

We then take this well-trained SSD model as the pre-trained model for the DSSD. For the first stage, we only train the extra deconvolution side by freezing all the weights of original SSD model. We set the learning rate at  $10^{-3}$ for the first 80k iterations, then continue training for 50k iterations with a $10^{-4}$ learning rate. We didn't run the second stage training here, based on the Table ~\ref{tab:ablation} results.

From Table~\ref{tab:coco}, we see that SSD300* is already better than Faster R-CNN~\cite{faster-RCNN} and ION~\cite{insideout-net} even with a very small input image size ($300\times 300$). By replacing VGGNet with Residual-101, we saw a big improvement (28.0\% vs. 25.1\%) with similar input image size (321 vs. 300). Interestingly, SSD321-Residual-101 is about 3.5\% better (29.3\% vs. 25.8\%) at higher Jaccard overlap threshold (0.75) while is only 2.3\% better at 0.5 threshold. We also observed that it is 7.9\% better for large objects and has no improvement on small objects. We think this demonstrates that Residual-101 has much better features than VGGNet which contributes to the great improvement on large objects. By adding the deconvolution layers on top of SSD321-Residual-101, we can see that it performs better on small objects (7.4\% vs. 6.2\%), unfortunately we don't see improvements on large objects. 

For bigger model, SSD513-Residual-101 is already 1.3\%(31.2\% vs. 29.9\%) better than the state-of-the-art method R-FCN ~\cite{R-FCN}. Switching to Residual-101 gives a boost mainly in large and medium objects. The DSSD513-Residual-101 shows improvement on all sizes of objects and achieves 33.2 \% mAP which is 3.3\% better than R-FCN~\cite{R-FCN}. According to this observation, we speculation that DSSD will benefit more when increasing the input image size, albeit with much longer training and inference time.

\begin{table*}[ht]
\begin{center}
	\centering
    \setlength{\tabcolsep}{3pt}
    \begin{tabular*}{\textwidth}
    {l|C{6em}|C{3.3em}|C{3em}|C{3em}|C{3em}|C{3em}|c|C{3.6em}|c}

	\multirow{2}{*}{Method} & \multirow{2}{*}{network} & \multirow{2}{*}{mAP} & \multicolumn{2}{c|}{\footnotesize{With BN layers} }& \multicolumn{2}{c|}{\footnotesize{BN layers removed}} & \multirow{2}{*}{\# Proposals} & \multirow{2}{*}{GPU} & \multirow{2}{*}{Input resolution}        \\
    
    				\hhline{~~~----~~~}
 	&&&FPS&\small{\makecell{batch \\ size}} &FPS &\small{\makecell{batch \\ size}}&&&
 					\\
                    \hline

        Faster R-CNN \cite{faster-RCNN} &VGG16 &73.2 & 7 &  1&- &- & 6000& Titan X &  $\sim1000\times600$\\
        Faster R-CNN \cite{Residual} & Residual-101 & 76.4 & 2.4&1&-&-& 300 & K40 &   $\sim1000\times600$\\
		R-FCN \cite{R-FCN}& Residual-101 &   80.5 & 9 &1 &-&-& 300& Titan X & $\sim1000\times600$\\

        \hline
        SSD300*\cite{SSD} & VGG16 & 77.5 & 46& 1 & - & -  &8732 & Titan X & $300\times300$ \\
        SSD512*\cite{SSD} & VGG16 & 79.5 & 19& 1 & -& -  &24564 & Titan X & $512\times512$ \\

        \hline 
        SSD321 & Residual-101 & 77.1 & 11.2 &  1 &  16.4& 1&17080&Titan X & $321\times321$ \\
        SSD321 & Residual-101 & 77.1 & 18.9 & 15 & 22.1& 44&17080&Titan X & $321\times321$ \\
        DSSD321 & Residual-101 &78.6 &9.5 &  1 &11.8 & 1 & 17080 & Titan X & $321\times321$ \\
        DSSD321 & Residual-101 &78.6 &13.6  & 12& 15.3 & 36 & 17080 & Titan X & $321\times321$ \\
   
        \hline
        SSD513 & Residual-101 & 80.6 & 6.8 &  1 & 8.0&  1 & 43688& Titan X &  $513\times513$ \\ 
        SSD513 & Residual-101 & 80.6 & 8.7 &  5 & 11.0& 16 & 43688& Titan X &  $513\times513$ \\ 
        DSSD513 & Residual-101 & 81.5 & 5.5 & 1 &6.4 & 1 & 43688  & Titan X &  $513\times513$ \\
         DSSD513 & Residual-101 & 81.5 & 6.6 &  4 &6.3 & 12 & 43688  & Titan X &  $513\times513$ \\

	\end{tabular*}
    \end{center}
    \caption{\textbf{Comparison of Speed \& Accuracy on PASCAL VOC2007 \texttt{test}.} }
    
    \label{tab:inference}
\end{table*}

\subsection*{Inference Time}
In order to speed up inference time, we use the following equations to remove the batch normalization layer in our network at test time. In Eq. \ref{eq:batchnorm_x}, the output of a convolution layer will be normalized by subtracting the mean,  dividing the square root of the variance plus $\epsilon$ ($\epsilon = 10^{-5}$), then scaling and shifting by parameters which were learned during training. To simplify and speed up the model during testing, we can rewrite the weight (Eq.~\ref{eq:batchnorm_w}) and bias (Eq.~\ref{eq:batchnorm_b}) of a convolution layer and remove the batch normalization related variables as shown in Eq. \ref{eq:batchnorm_final}. We found that this trick improves the speed by $1.2\times$ - $1.5\times$ in general and reduces the memory footprint up to three times.

\begin{align}
y &= scale\left( \frac{(wx + b) - \mu}{\sqrt[]{var + \epsilon}} \right) + shift 
\label{eq:batchnorm_x}\\
\hat{w} &= scale \left( \frac{w}{\sqrt[]{var + \epsilon}} \right)
\label{eq:batchnorm_w}\\
\hat{b} &= scale\left( \frac{b - \mu}{\sqrt[]{var + \epsilon}} \right) + shift 
\label{eq:batchnorm_b} \\
y &= \hat{w}x + \hat{b}
\label{eq:batchnorm_final}
\end{align}

 The model we proposed is not as fast as the original SSD for multiple reasons. First, the Residual-101 network, which has much more layers, is slower than the (reduced) VGGNet. Second, the extra layers we added to the model, especially the prediction module and the deconvolutional module, introduce extra overhead. A potential way to speed up DSSD is to replace the deconvolution layer by a simple bilinear up-sampling. Third, we use much more default boxes. Table \ref{tab:inference} shows that we use 2.6 times more default boxes than a previous version of SSD (43688 vs. 17080). These extra default boxes take more time not only in prediction but also in following non-maximum suppression.

We test our model using either a batch size of 1, or the maximum size that can fit in the memory of a Titan X GPU.
 We compiled the results in Table \ref{tab:inference} using a Titan X GPU and cuDNN v4 and an Intel Xeon E5-2667v3@3.20GHz. Compared to R-FCN \cite{R-FCN}, the SSD 513 model has similar speed and accuracy. Our DSSD 513 model has better accuracy, but is slightly slower. The DSSD 321 model maintains a speed advantage over R-FCN, but with a small drop in accuracy. Our proposed DSSD model achieves state-of-the-art accuracy while maintaining a reasonable speed compared to other detectors.

\subsection*{Visualization}
In Figure \ref{fig:coco} , we show some detection examples on
COCO test-dev with the SSD321 and DSSD321 models. Compared to SSD, our DSSD model improves in two cases. The first case is in scenes containing small objects or dense objects as shown in Figure \ref{sfig:dense}. Due to the small input size, SSD does not work well on small objects, but DSSD shows obvious improvement. The second case is for  certain classes that have distinct context. In Figure ~\ref{sfig:context}, we can see the results of classes with specific relationships can be improved: tie and man in suit, baseball bat and baseball player, soccer ball and soccer player, tennis racket and tennis player, and skateboard and jumping person. 

\section{Conclusion}
	We propose an approach for adding context to a state-of-the-art object detection framework, and demonstrate its effectiveness on benchmark datasets. While we expect many improvements in finding more efficient and effective ways to combine the features from the encoder and decoder, our model still achieves state-of-the-art detection results on PASCAL VOC and COCO. Our new DSSD model is able to outperform the previous SSD framework, especially on small object or context specific objects, while still preserving comparable speed to other detectors. While we only apply our encoder-decoder hourglass model to the SSD framework, this approach can be applied to other detection methods, such as the R-CNN series methods~\cite{R-CNN,fast-RCNN,faster-RCNN}, as well. 

\section{Acknowledgment}
This project was started as an intern project at Amazon Lab126 and continued at UNC. We would like to give special thanks to Phil Ammirato for helpful discussion and comments. We also thank NVIDIA for providing great help with GPU clusters and acknowledge support from NSF 1533771.

{\footnotesize
\bibliographystyle{ieee}
\bibliography{egbib}
}

\clearpage

\begin{figure*}
	\centering
	
    \begin{subfigure}{\textwidth}
    \includegraphics[width=0.24\linewidth]{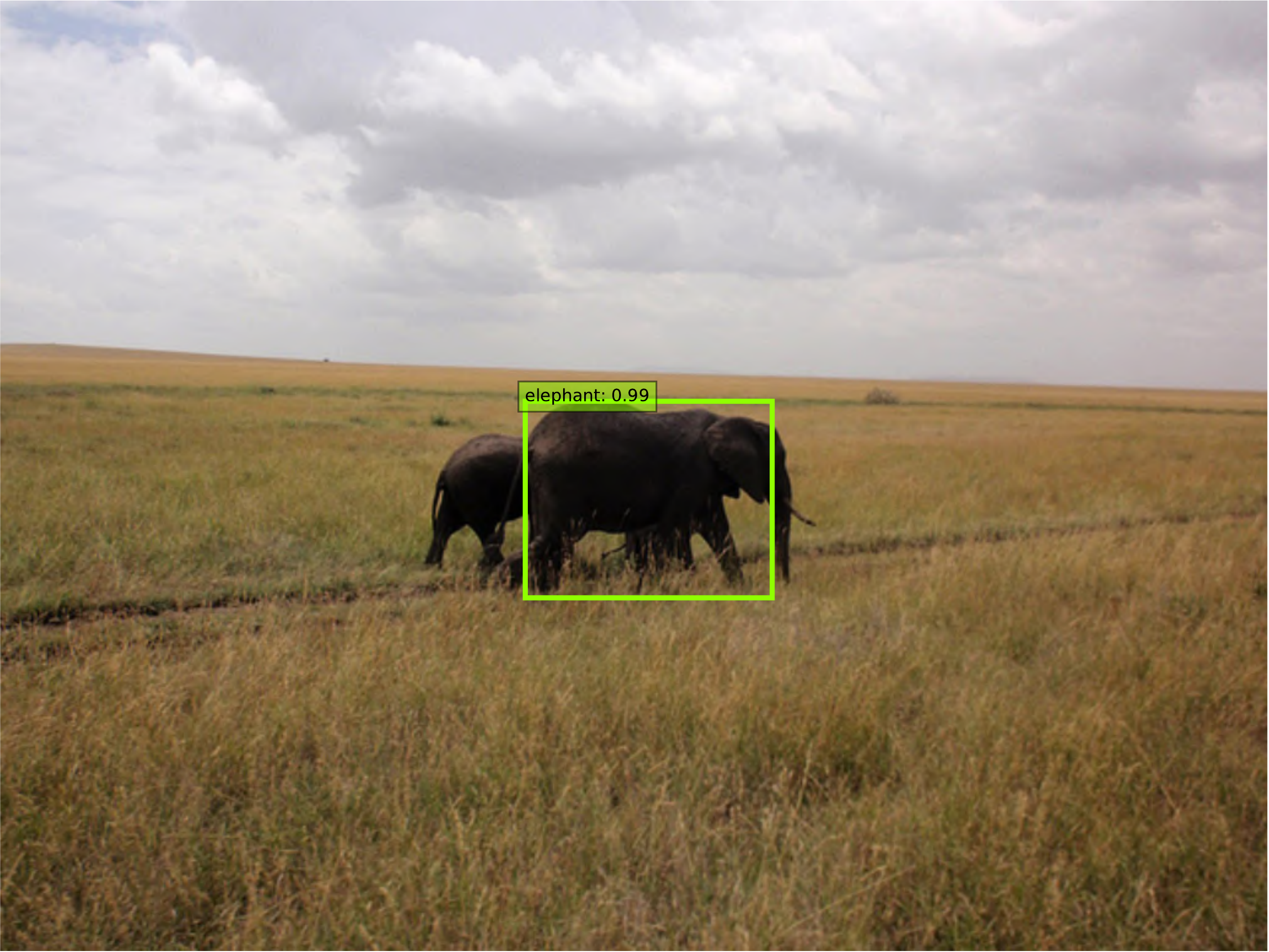}
    \includegraphics[width=0.24\linewidth]{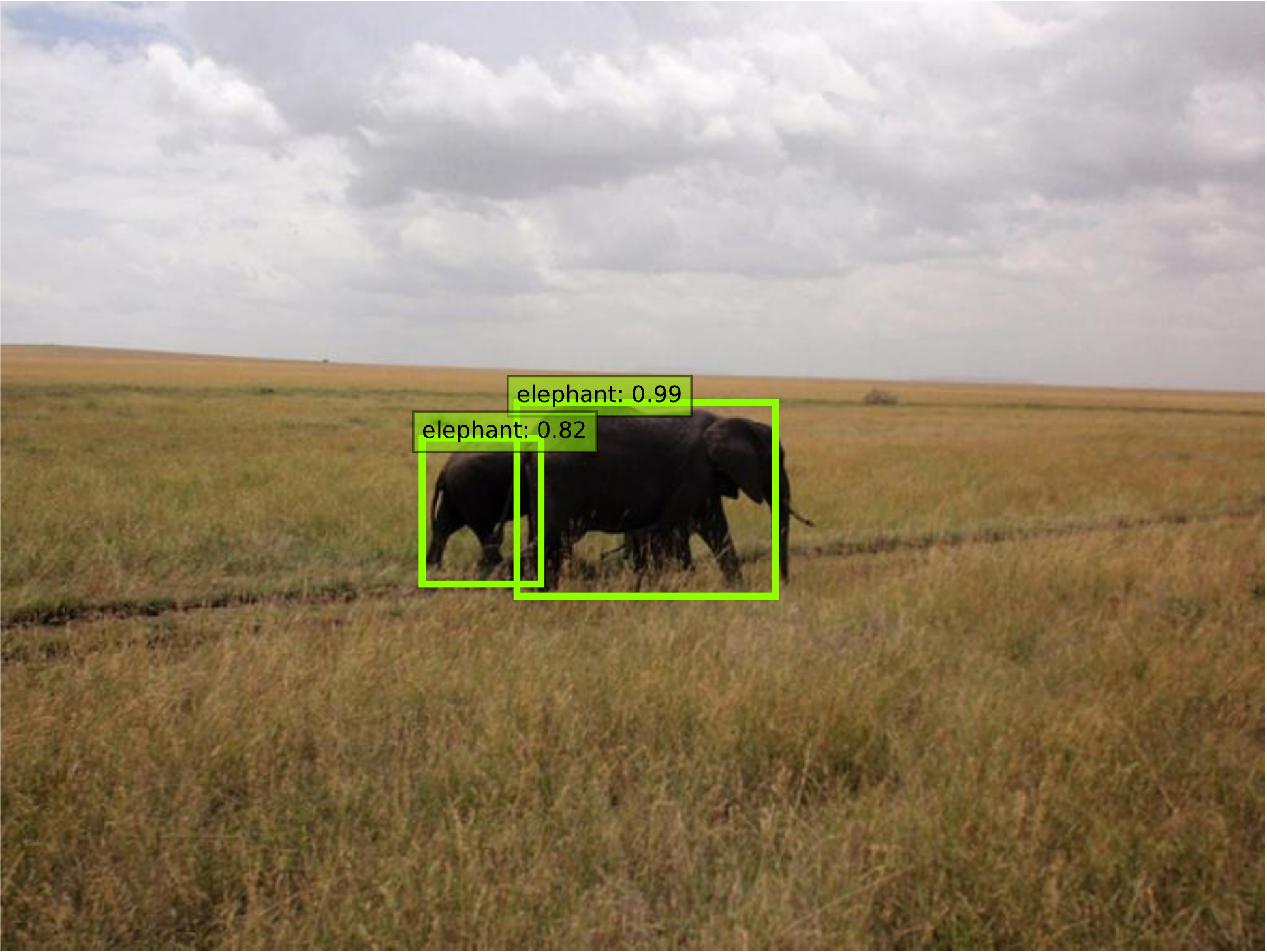}
	\includegraphics[width=0.24\linewidth]{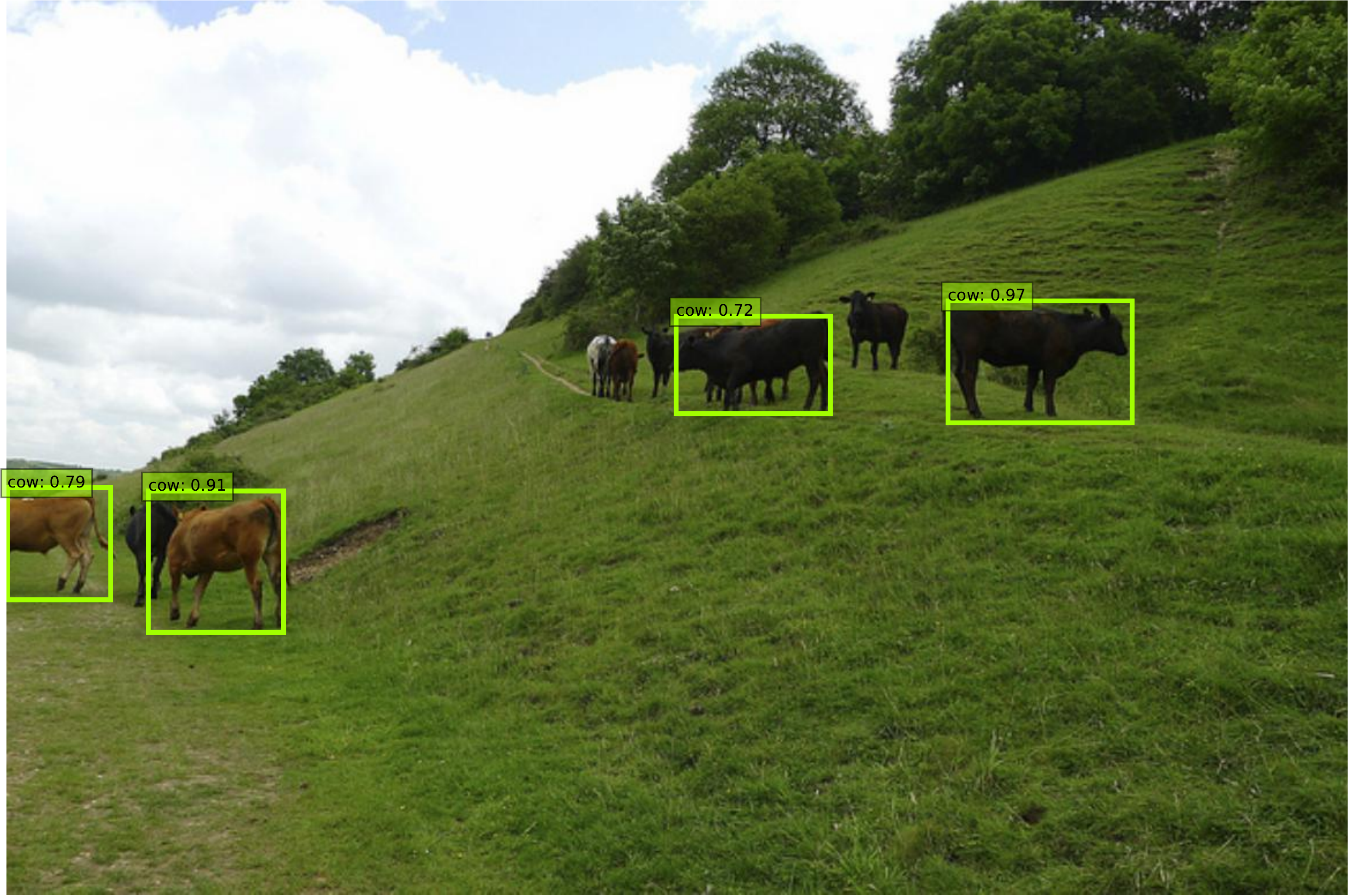}
    \includegraphics[width=0.24\linewidth]{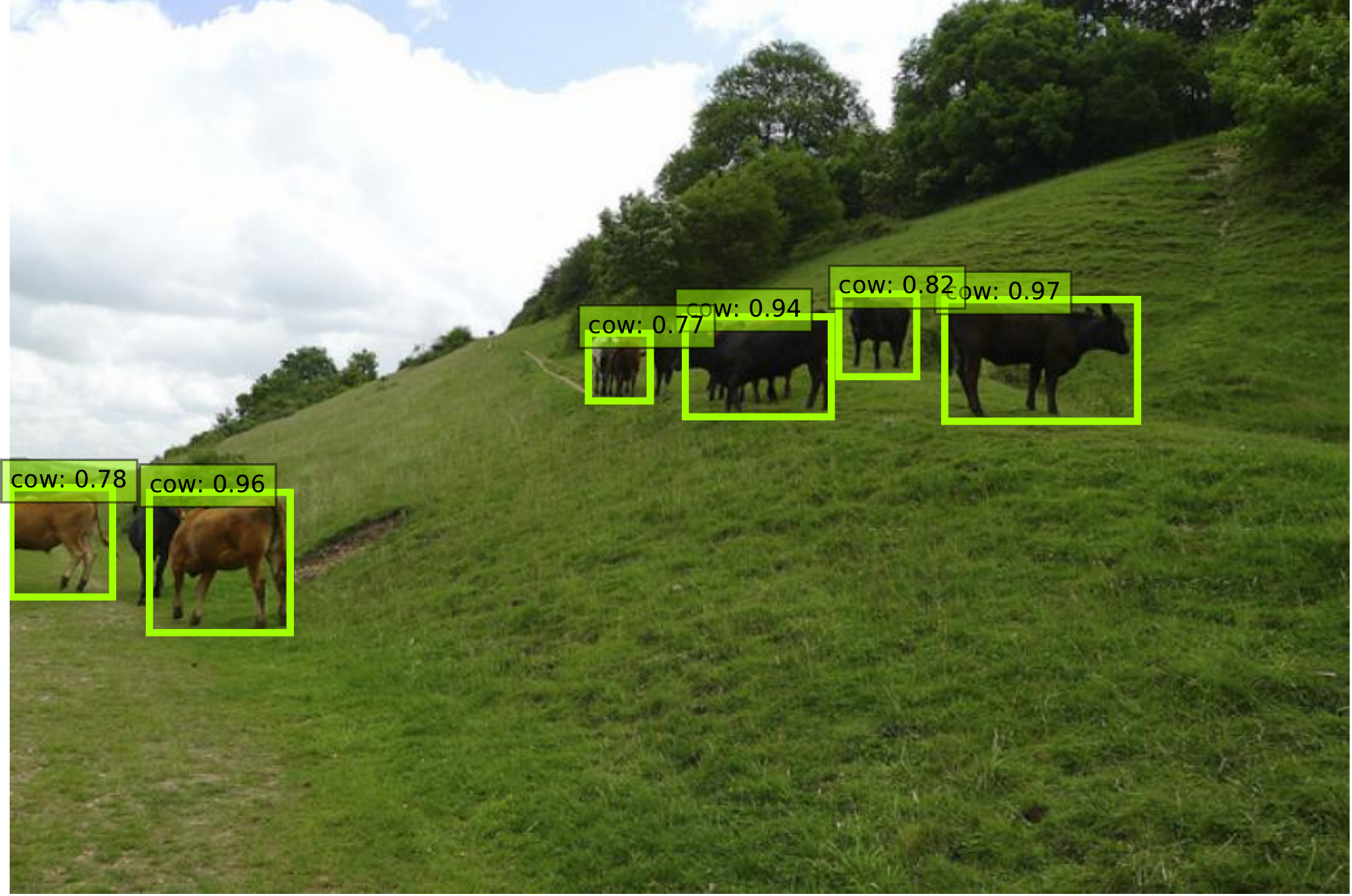}
\\
	\includegraphics[width=0.24\linewidth]{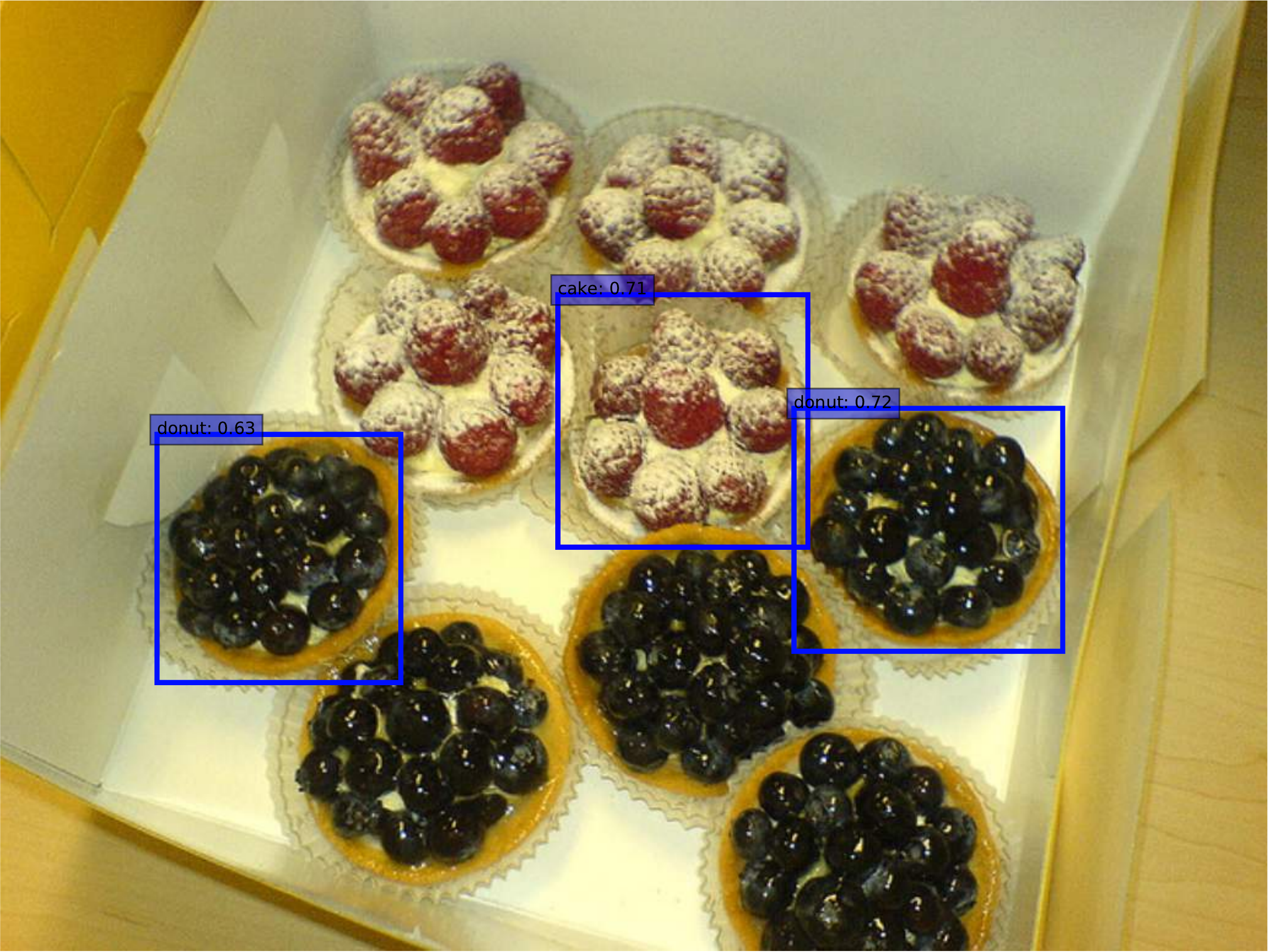}
    \includegraphics[width=0.24\linewidth]{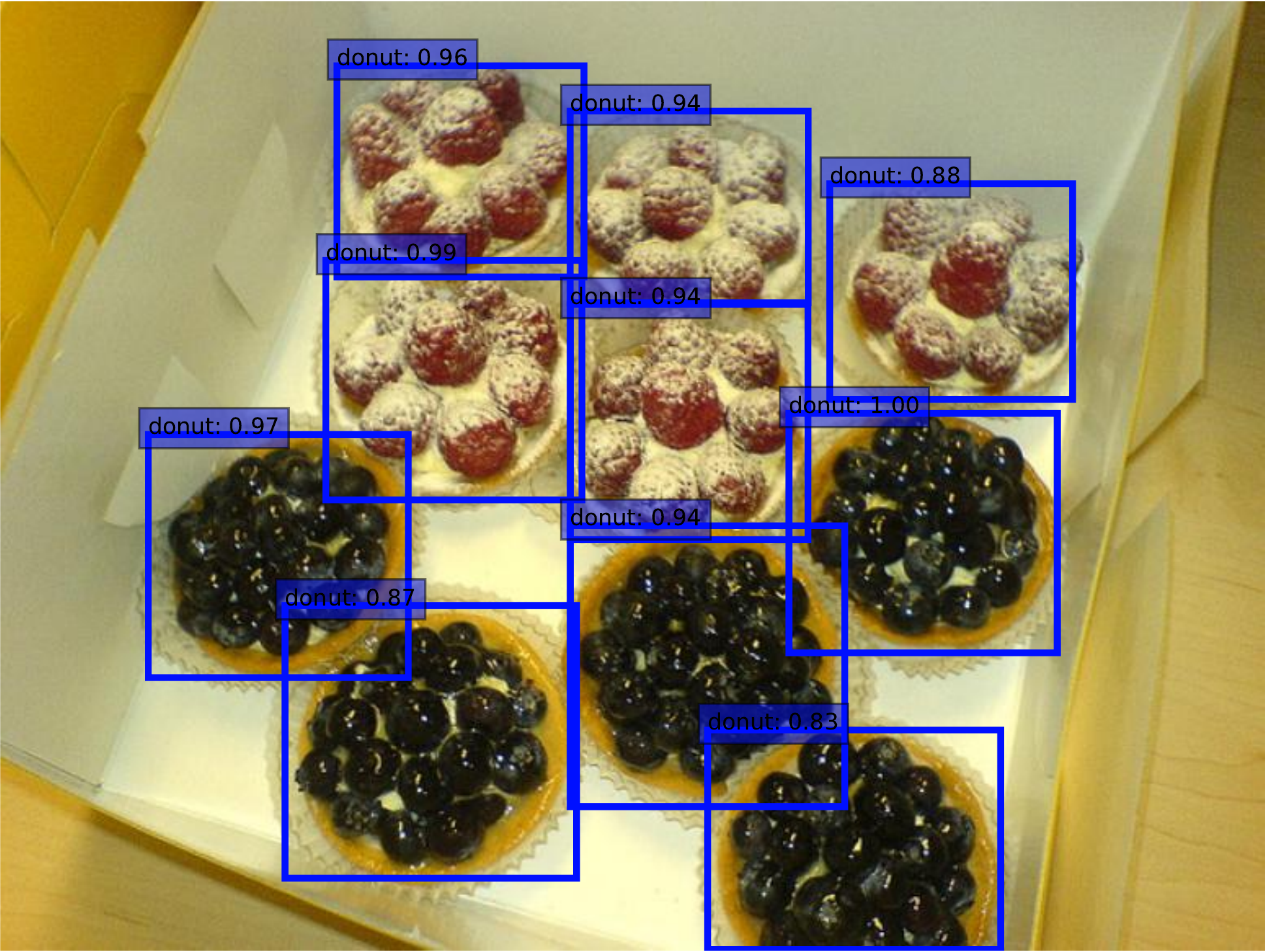}
    \includegraphics[width=0.24\linewidth]{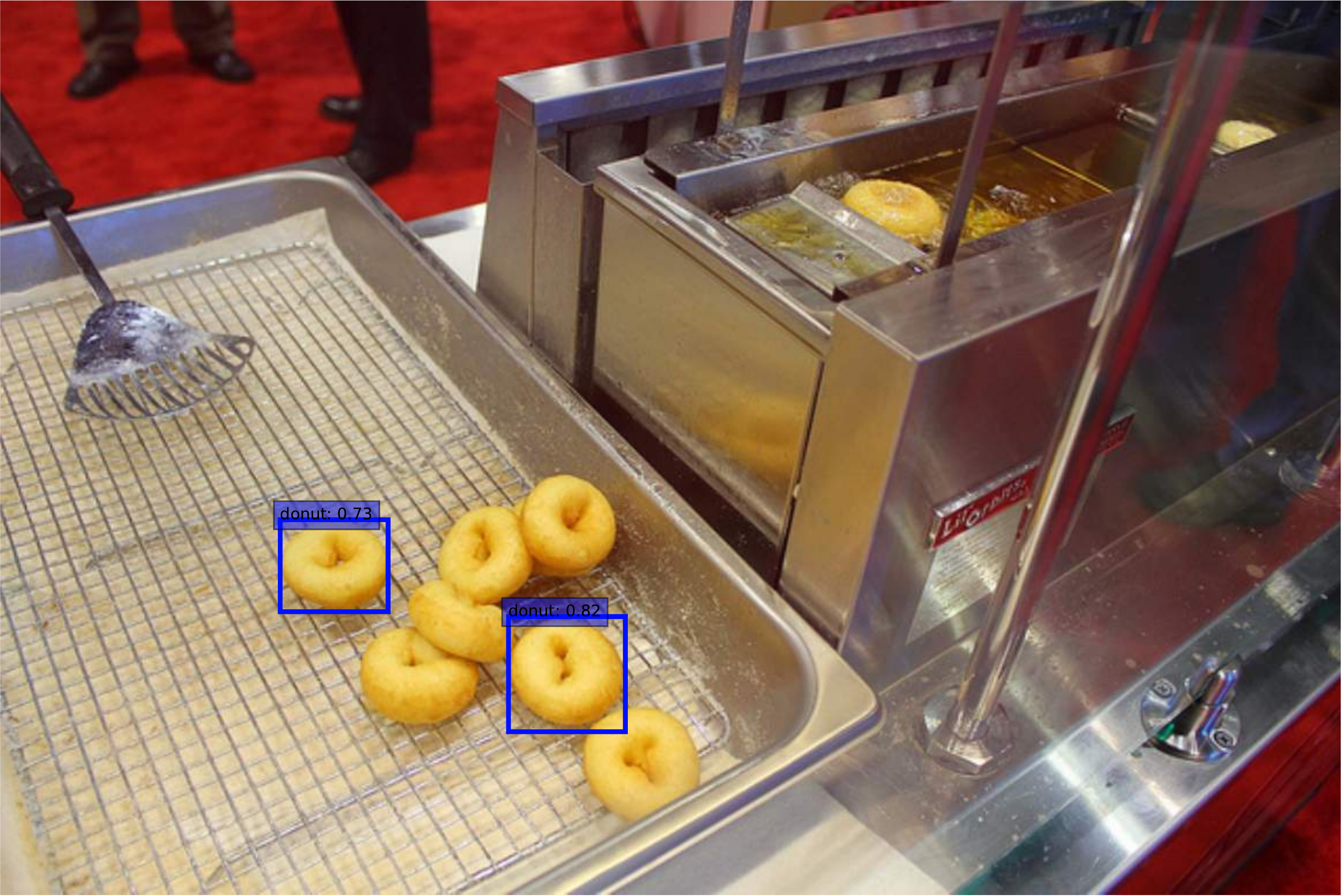}
    \includegraphics[width=0.24\linewidth]{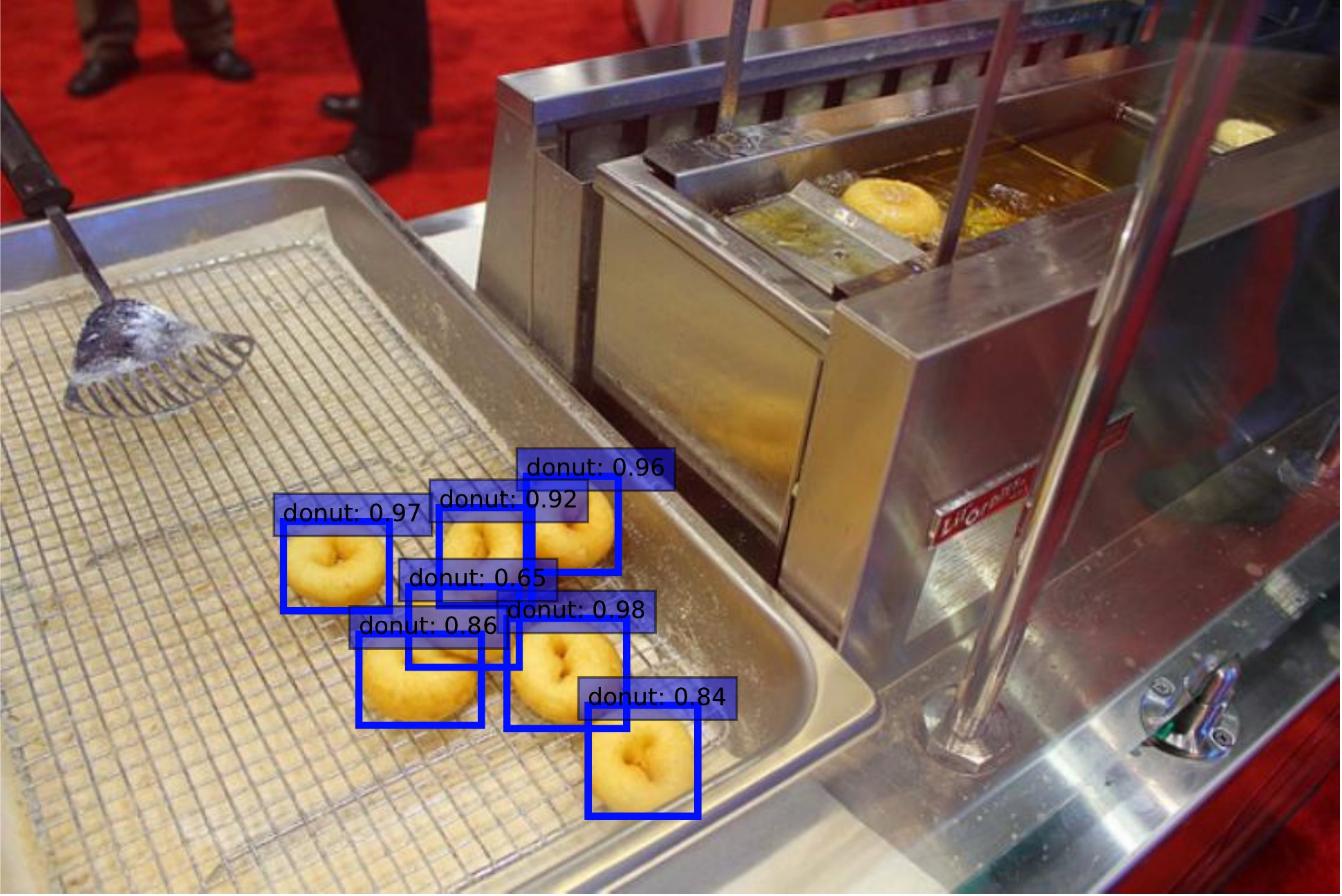}
 \\
 	\includegraphics[width=0.24\linewidth]{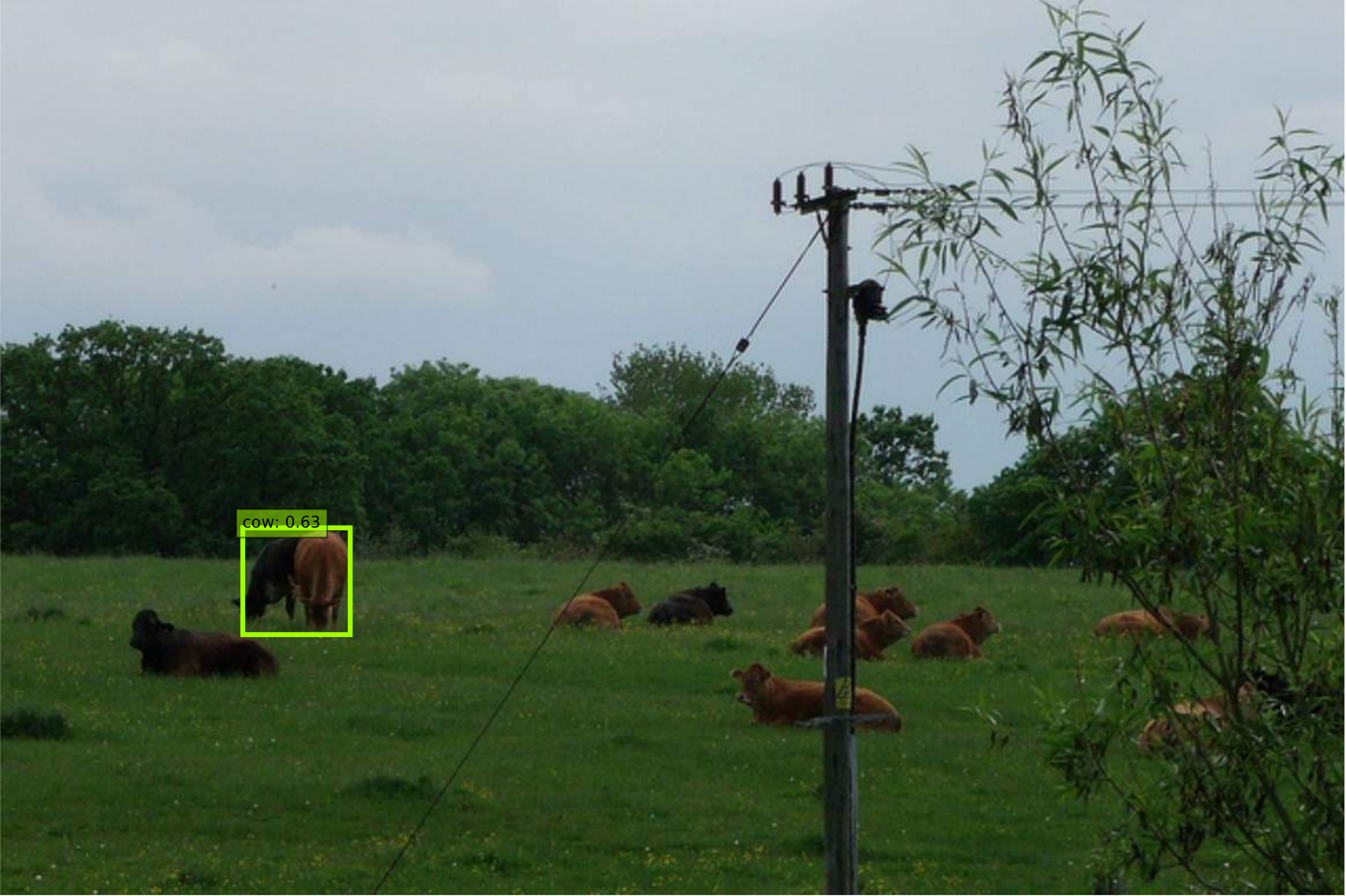}
    \includegraphics[width=0.24\linewidth]{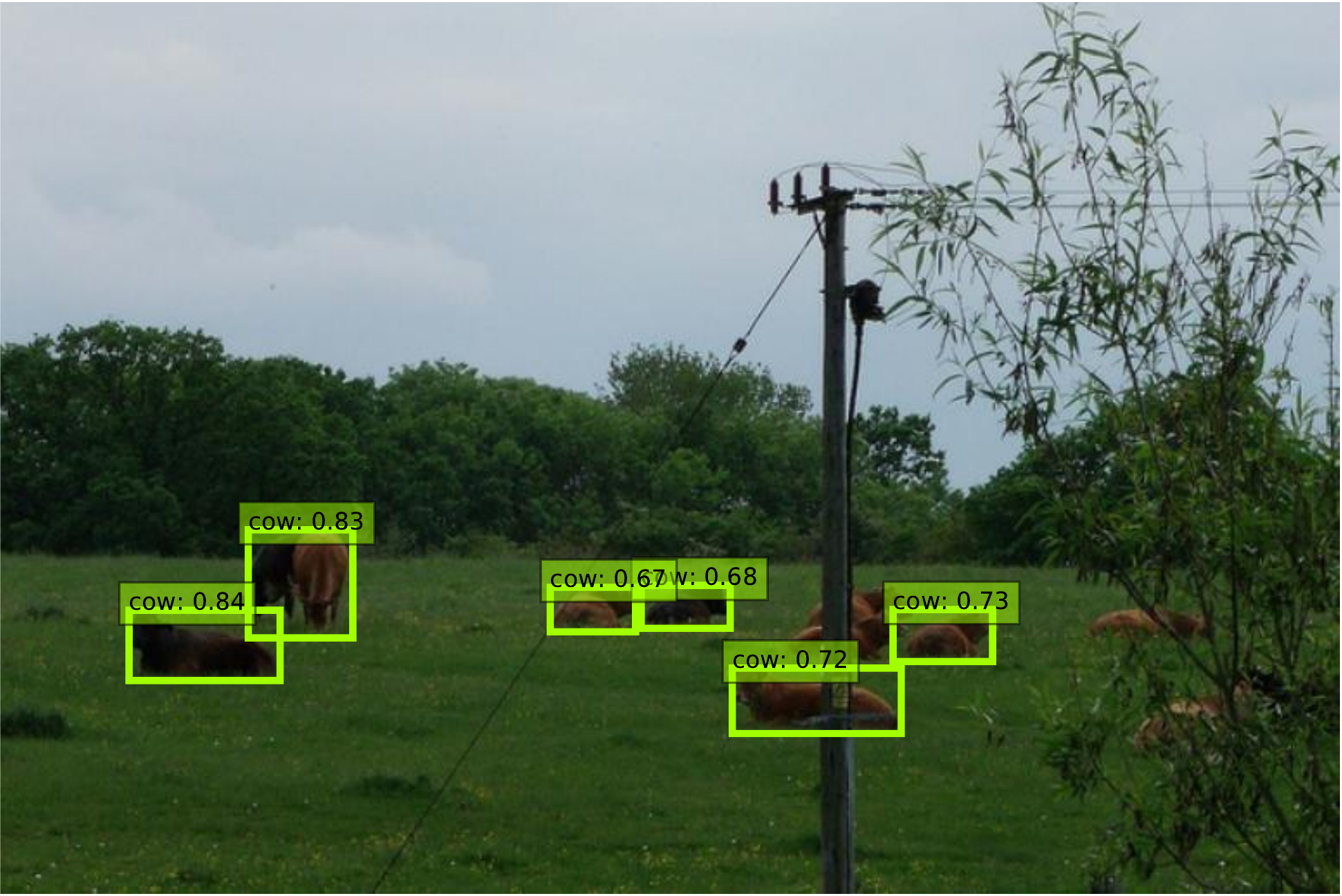}
    \includegraphics[width=0.24\linewidth]{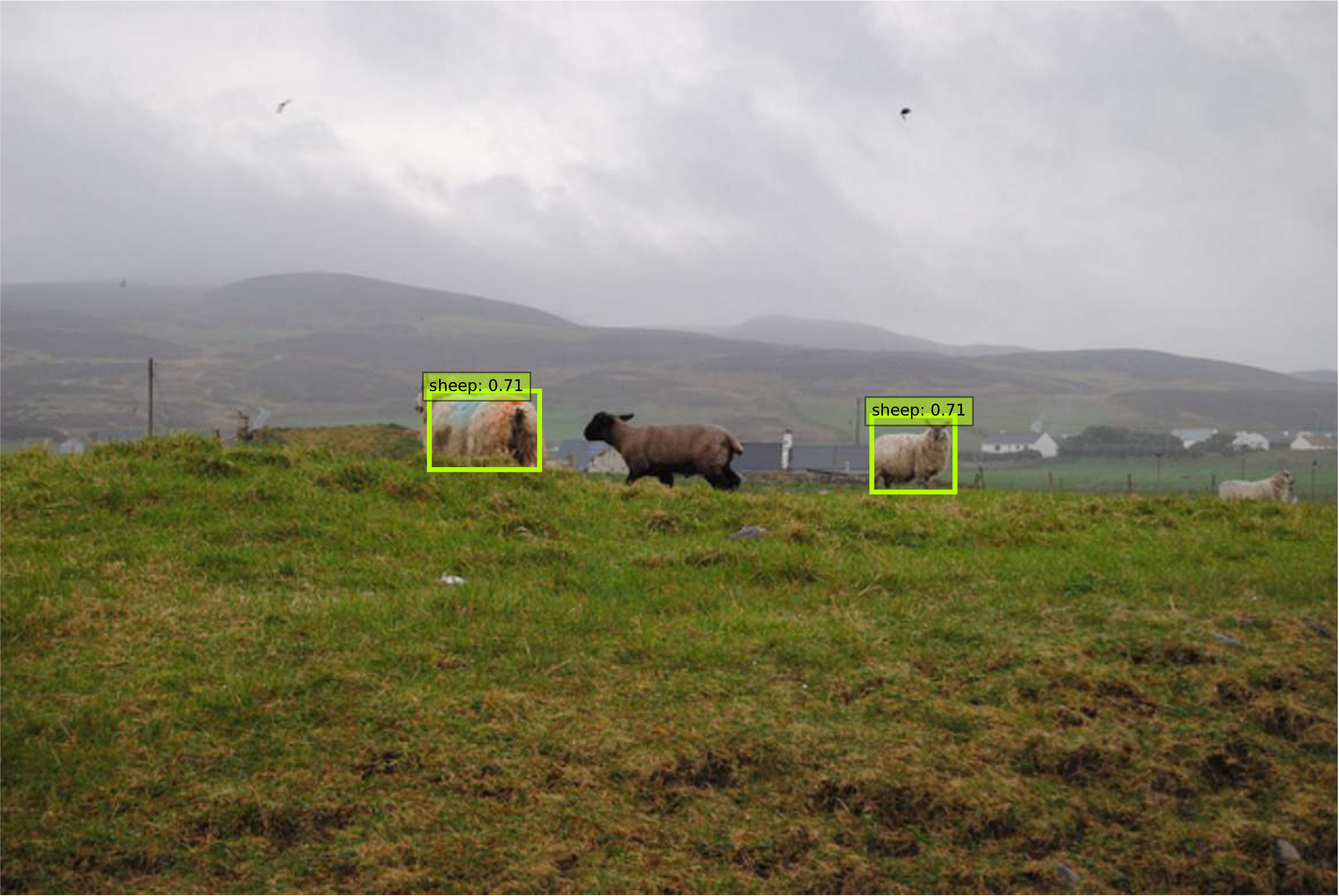}
    \includegraphics[width=0.24\linewidth]{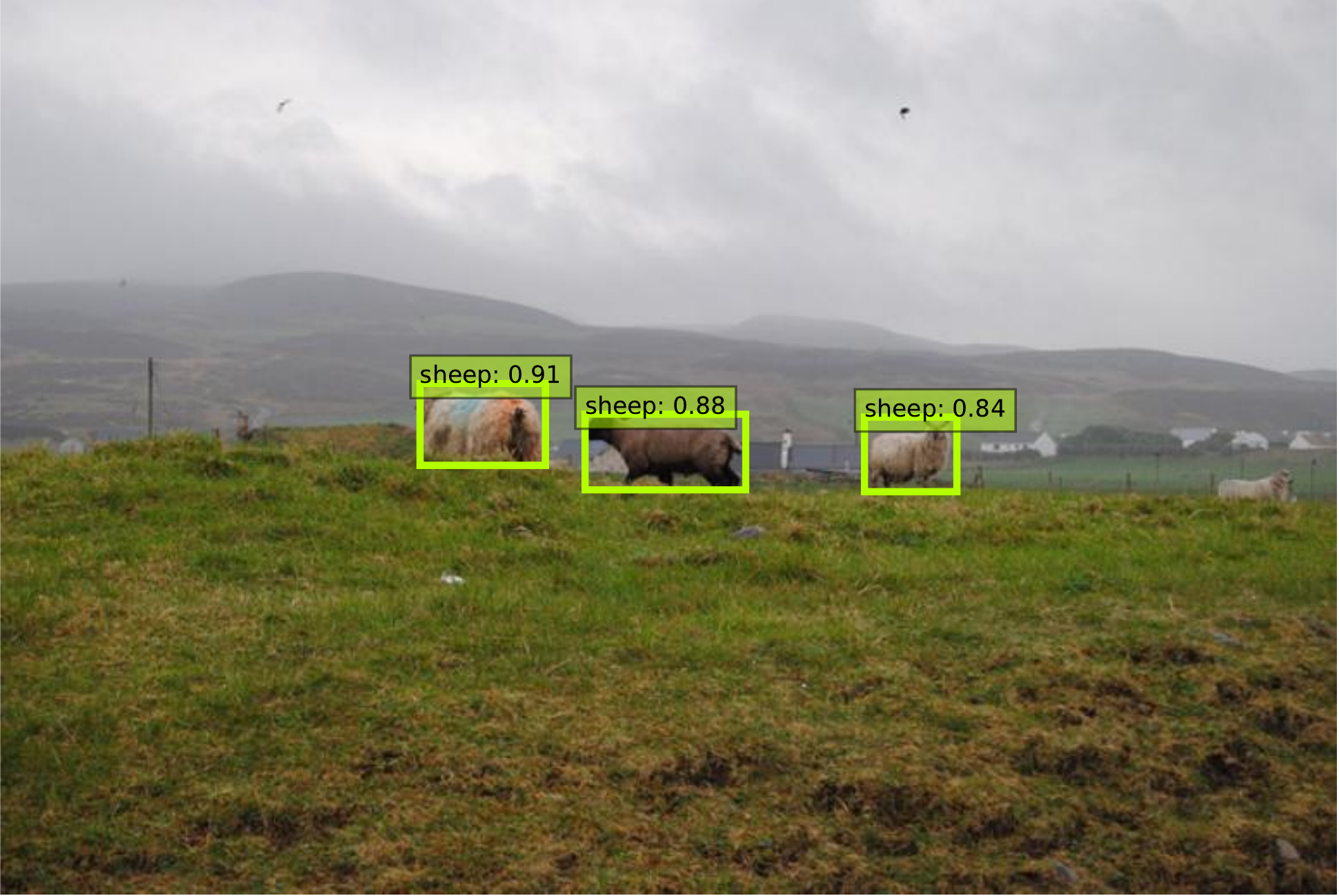}
 \\
  	\includegraphics[width=0.24\linewidth]{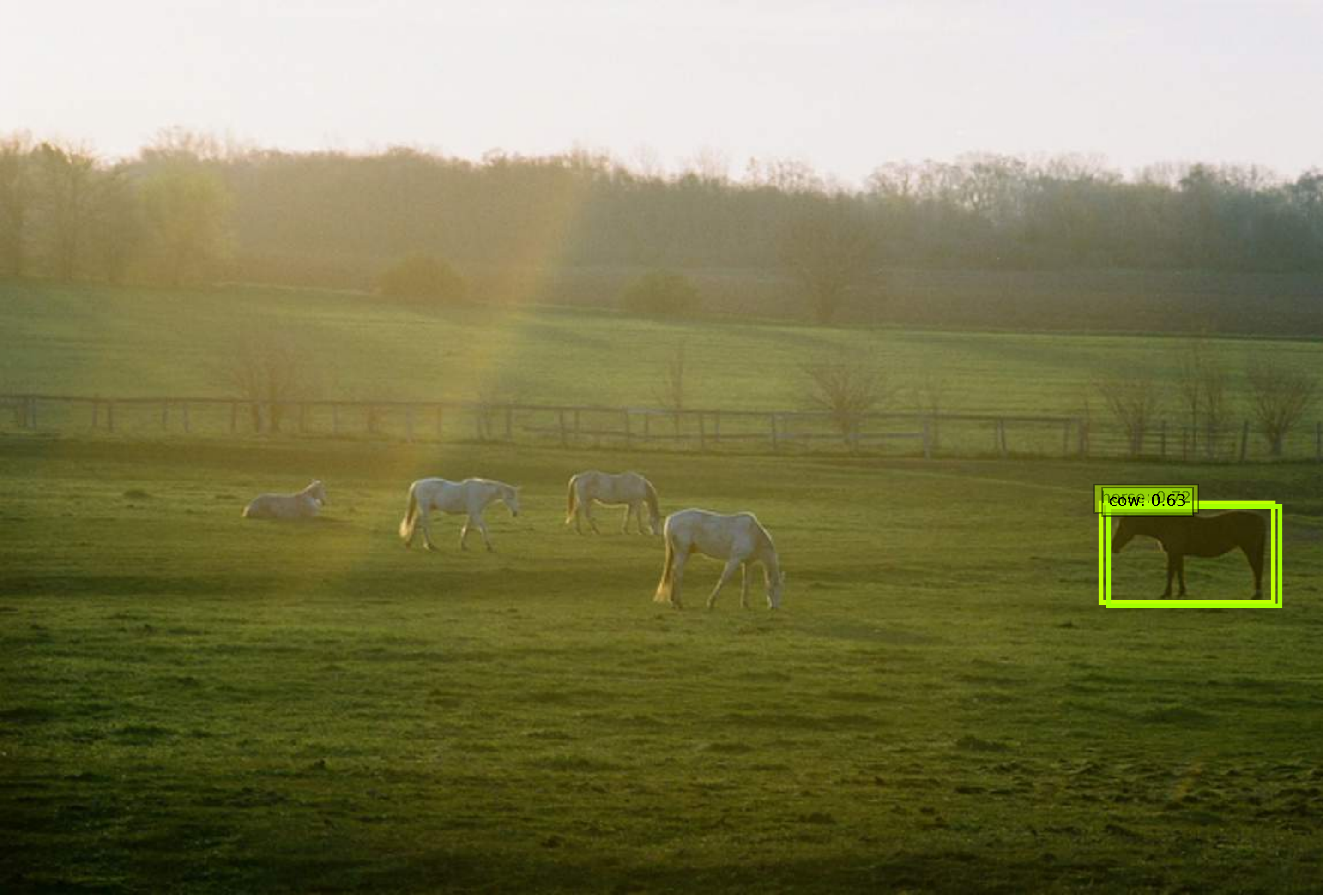}
    \includegraphics[width=0.24\linewidth]{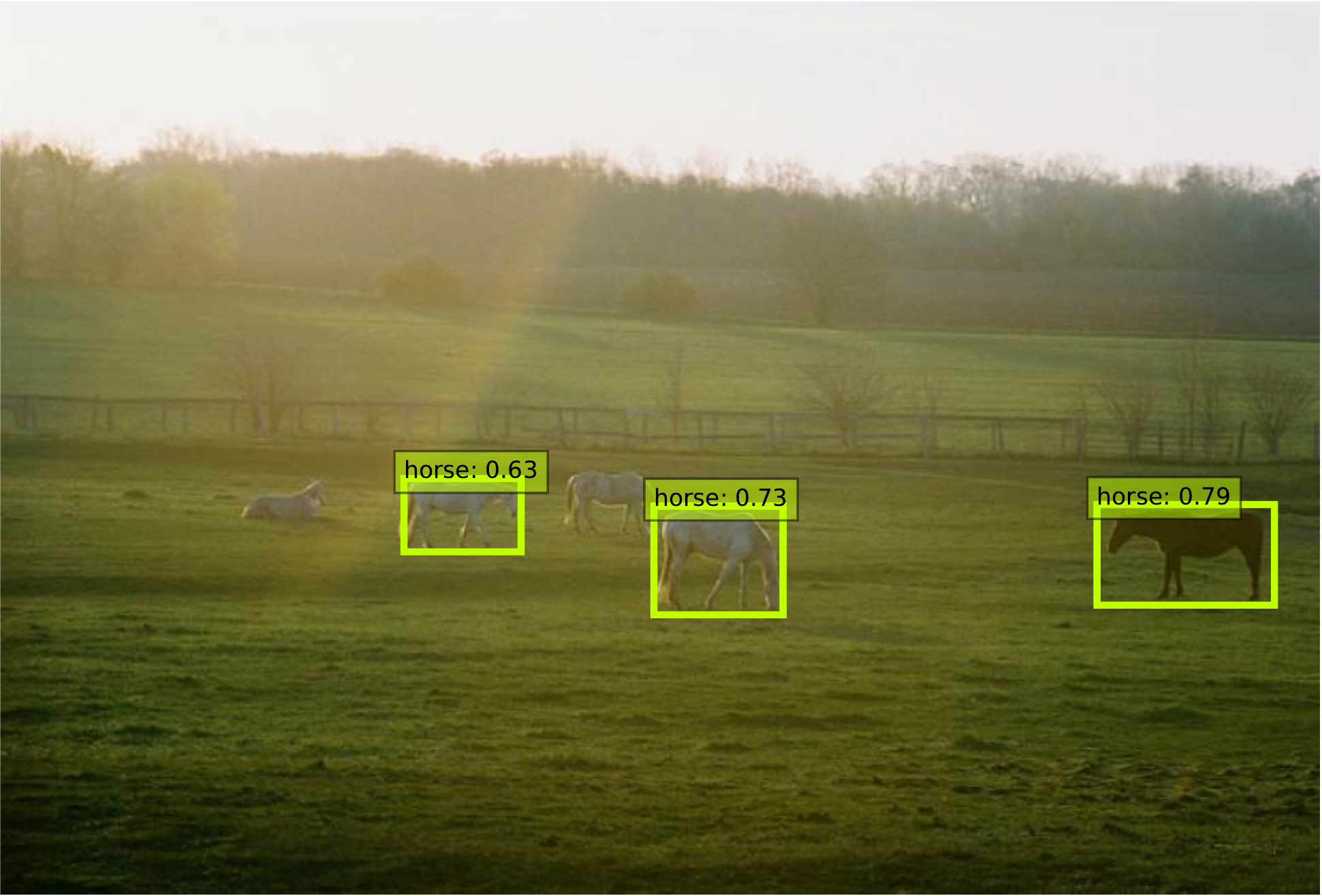}
    \includegraphics[width=0.24\linewidth]{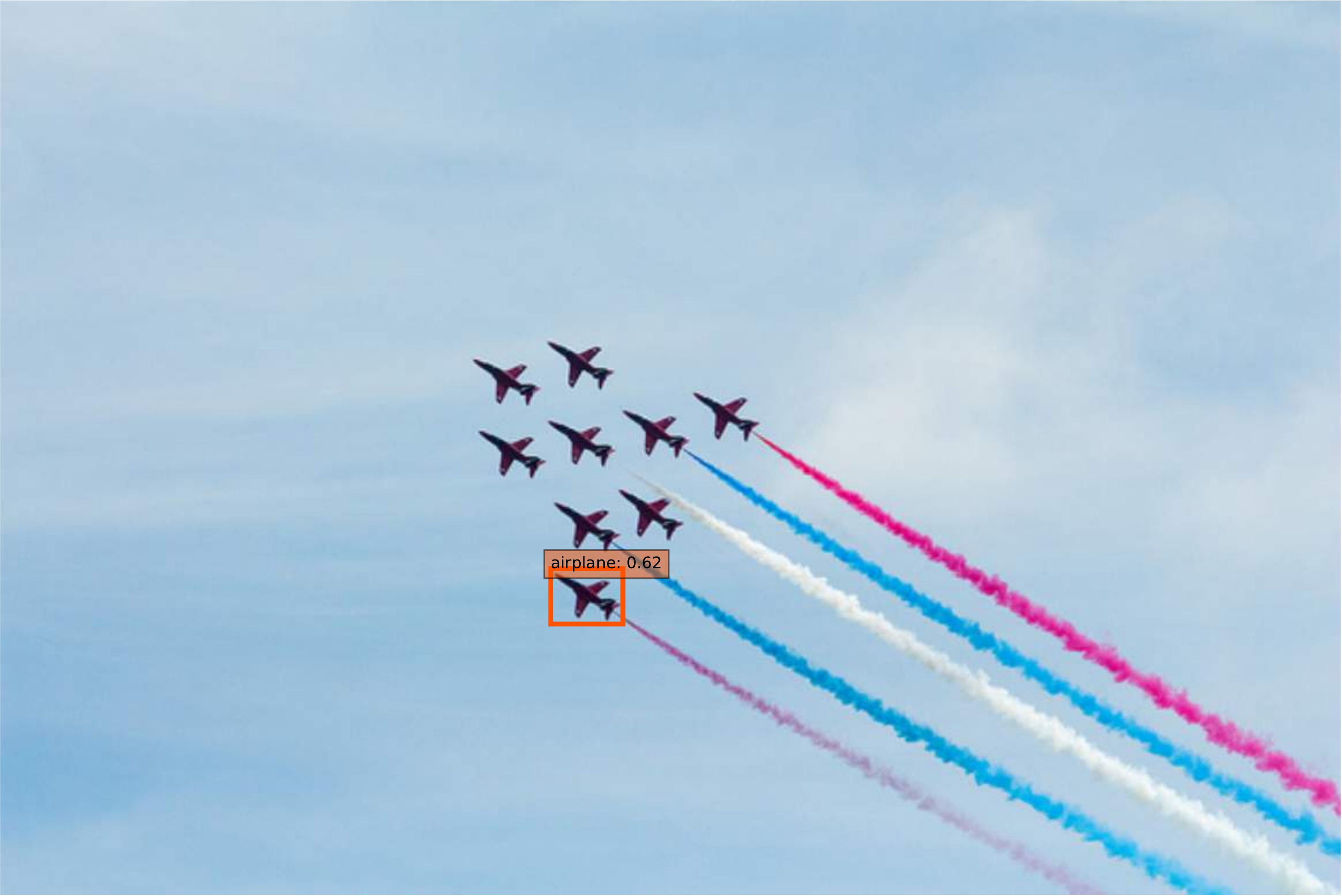}
    \includegraphics[width=0.24\linewidth]{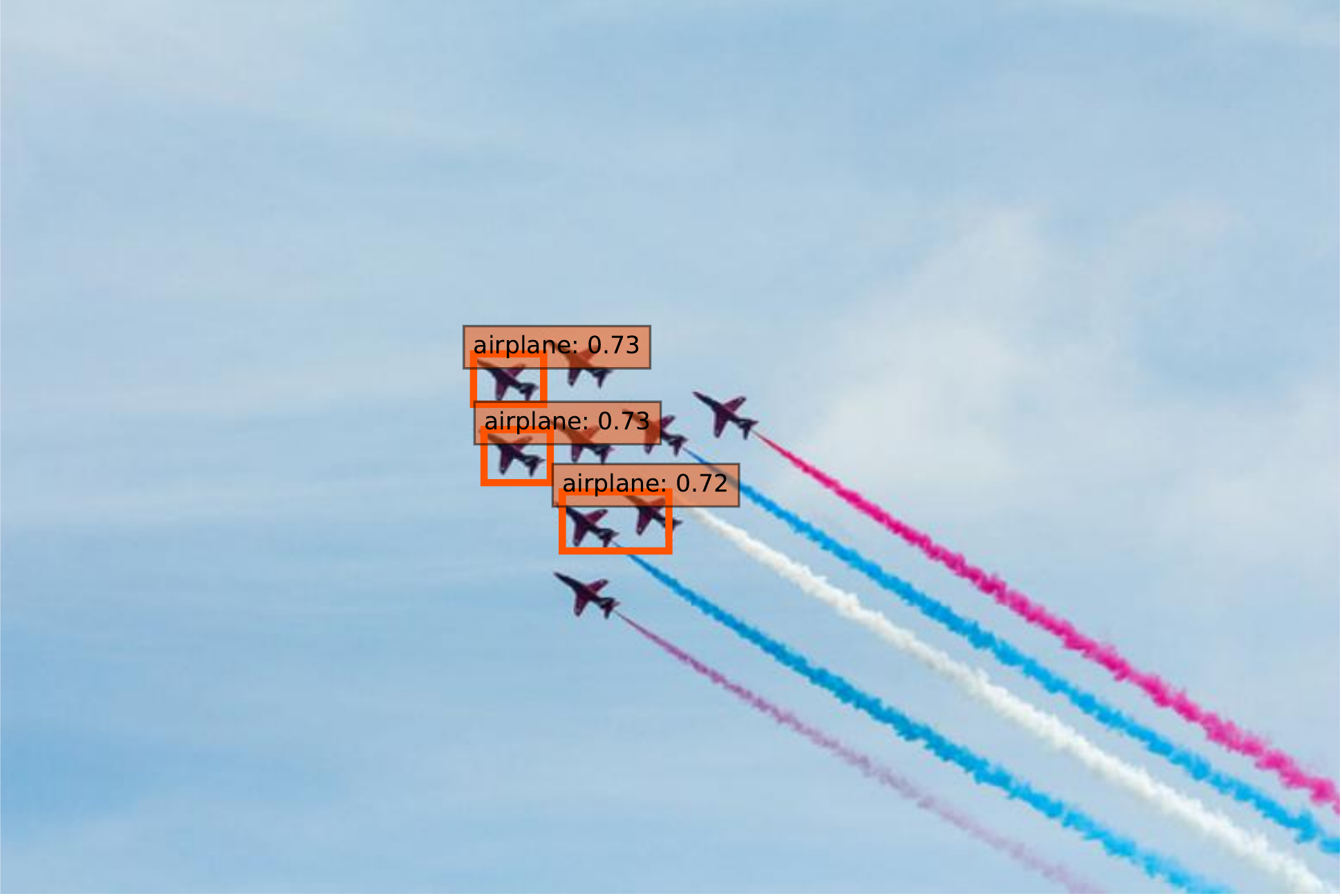}
 \\
    \includegraphics[width=0.24\linewidth]{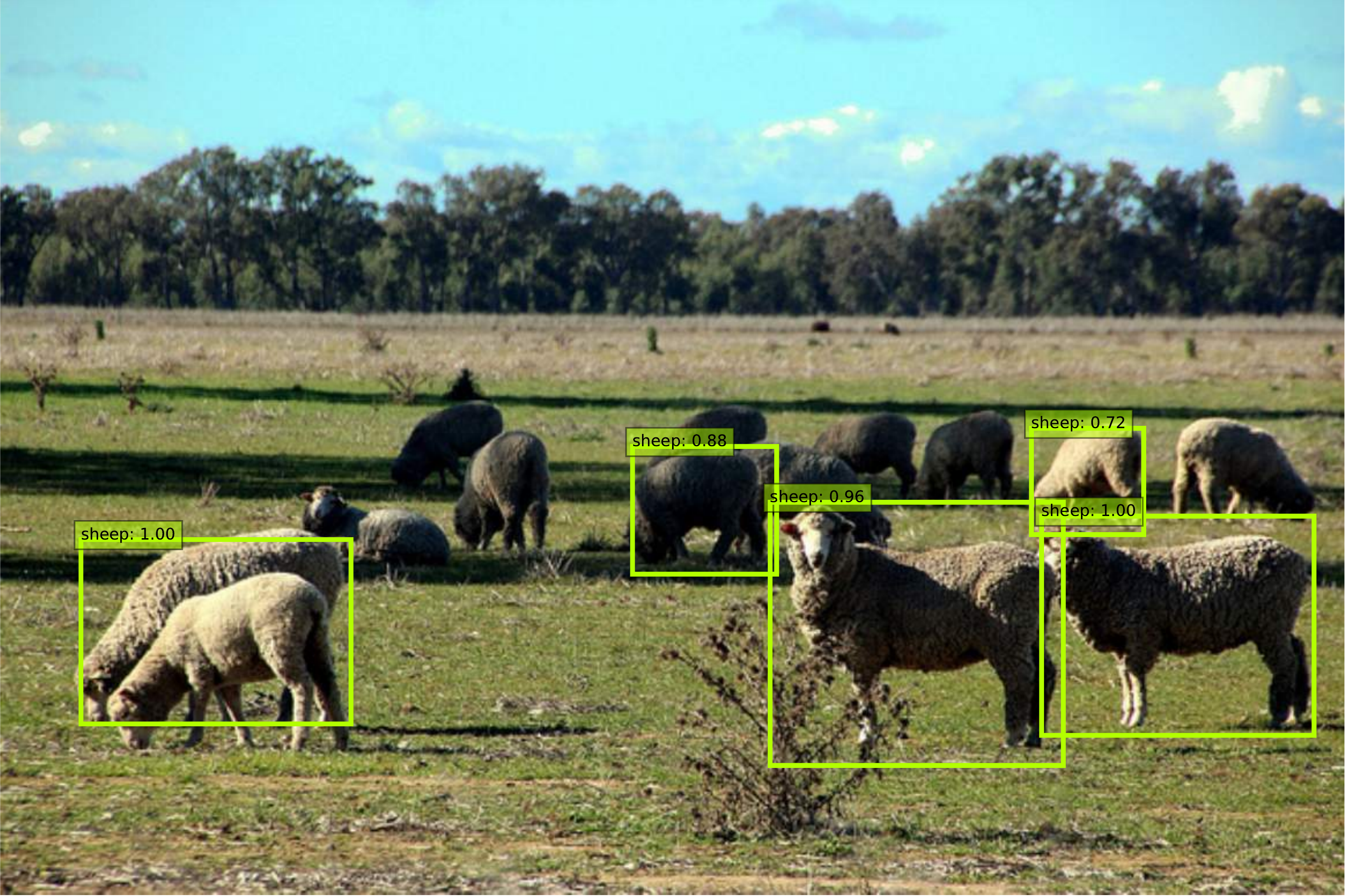}
    \includegraphics[width=0.24\linewidth]{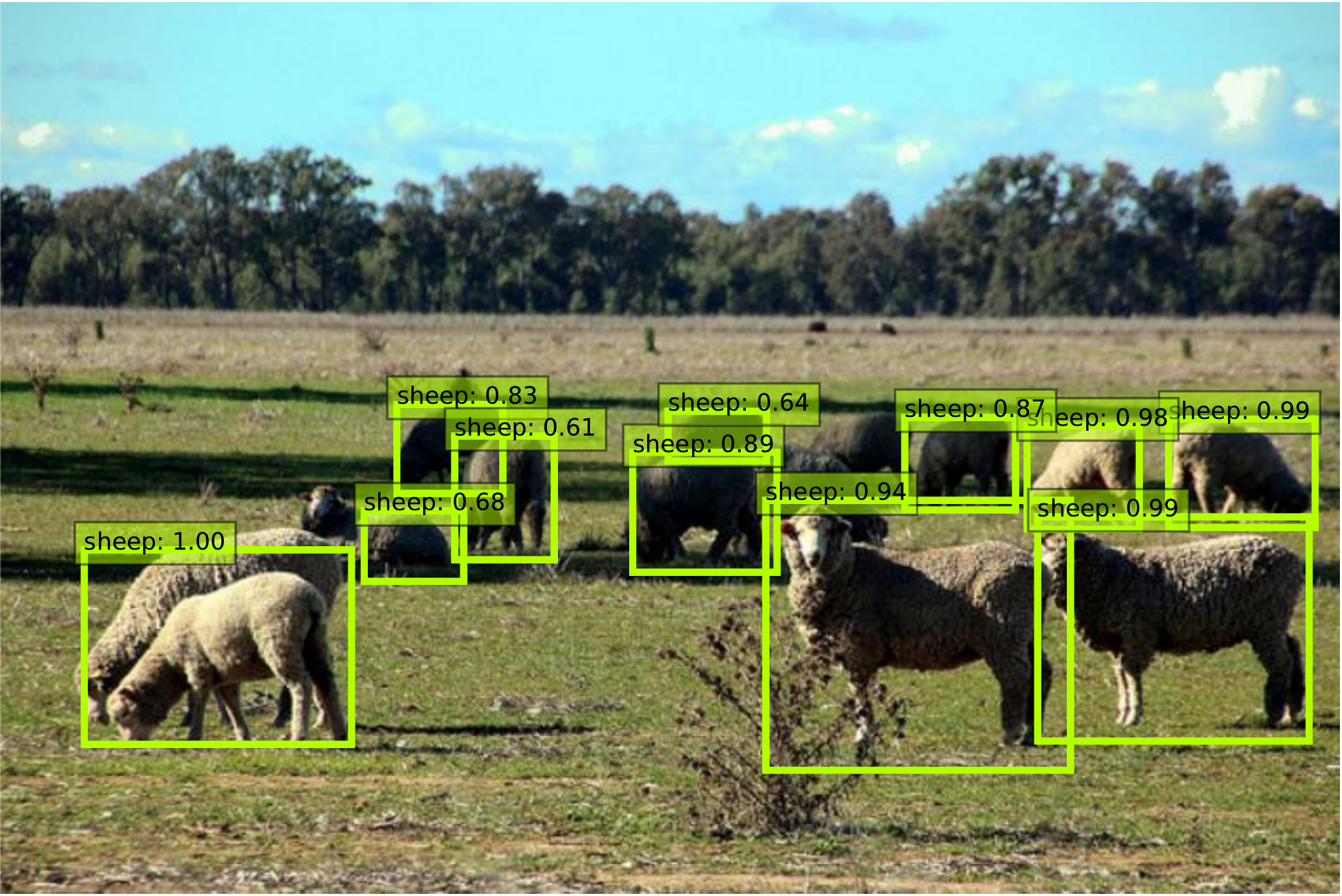}
    \includegraphics[width=0.24\linewidth]{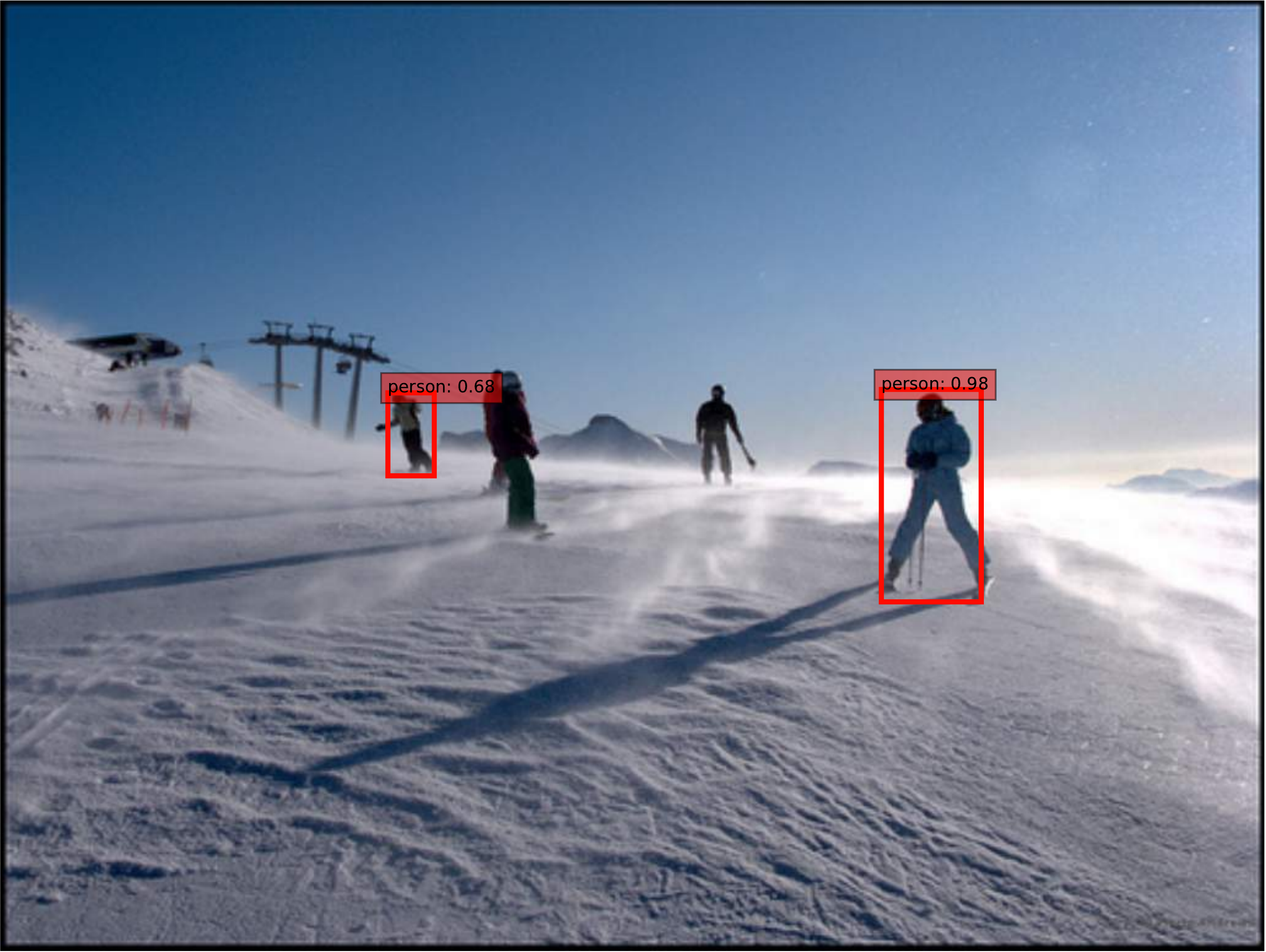}
    \includegraphics[width=0.24\linewidth]{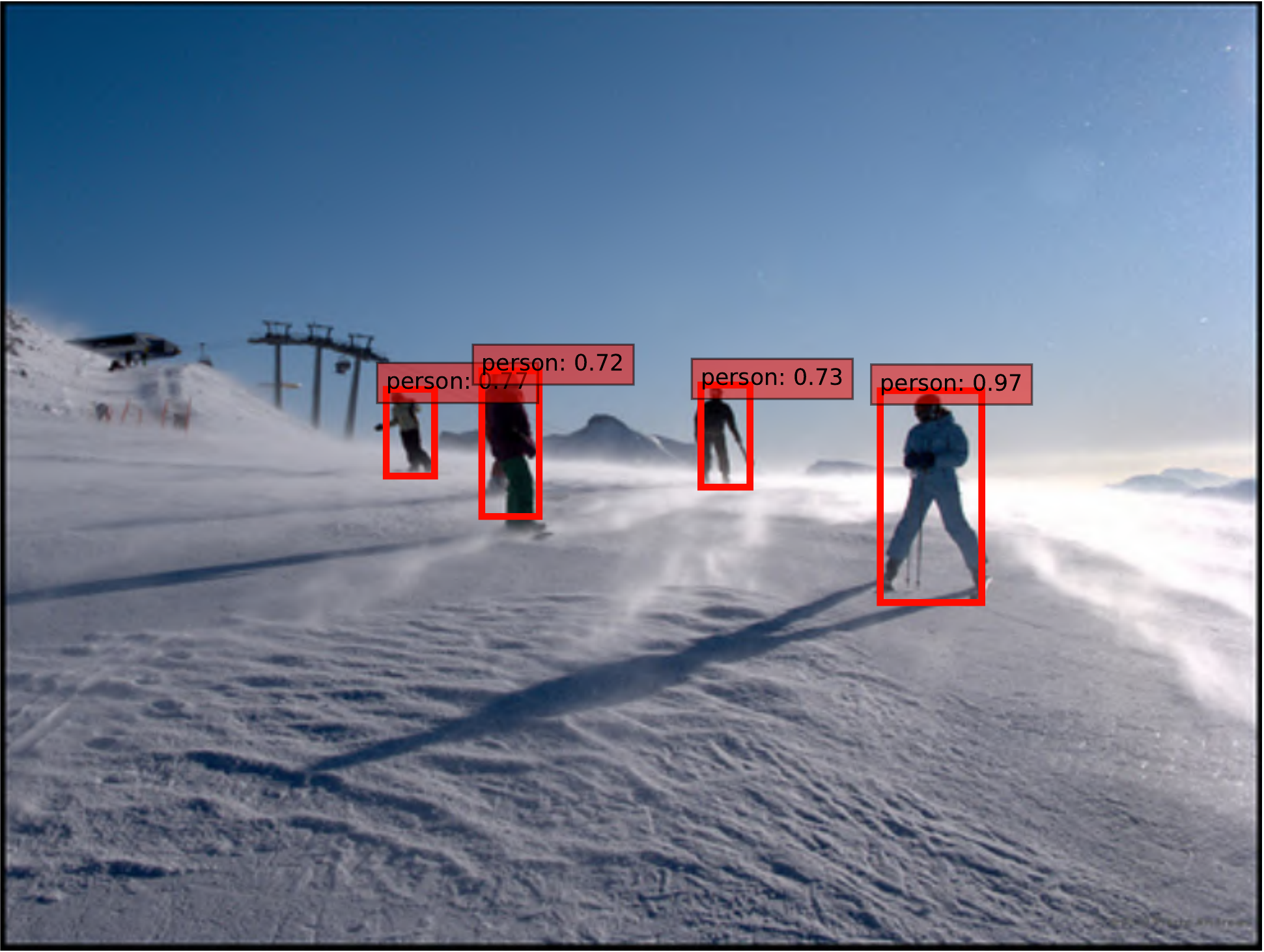}
  \\
    \includegraphics[width=0.24\linewidth]{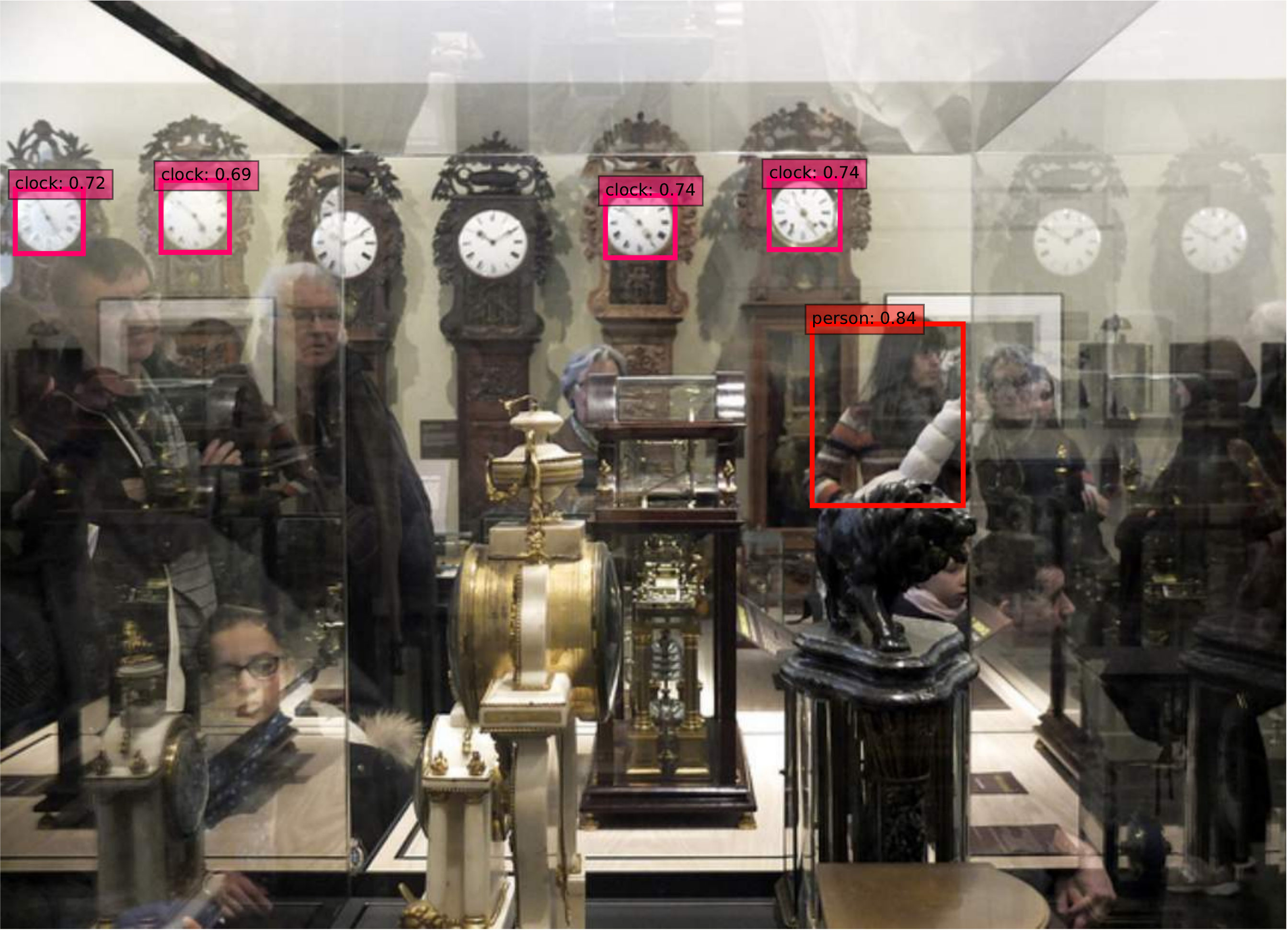}
    \includegraphics[width=0.24\linewidth]{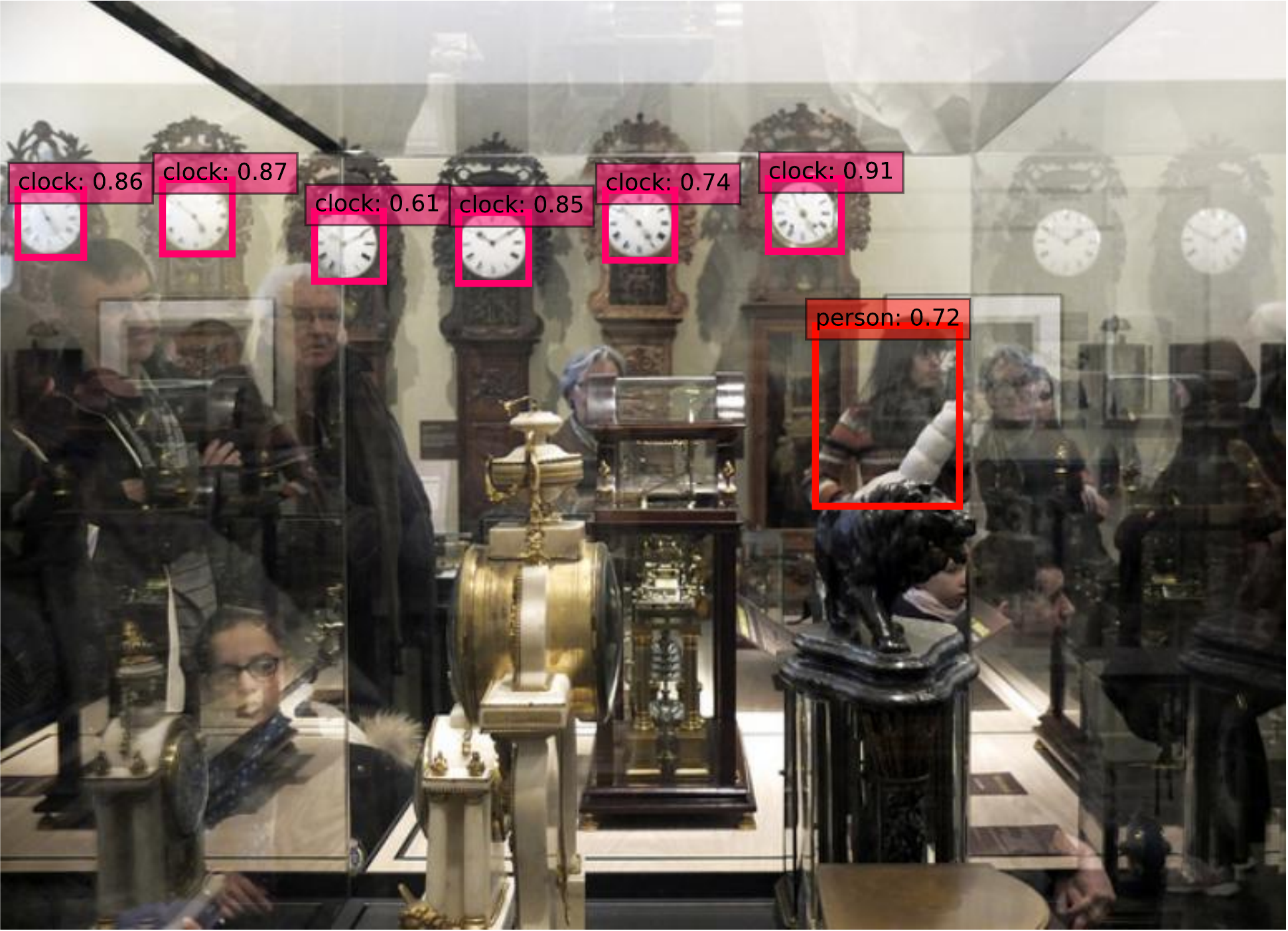}
        \includegraphics[width=0.24\linewidth]{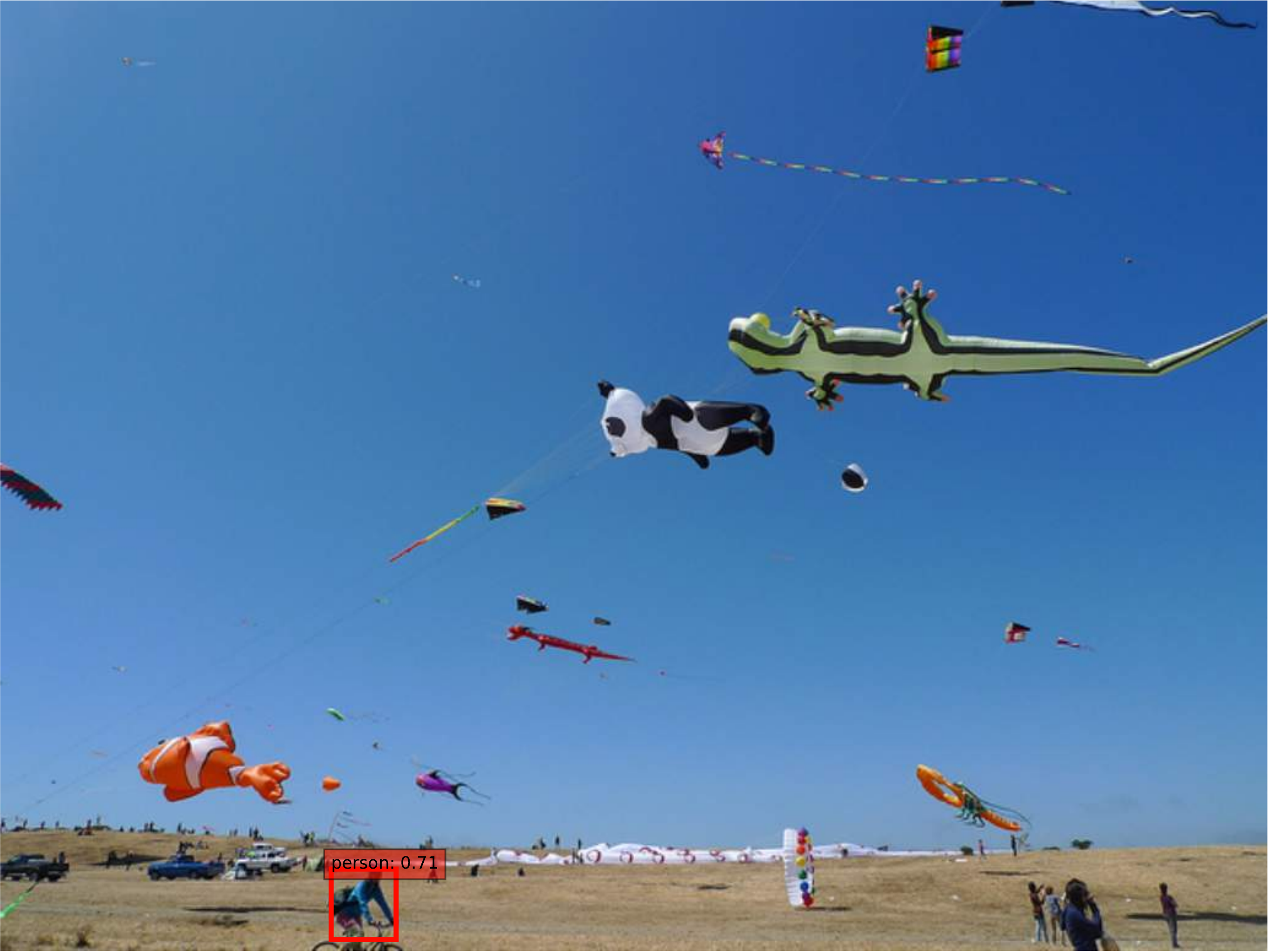}
    \includegraphics[width=0.24\linewidth]{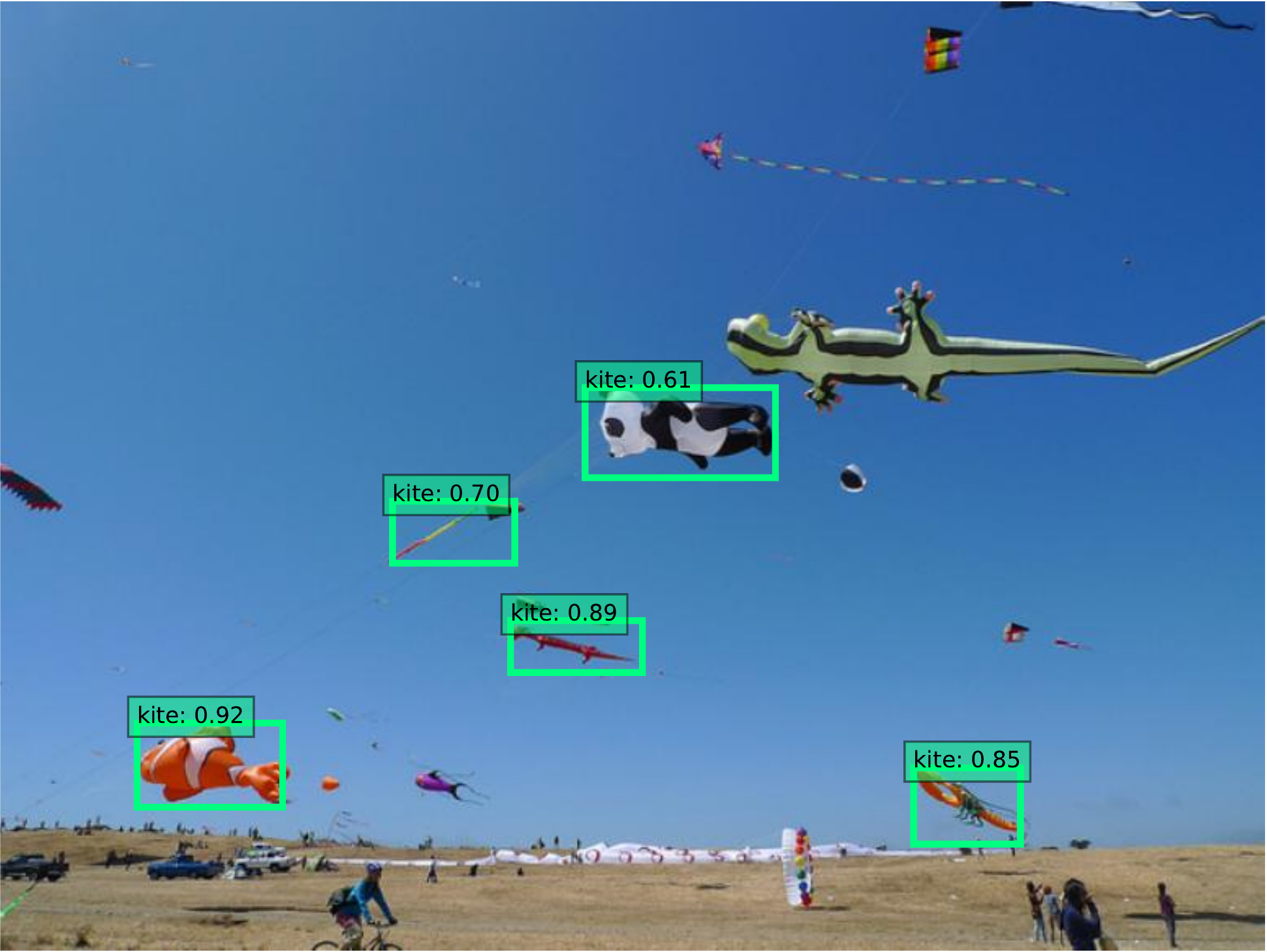}
  \caption{\textbf{Dense Cases}:DSSD considers context more compared to SSD. This yields better performance on small objects and dense scenes. }
  \label{sfig:dense}
  \end{subfigure}
  \phantomcaption
\end{figure*}

\begin{figure*}
\ContinuedFloat
	\centering
    \begin{subfigure}{\textwidth}
	\includegraphics[width=0.24\linewidth]{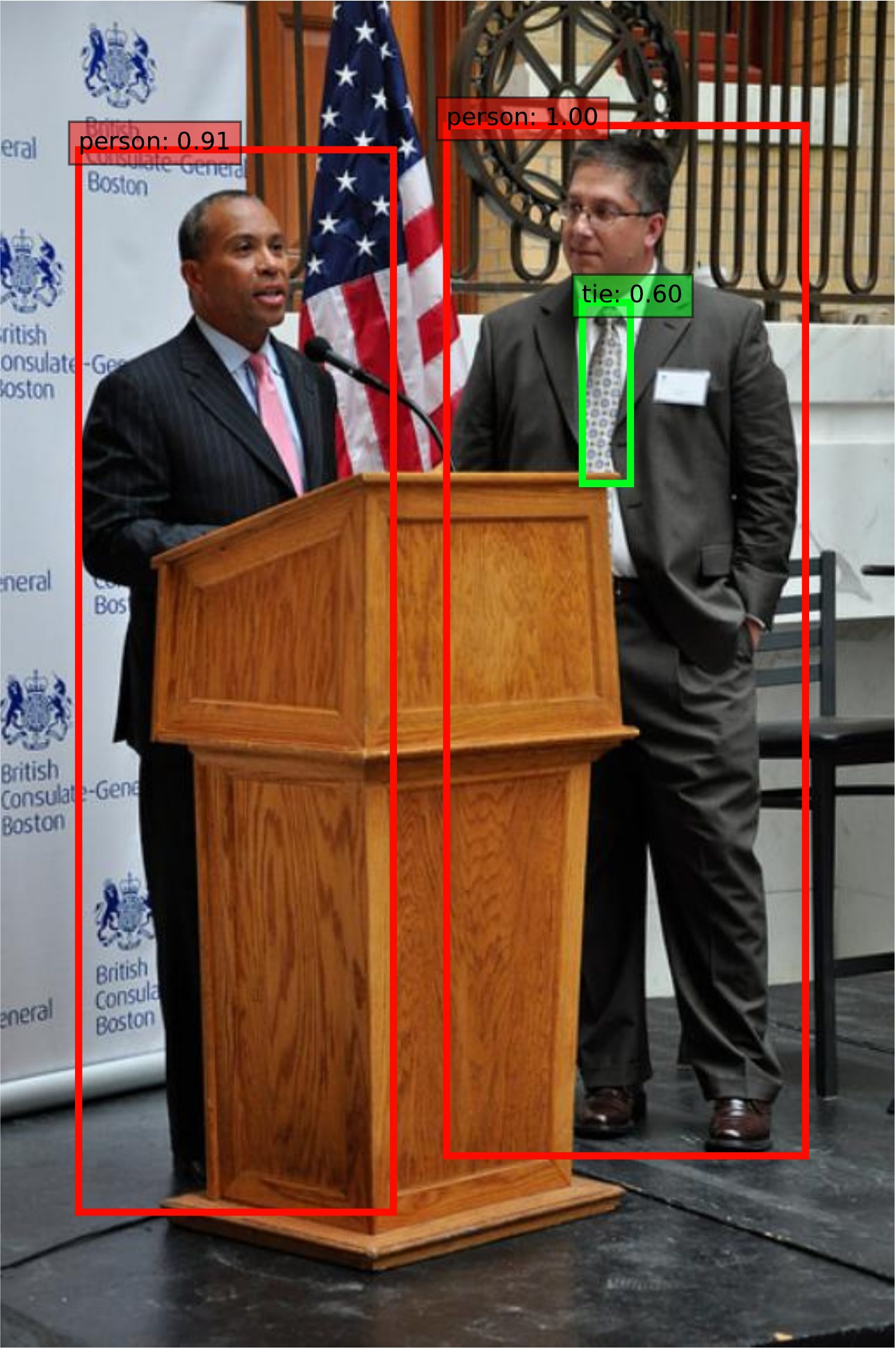}
    \includegraphics[width=0.24\linewidth]{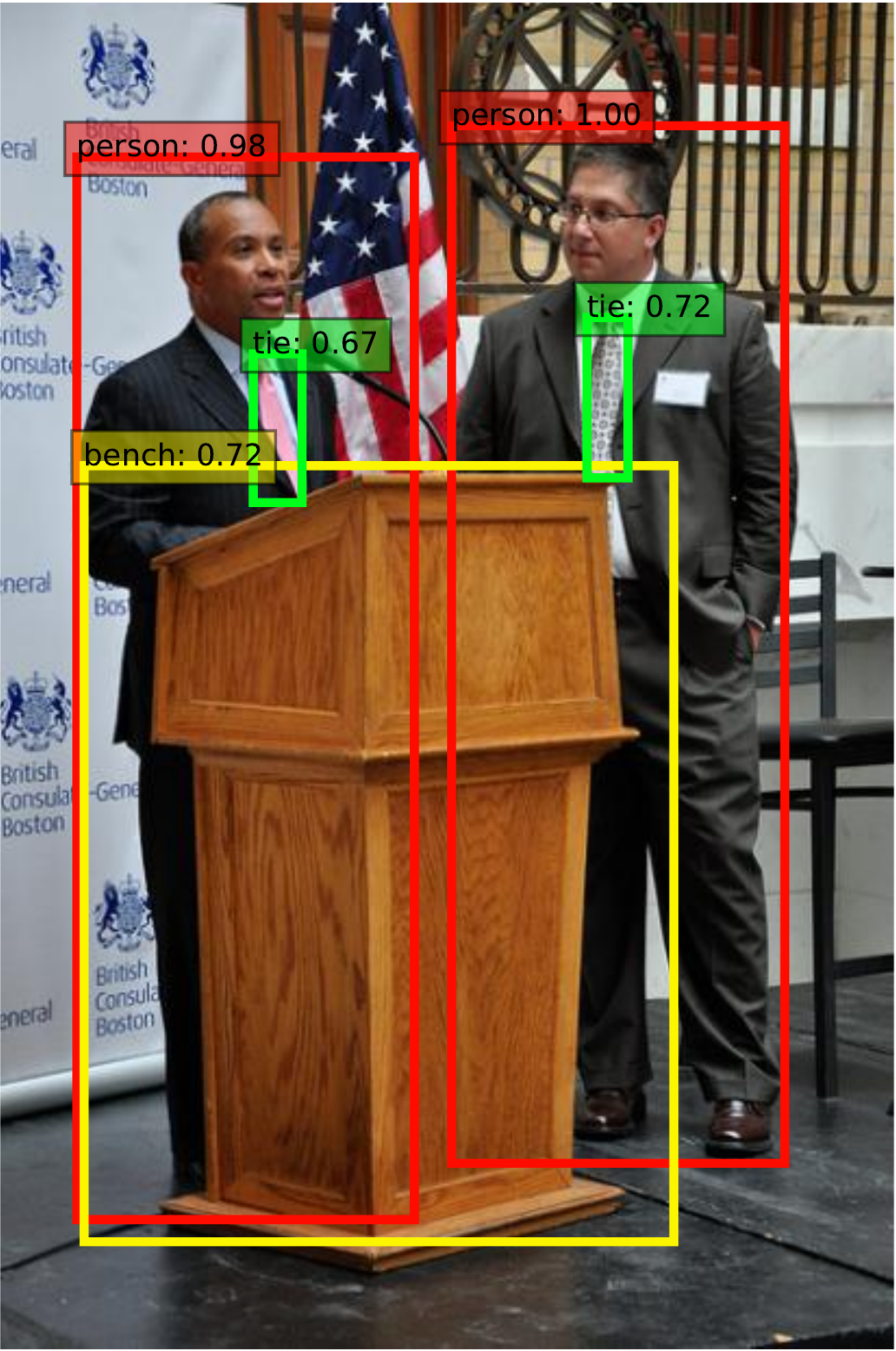}
    \includegraphics[width=0.24\linewidth]{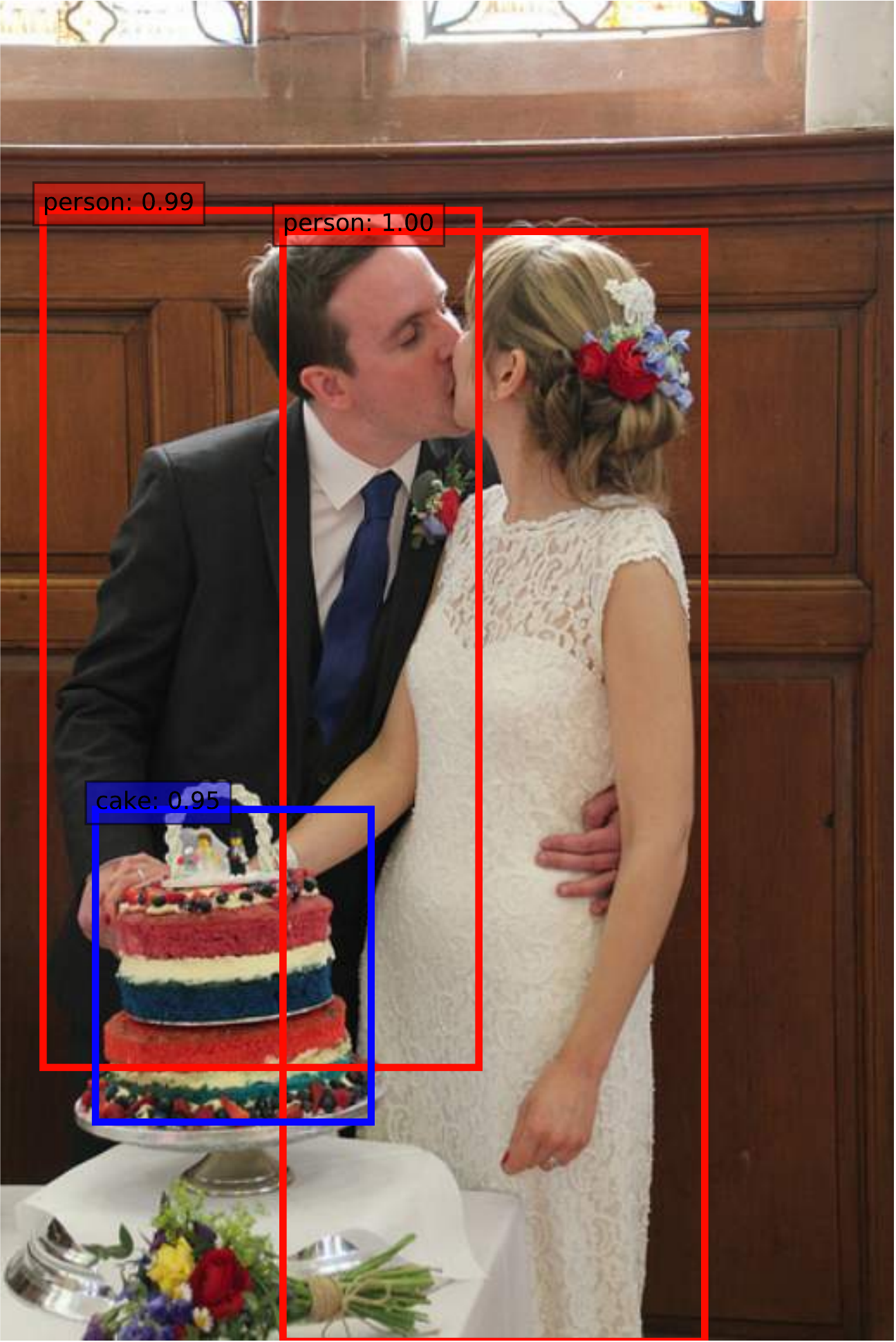}
    \includegraphics[width=0.24\linewidth]{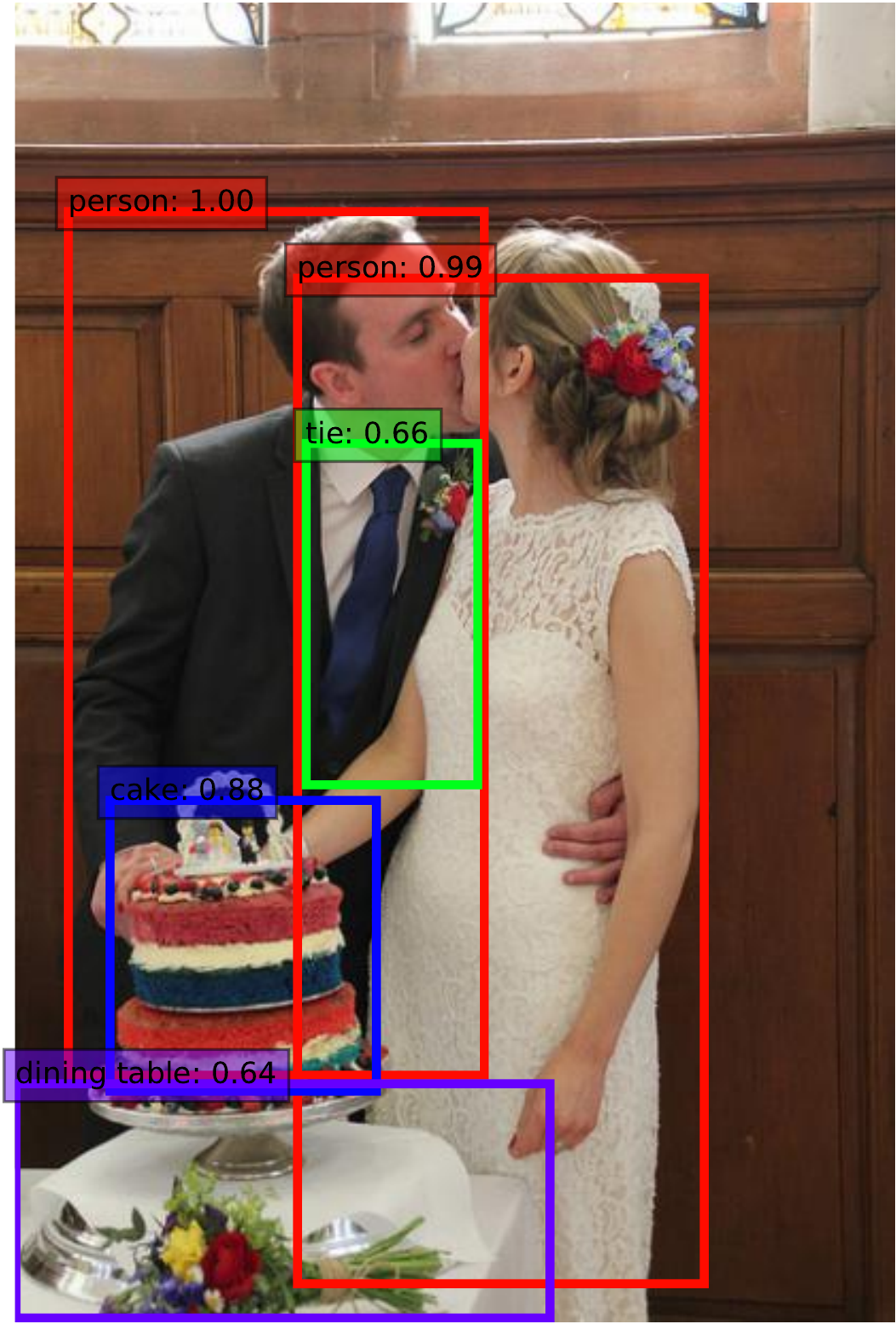}
    \\
    \includegraphics[width=0.24\linewidth]{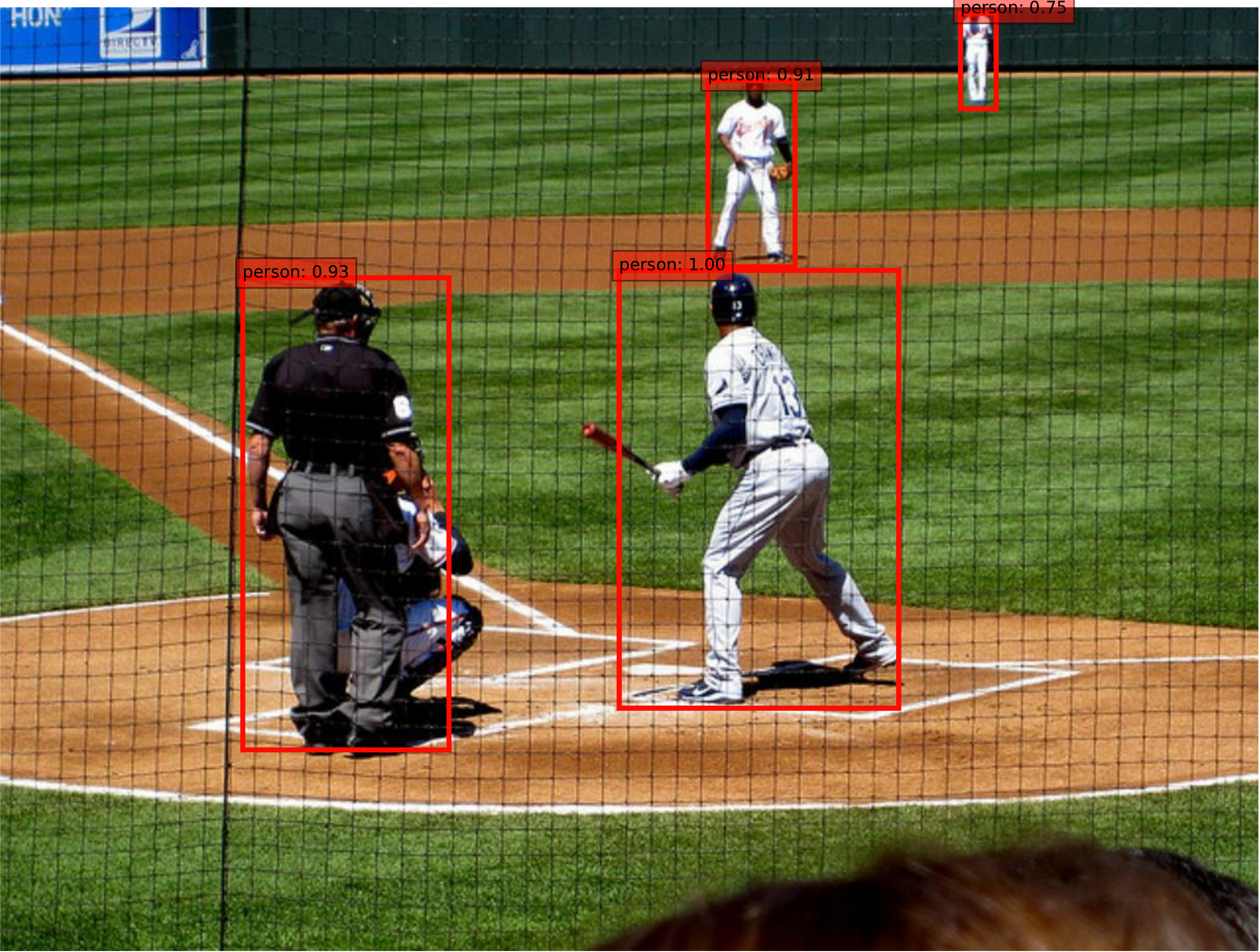}
    \includegraphics[width=0.24\linewidth]{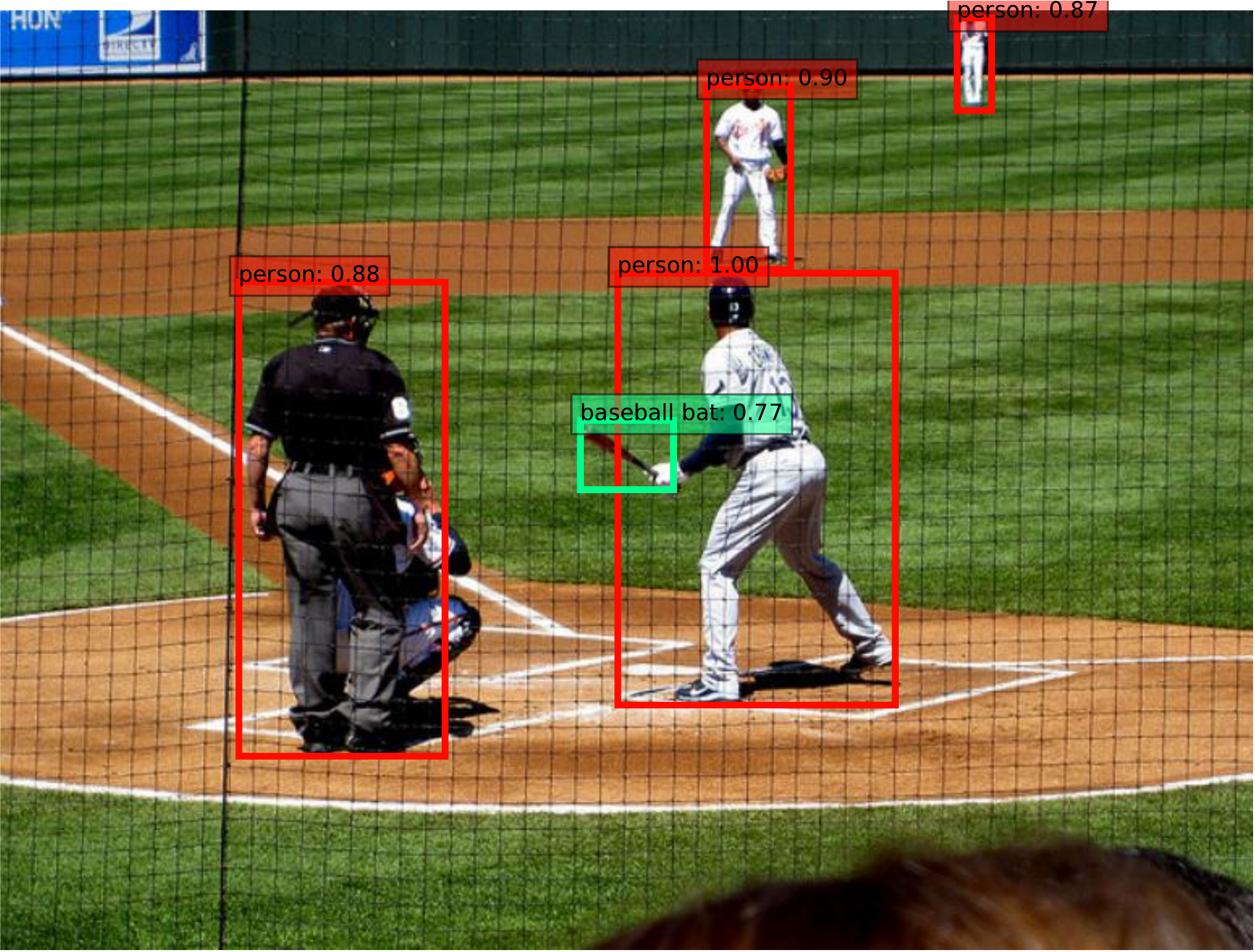}
    \includegraphics[width=0.24\linewidth]{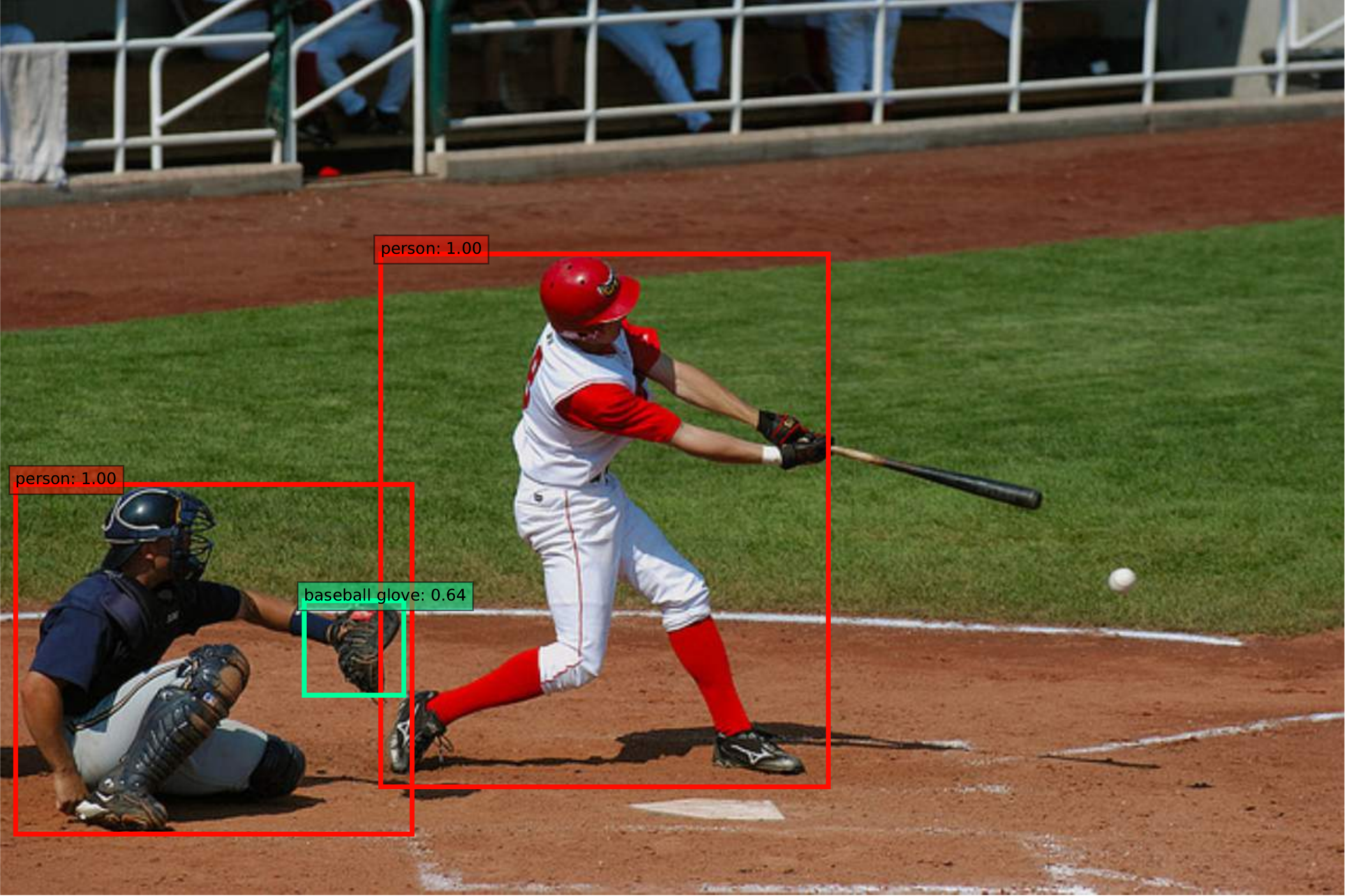}
    \includegraphics[width=0.24\linewidth]{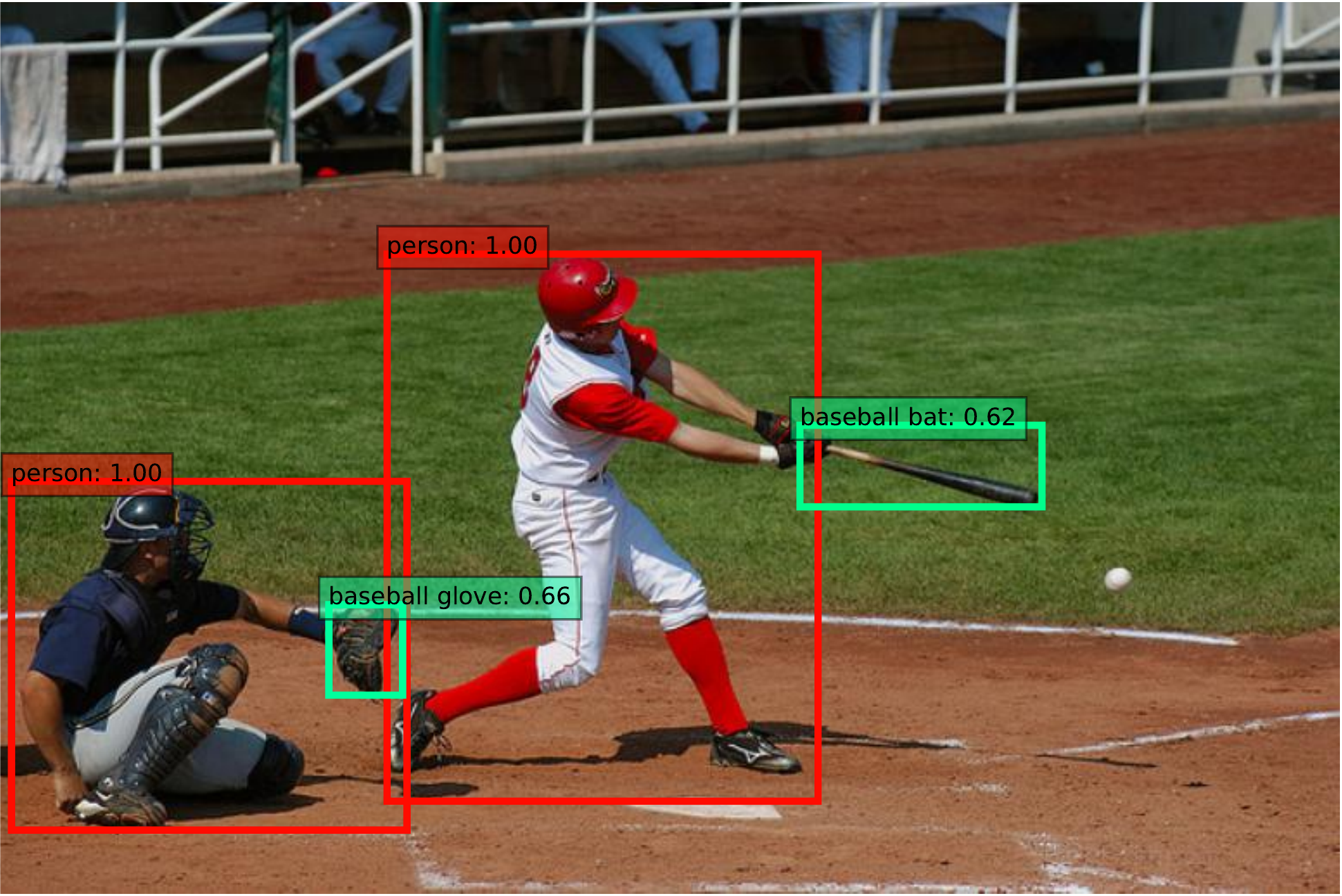}
    \\

    \includegraphics[width=0.24\linewidth]{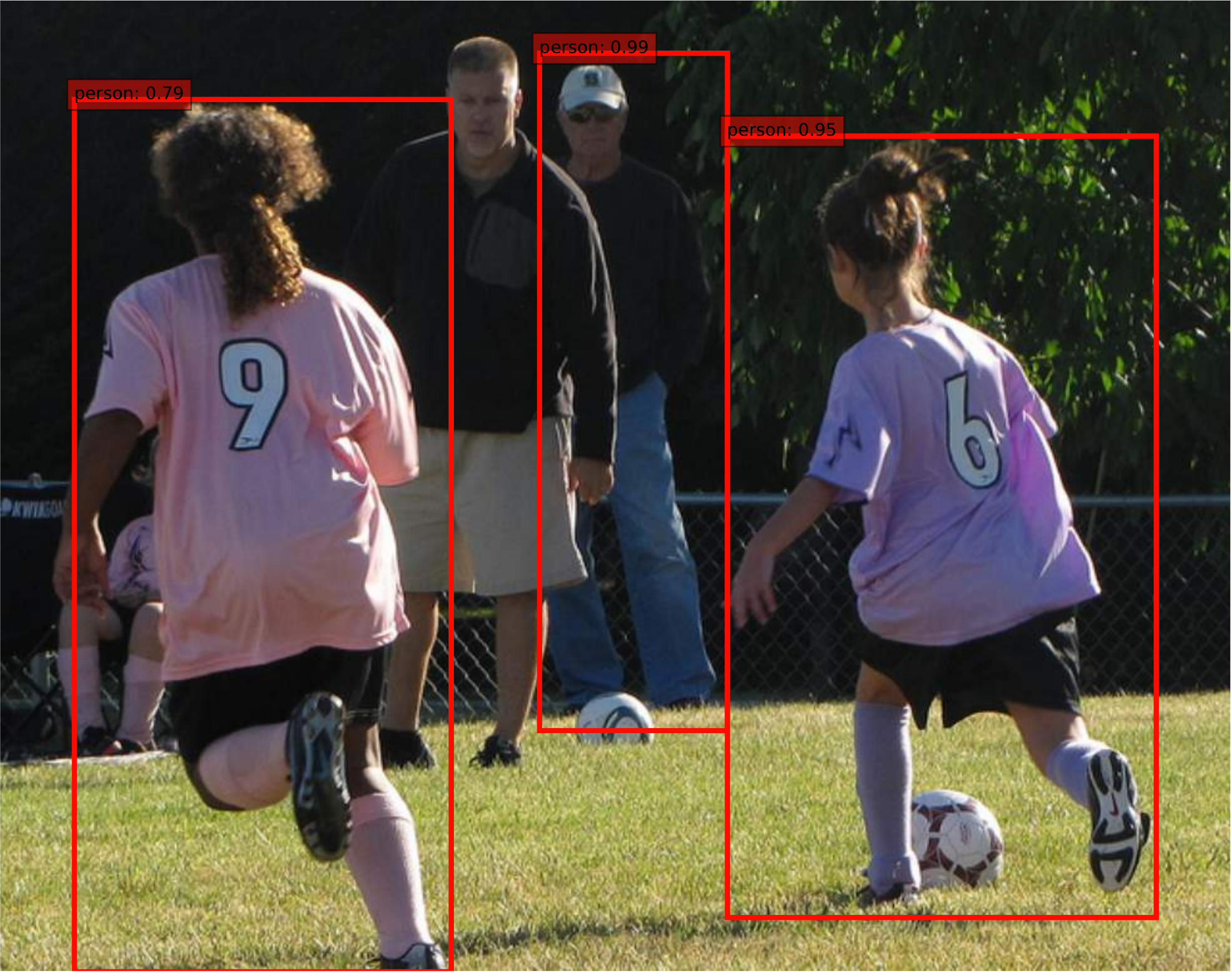}
    \includegraphics[width=0.24\linewidth]{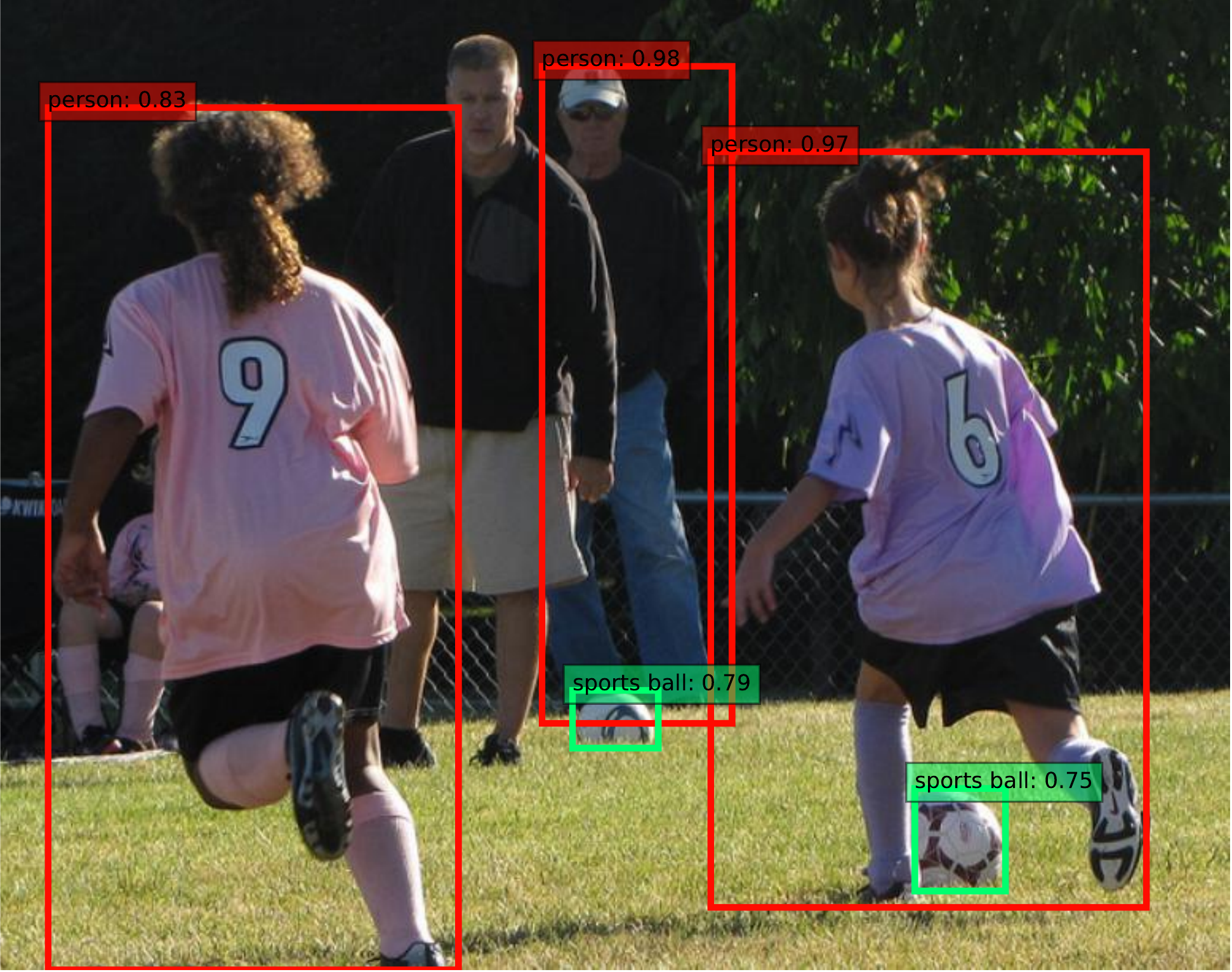}
    \includegraphics[width=0.24\linewidth]{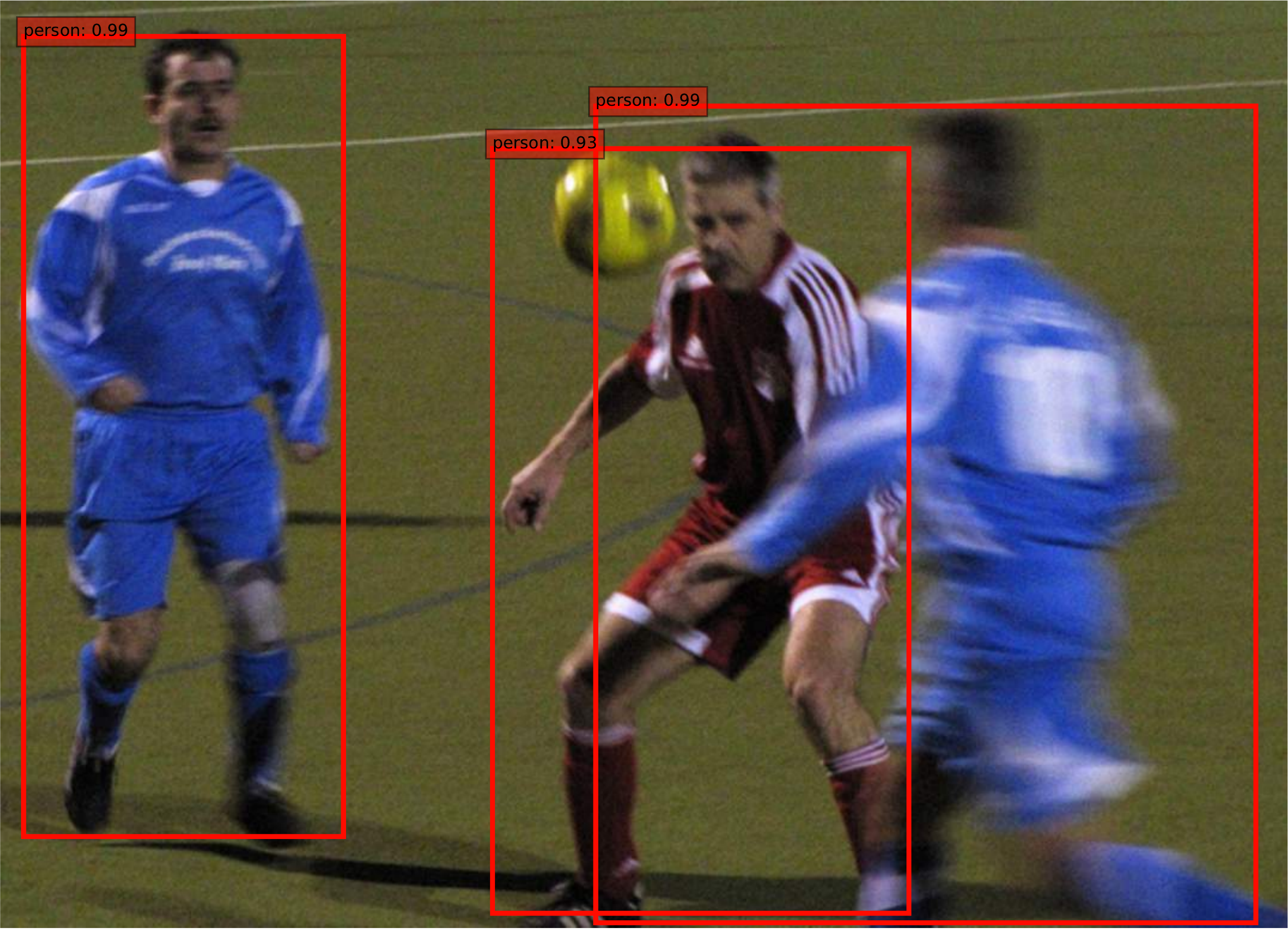}
    \includegraphics[width=0.24\linewidth]{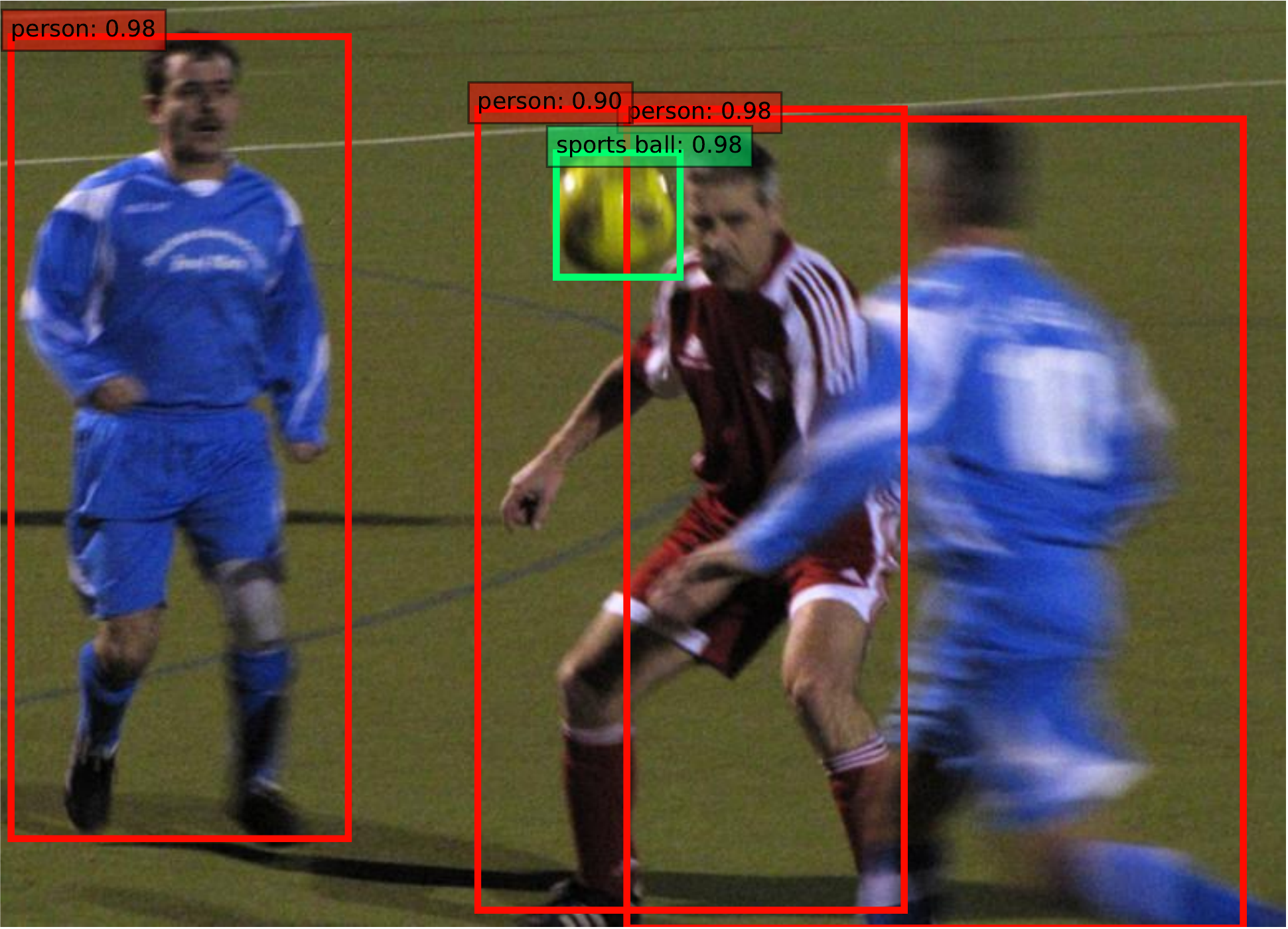}
    \\
    \includegraphics[width=0.24\linewidth]{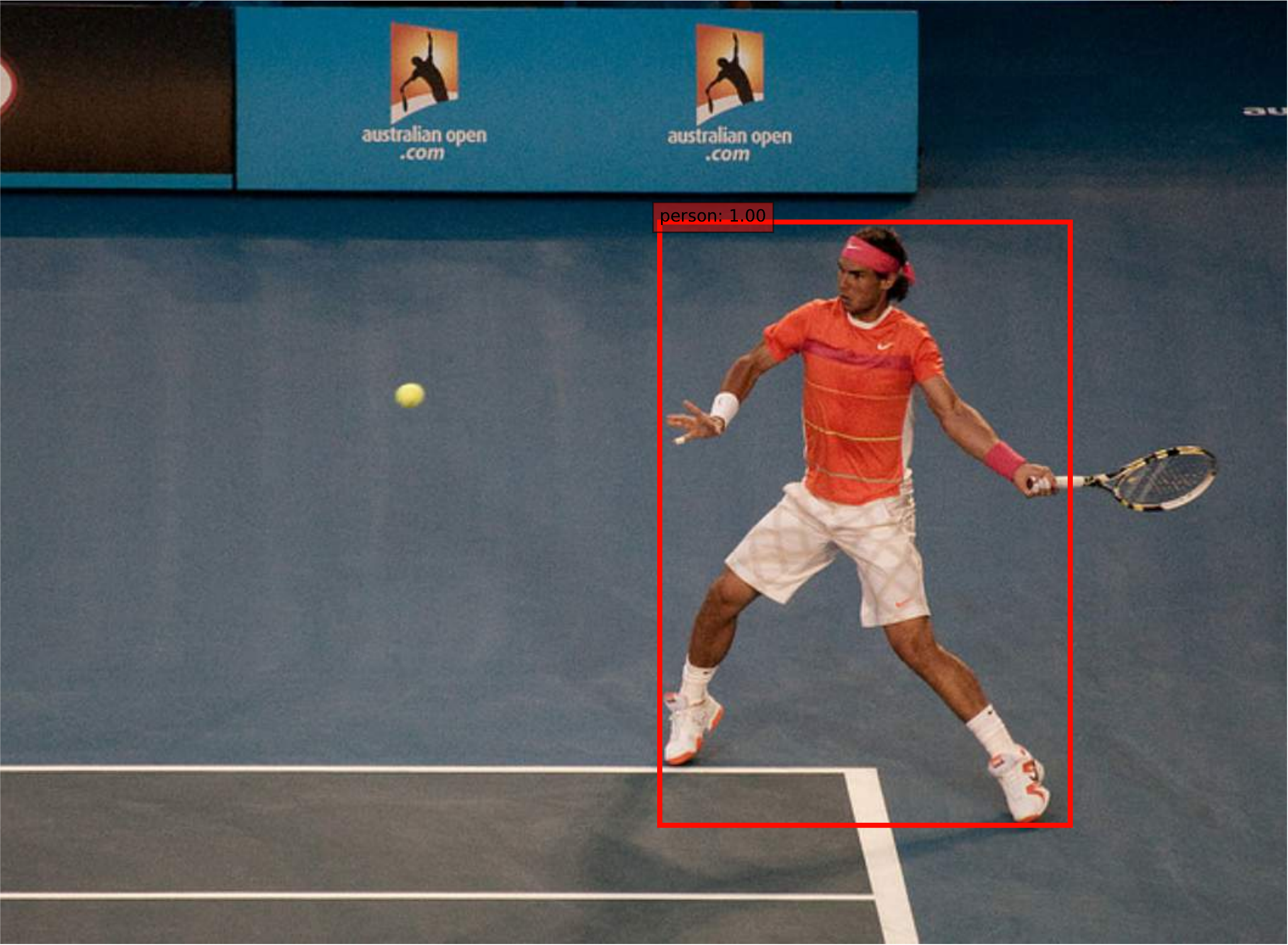}
    \includegraphics[width=0.24\linewidth]{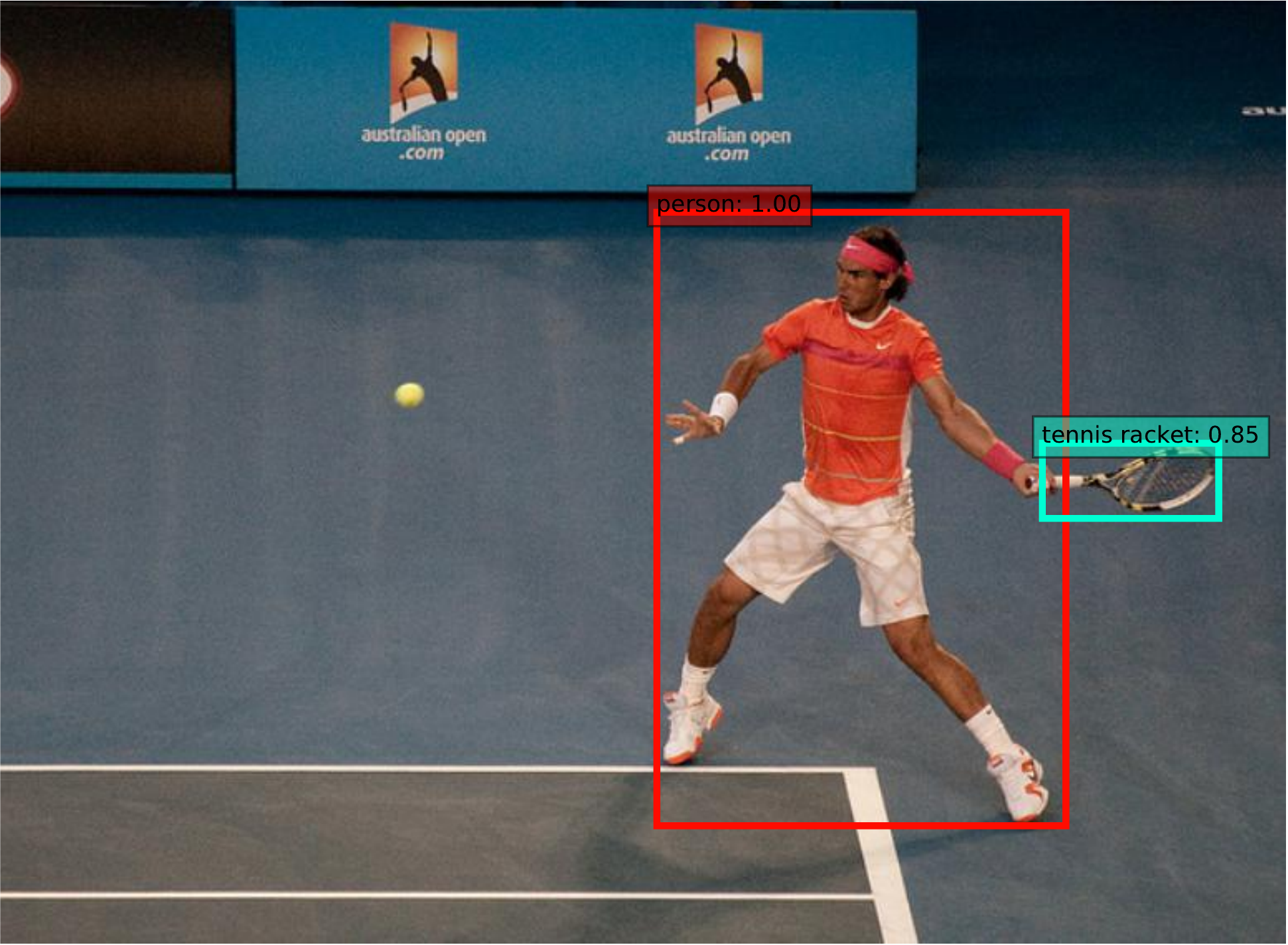}
    \includegraphics[width=0.24\linewidth]{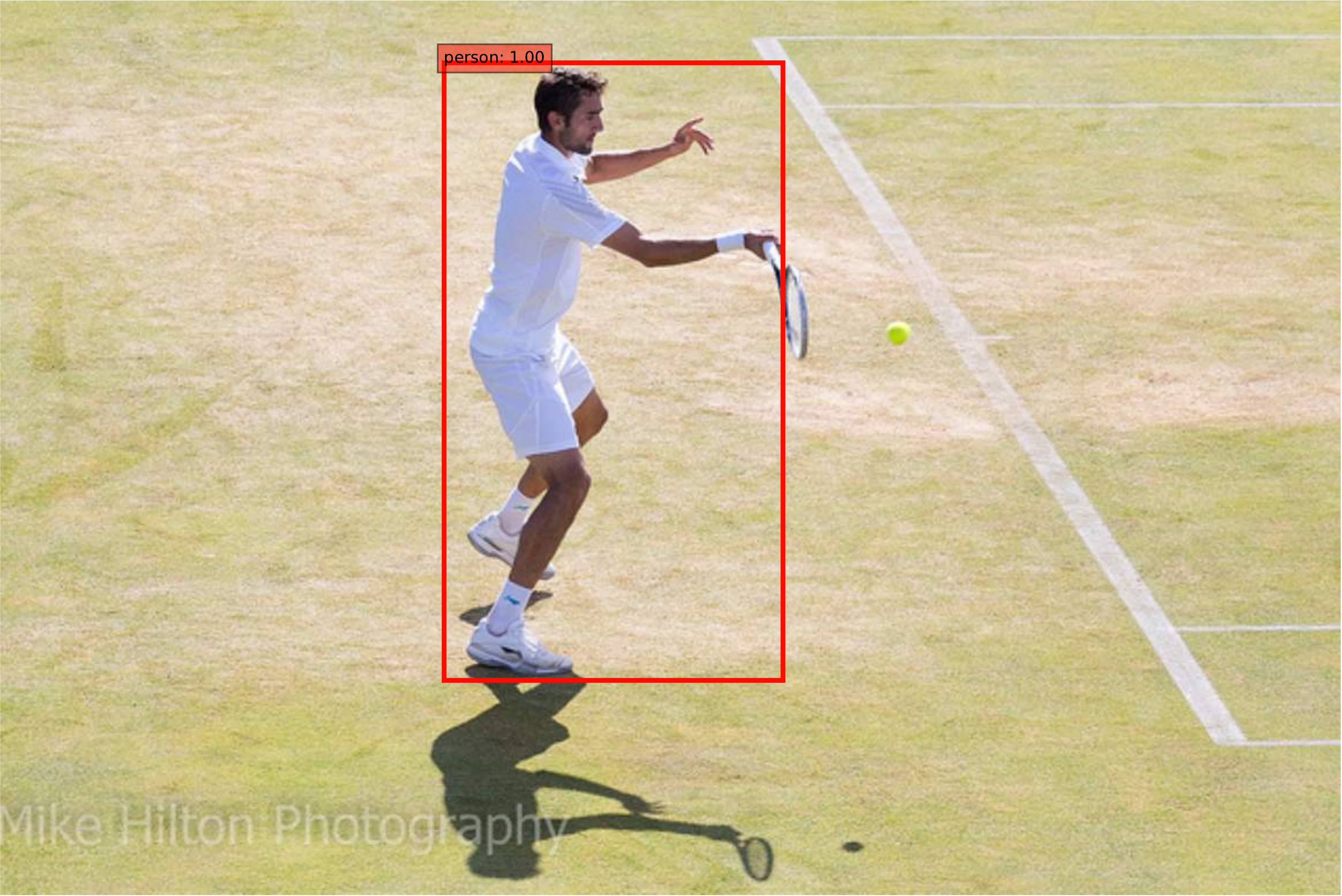}
    \includegraphics[width=0.24\linewidth]{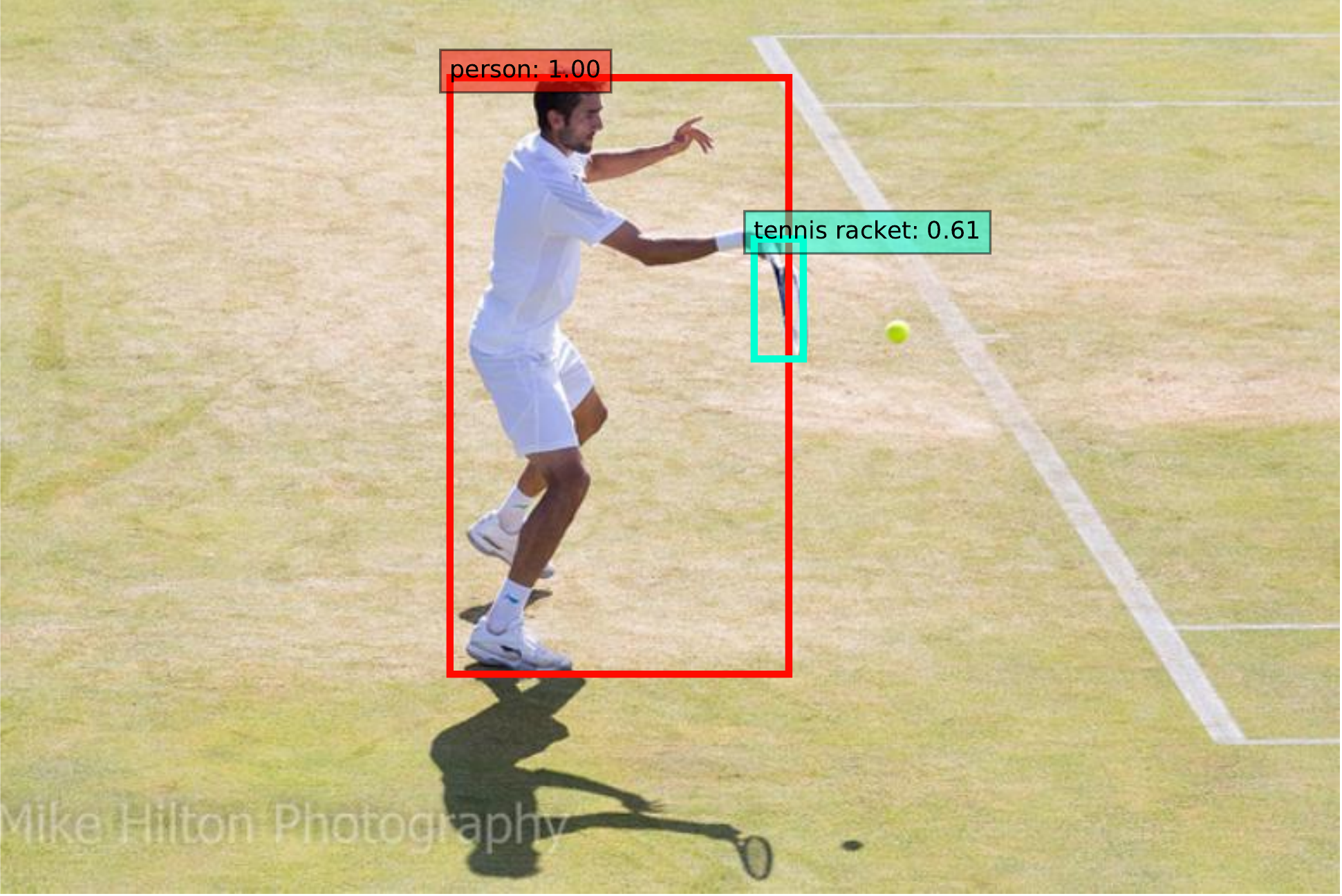}
    \\
    \includegraphics[width=0.24\linewidth]{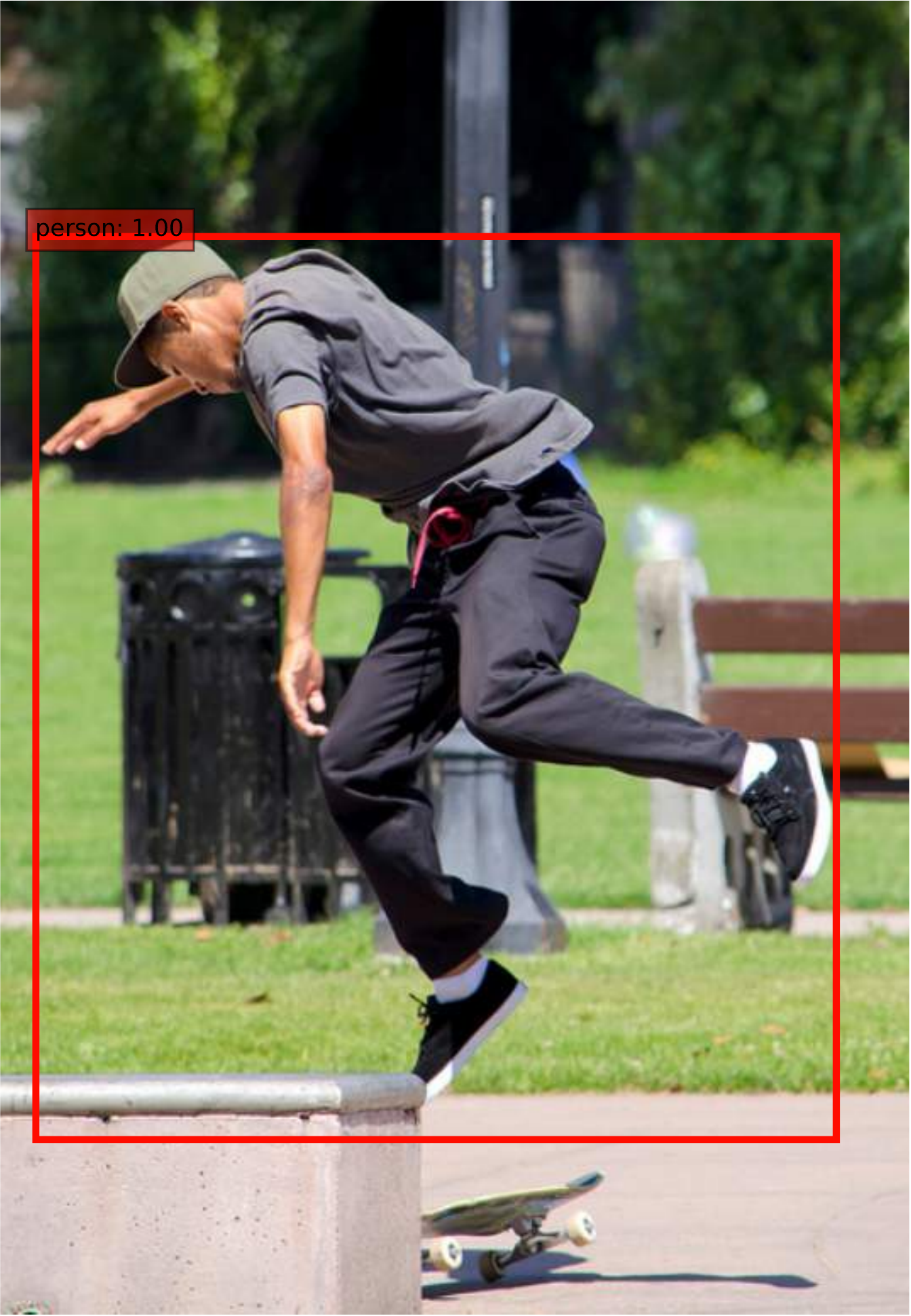}
    \includegraphics[width=0.24\linewidth]{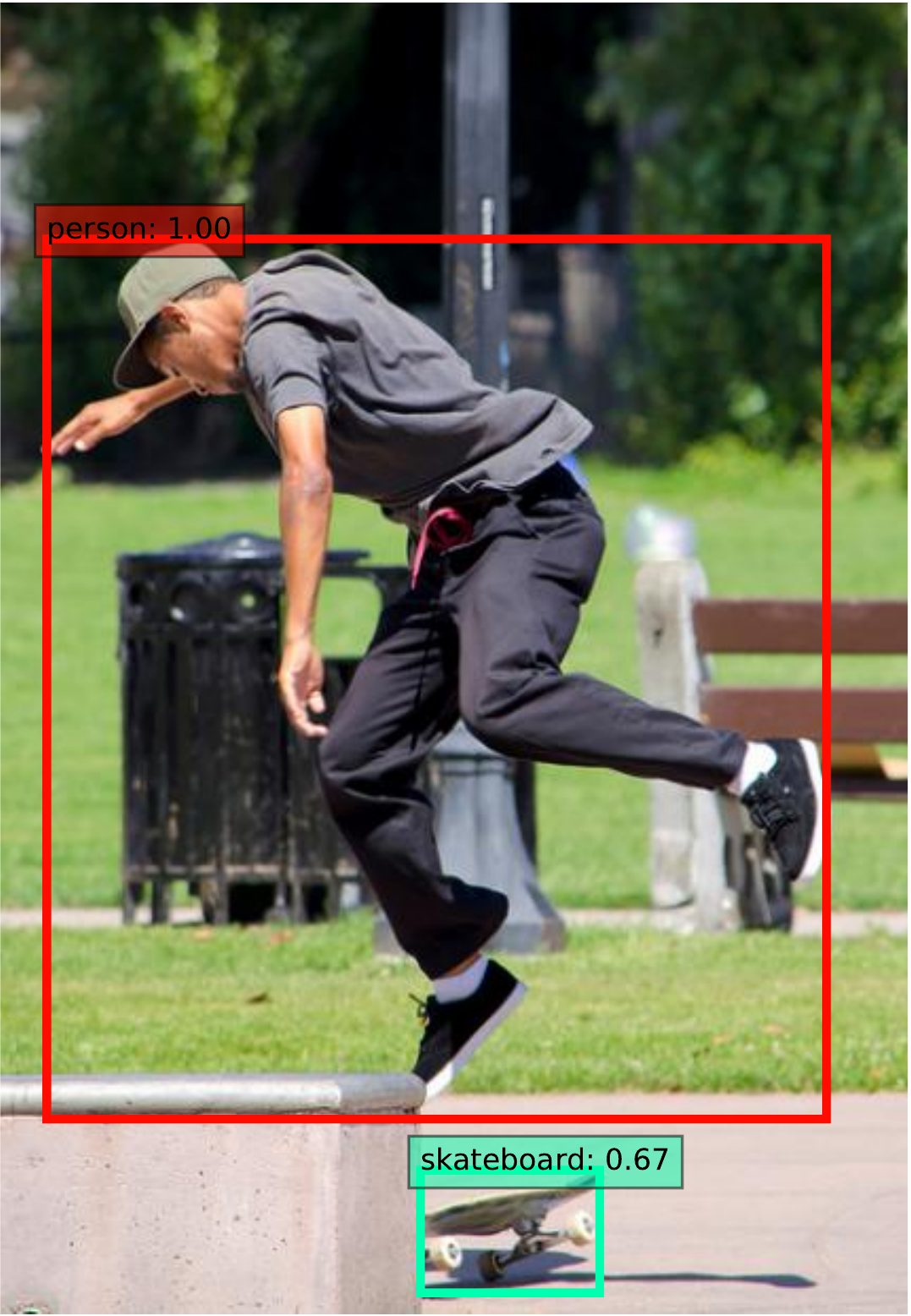}
    \includegraphics[width=0.24\linewidth]{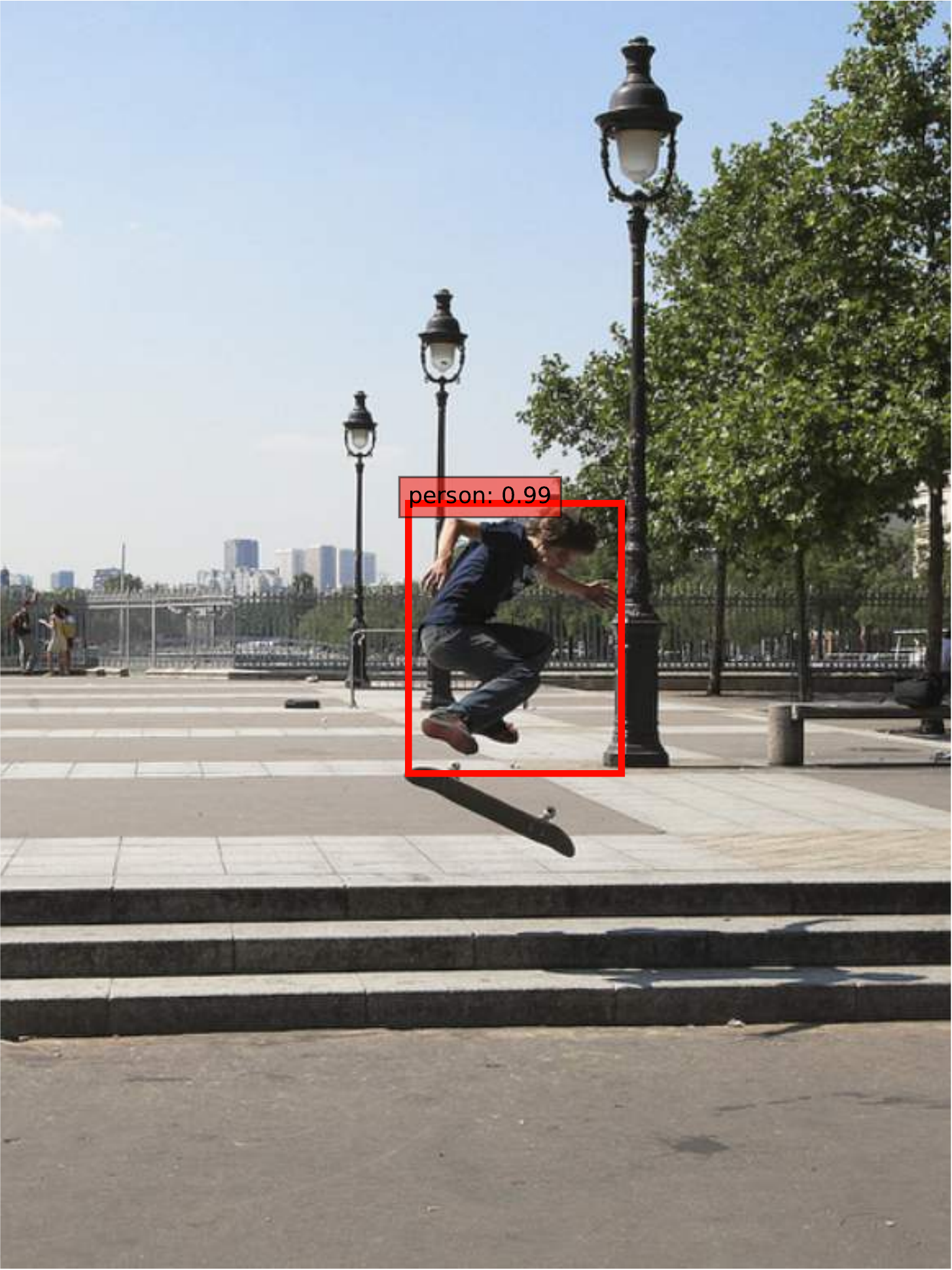}
    \includegraphics[width=0.24\linewidth]{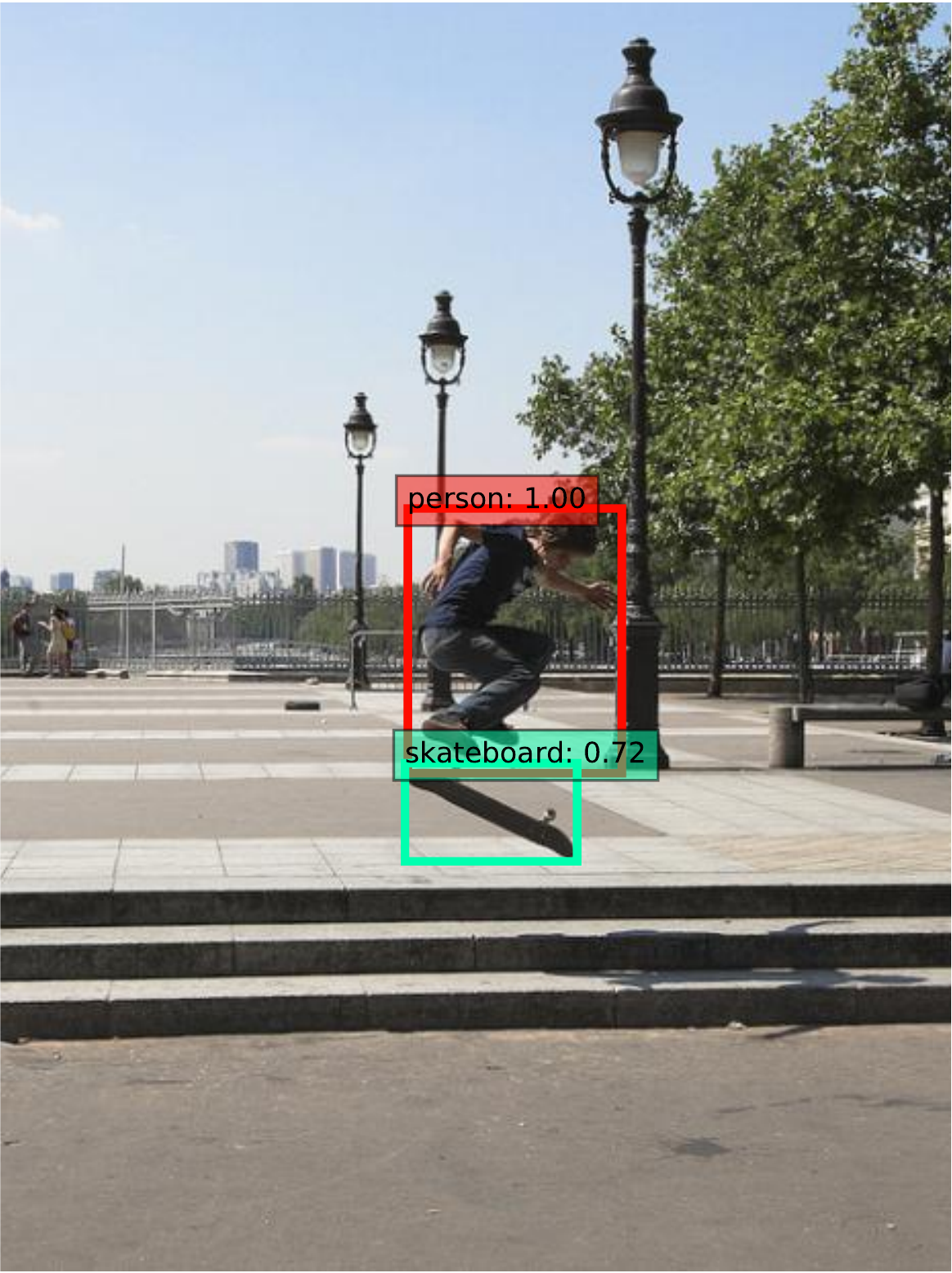}
    \\
     \caption{\textbf{Scene Context Cases}: DSSD also captures scene context, cleanly handling a wide variety of scenes and geometric configurations.}
    \end{subfigure}
    
	\caption{
    (a) on previous page. (b) above. 
    \textbf{Detection examples on COCO \texttt{test-dev} with SSD321/DSSD321 model.} For each pair, the left side is the result of SSD and right side is the result of DSSD. We show detections with scores higher than 0.6. Each color corresponds to an object category. }
      \label{sfig:context}
    \label{fig:coco}
\end{figure*}

\clearpage

\end{document}